\documentclass{article}

\usepackage{microtype}
\usepackage{graphicx}
\usepackage{subcaption}
\usepackage{booktabs} 

\usepackage{hyperref}

\usepackage[accepted]{icml2026}

\usepackage{amsmath}
\usepackage{amssymb}
\usepackage{mathtools}
\usepackage{amsthm}
\usepackage{bm}

\usepackage{booktabs} 
\usepackage[T1]{fontenc}  
\usepackage{xcolor,colortbl}
\usepackage{makecell}
\usepackage{url}            
\usepackage[capitalize,noabbrev]{cleveref}
\newcommand{\norm}[1]{\left\lVert #1 \right\rVert}

\usepackage{multicol}
\usepackage{multirow}

\usepackage[disable,textsize=tiny]{todonotes}

\theoremstyle{plain}
\newtheorem{theorem}{Theorem}[section]
\newtheorem{lemma}[theorem]{Lemma}
\newtheorem{assumption}[theorem]{Assumption}

\begin{document}


\twocolumn[
  \icmltitle{Cooperative Variance Estimation and Bayesian Neural Networks for Disentangling Aleatoric and Epistemic Uncertainties}



  \icmlsetsymbol{equal}{*}

\begin{icmlauthorlist}
\icmlauthor{Jiaxiang Yi}{delft}
\icmlauthor{Miguel A. Bessa}{brown}

\end{icmlauthorlist}

\icmlaffiliation{delft}{Faculty of Mechanical Engineering, Delft University of Technology, Mekelweg 2, Delft, 2628 CD, The Netherlands}
\icmlaffiliation{brown}{School of Engineering, Brown University, 184 Hope St., Providence, RI 02912, USA}

\icmlcorrespondingauthor{Miguel Bessa}{miguel\_bessa@brown.edu}

  \icmlkeywords{Machine Learning, ICML}

  \vskip 0.3in
]



\printAffiliationsAndNotice{}  

\begin{abstract}

Real-world data contains aleatoric uncertainty -- irreducible noise arising from imperfect measurements or from incomplete knowledge about the data generation process. Mean-variance estimation networks can learn this type of uncertainty but require ad-hoc regularization strategies to avoid overfitting and are unable to predict epistemic uncertainty (model uncertainty). Conversely, Bayesian neural networks predict epistemic uncertainty but are notoriously difficult to train due to the approximate nature of Bayesian inference. We propose to cooperatively train a variance estimation network with a Bayesian neural network and empirically demonstrate that the resulting model disentangles aleatoric and epistemic uncertainties while improving the mean estimation. We demonstrate the effectiveness and scalability of this method across a diverse range of datasets, including a time-dependent heteroscedastic regression dataset we created where the aleatoric uncertainty is known.  The proposed method is straightforward to implement, robust, and adaptable to various model architectures. Code is available at \url{https://github.com/bessagroup/VeBNN}.

\end{abstract}

\section{Introduction}
\label{sec:introduction} 

Non-probabilistic neural networks that only estimate the mean (expected value) tend to be overconfident and vulnerable to adversarial attacks \cite{guo2017calibration}. Quantifying aleatoric (or data) uncertainty alleviates these issues by characterizing noise \cite{Skafte2019, Seitzer2022}. Estimating epistemic (or model) uncertainty enables active learning and risk-sensitive decision-making \cite{Kendall2017,depeweg2018decomposition}. Consequently, except in cases with negligible or homoscedastic data noise, simultaneously predicting aleatoric and epistemic uncertainties is essential for a wide range of safety-critical applications \cite{hullermeier2021aleatoric}. In such cases, an outcome with good mean performance but large aleatoric uncertainty may be undesired \cite{Lakshminarayanan2017}. Therefore, the principle of reducing epistemic uncertainty behind active learning or decision-making needs to be balanced by the respective aleatoric uncertainty. This is particularly essential when the aleatoric uncertainty is heteroscedastic (input-dependent) due to imperfect measurements, environmental variability, and other factors \cite{smith2024rethinkingaleatoricepistemicuncertainty}.

\begin{figure*}[h]
\centering
\centerline{\includegraphics[width=0.80\textwidth]{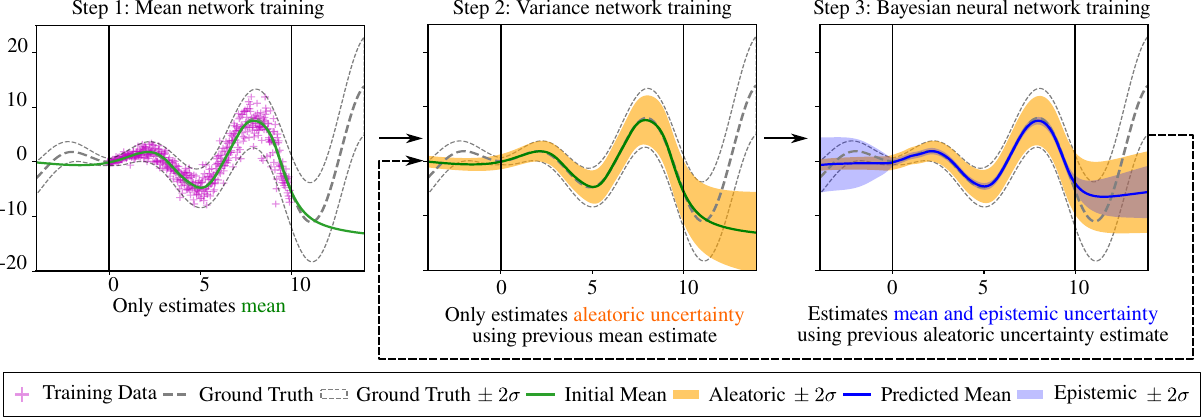}}
\caption{\textbf{Illustration of the proposed cooperative training strategy leading to Variance estimation Bayesian Neural Networks (VeBNNs).} The left figure shows the unseen ground truth mean as well as the respective training data. The method starts by training the mean network to only estimate the mean (green solid line in Step 1). Then, without updating the mean estimate, a variance network is trained to only predict aleatoric uncertainty (orange credible interval in Step 2). Subsequently, considering this aleatoric uncertainty estimation, a Bayesian neural network is trained to obtain an updated mean and corresponding epistemic uncertainty (solid blue line for the new mean, and shaded blue credible interval for the epistemic uncertainty in Step 3). If needed, the method can iterate between steps 3 and 2 to improve the disentanglement of uncertainties. Note the disentanglement of uncertainties together with the improvement of the mean estimation away from the data support ($x<0$ and $x>10$ in Step 3). }
\label{fig:illustration}
\vskip -0.15in
\end{figure*}

\textbf{Summary of contributions.} We propose a cooperative learning strategy for uncertainty disentanglement based on sequential training of (1) a mean network, (2) a variance network, and (3) a Bayesian neural network. \cref{fig:illustration} illustrates the method for one-dimensional heteroscedastic regression, briefly describing it at the figure caption. 

\section{Related Work}
\label{sec:related work}

\subsection{Aleatoric Uncertainty}
\label{sec:aleatoric uncertainty}

Aleatoric uncertainty can be estimated by parametric or nonparametric models. The latter do not explicitly define the likelihood function and instead focus on learning to sample from the data distribution \cite{mohamed2016learning}. Their ability to estimate nontrivial aleatoric uncertainty comes at the cost of training difficulties and sampling inefficiencies \cite{Harakeh2023}. Therefore, parametric models are more common. They assume a parameterized observation distribution, usually a Gaussian as in Mean Variance Estimation (MVE) networks \cite{nix1994estimating}, and learn the corresponding parameters by minimizing the Negative Log-Likelihood (NLL) loss with associated regularization. Similar parametric models have been developed, replacing the Gaussian distribution with other distributions \cite{meyer2020learning}. However, training these models can be challenging. MVE networks have been observed to lead to good mean but overconfident variance estimations in regions within the data support, and exhibit generalization issues outside these regions \cite{Skafte2019}. As analyzed by different authors \cite{Skafte2019, Seitzer2022, sluijterman2024optimal}, these issues result from the loss gradients having very different magnitudes when calculated with respect to the mean or the aleatoric variance, as seen in \Cref{eq:muder_Lvar}, which creates imbalances when minimizing the NLL.

Different solutions have been proposed to improve the parametric estimation of aleatoric uncertainty, including a Bayesian treatment of variance \cite{Stirn2020}, calibration by distribution matching via maximum mean discrepancy loss \cite{cui2020calibrated}, or modifying the loss function to include an additional balance hyperparameter \cite{Seitzer2022}. However, \citet{sluijterman2024optimal} demonstrated that training an MVE network that simultaneously predicts mean and variance leads to significantly worse predictions for both estimations compared to separate training, even when considering the above-mentioned modified losses (we independently reached the same conclusion while conducting our work; see \cref{sec:losses_comparison}). Nevertheless, we note that parametric deterministic models are insufficient if they only estimate mean and aleatoric uncertainty, without accounting for epistemic uncertainty. This is also visible in \cref{fig:illustration} (Step 2), as the model has an overconfident mean outside the data support (see $x>10$). Furthermore, the inability to estimate epistemic uncertainty is problematic in itself, as discussed in \cref{sec:introduction}.

\subsection{Epistemic Uncertainty}

Epistemic uncertainty is usually estimated by probabilistic models that impose a prior distribution on the model parameters \cite{Abdar2021}. Instead of finding point estimates as in deterministic models, they predict a posterior predictive distribution (PPD). Unfortunately, accurate and computationally tractable determination of the PPD is challenging in most cases, except for a few models like Gaussian processes, where inference can be done exactly under strict assumptions and with limited data scalability \cite{Rasmussen2005}. Bayesian neural networks (BNNs) are more scalable, but the accuracy and scalability are strongly dependent on the type of Bayesian inference \cite{Wilson2020, Wenzel2020, izmailov2021bayesian}.

Bayesian inference is commonly performed using Markov Chain Monte Carlo (MCMC) sampling methods \cite{Neal1995, Welling2011, li2016preconditioned} or Variational Inference (VI) methods, which are faster to train but less accurate \cite{Graves2011, Blundell2015}. Avoiding formal Bayesian inference is also possible by adopting ensemble methods such as Monte Carlo (MC) Dropout \cite{Gal2016} and deep ensembles \cite{Lakshminarayanan2017, fort2020deepensembleslosslandscape}. Although not strictly Bayesian, MC-Dropout can be interpreted as a Variational Bayesian approximation that has additional scalability but even lower accuracy (no free lunch).

The practical difficulties of training BNNs have limited their widespread use. Therefore, they are often trained by disregarding aleatoric uncertainty or assuming homoscedastic noise, which can be set as a hyperparameter but is impractical for complex problems \cite{Abdar2021}. Instead, the end-to-end training of an MVE network by MC-Dropout \cite{Kendall2017} or deep ensembling \cite{wang2025credal} is reported to approximately disentangle aleatoric and epistemic uncertainties for heteroscedastic regression and classification. However, the essential challenges faced when training MVE networks are not solved by ensembling them, and the lack of accuracy associated with MC-Dropout raises questions about their ability to make high-quality predictions and truly disentangle epistemic and aleatoric uncertainties \cite{ValdenegroToro2022, mucsanyi2024benchmarking}. This has motivated other authors to propose different solutions. A non-Bayesian solution to separate uncertainties was proposed by introducing a high-order evidential distribution, i.e., considering priors over the likelihood function instead of over network weights \cite{Amini2020}. Another proposal has been to use a natural reparameterization combined with an approximate Laplace expansion to estimate epistemic uncertainty \cite{Immer2023}. Still, a recent investigation \cite{mucsanyi2024benchmarking} suggests that no current method achieves reliable uncertainty disentanglement.

\section{Methodology}

Consider a dataset $\mathcal{D} = \{\mathbf{x}_n, y_n\}^N_{n=1}$ with \emph{i.i.d.} data points, where $\mathbf{x}_n \in \mathbb{R}^{d}$ represents the input features and $y_n \in \mathbb{R}$ the corresponding outputs\footnote{The equations become simpler when writing for one-dimensional outputs, but this article includes examples with multi-dimensional outputs $\mathbf{y}_n \in \mathbb{R}^m$, as shown later.}. The heteroscedastic regression problem can be formulated as:
\begin{equation} 
\label{eq:problem_set_up}
     y = f(\mathbf{x}) + \epsilon(\mathbf{x})
\end{equation}
where $f(\mathbf{x})$ denotes the underlying noiseless ground truth function (expected mean), and $\epsilon(\mathbf{x})$ is the corresponding heteroscedastic noise (aleatoric uncertainty). If the noise is Gaussian, then $\epsilon(\mathbf{x}) \sim \mathcal{N}(0, s^2(\mathbf{x}))$ where $ s^2(\mathbf{x})$ represents its ground truth input-dependent variance.

\subsection{Challenges of Joint Training of MVE}
\label{sec:mean_training}

The NLL loss used for training MVE networks resulting from a Gaussian observation distribution with heteroscedastic aleatoric uncertainty is:
\begin{equation}
\label{eq:Lvar}
    \mathcal{L}_1(\bm{\theta}, \bm{\phi}) = \sum_{n=1}^N \left[\frac{(y_n - \mu(\mathbf{x}_n; \bm{\theta}))^2}{2 \sigma_a^2(\mathbf{x}_n;\bm{\phi})} + \frac{1}{2}\log\big(\sigma_a^2(\mathbf{x}_n;\bm{\phi})\big)\right]
\end{equation}
where $\sigma_a^2(\mathbf{x};\bm{\phi}) > 0$ is the parameterized aleatoric variance, and $\mu(\mathbf{x}; \bm{\theta})$ the mean. The derivatives with respect to $\mu(\mathbf{x};\bm{\theta})$ and $\sigma_a^2(\mathbf{x};\bm{\phi})$ become:
\begin{align}
\label{eq:muder_Lvar}
    \nabla_{\mu} \mathcal{L}_1 &= \sum_{n=1}^{N}\left( \frac{\mu(\mathbf{x}_n;\bm{\theta}) - y_n}{\sigma_a^2(\mathbf{x}_n;\bm{\phi})}\right),  \notag \\
    \nabla_{\sigma_a^2} \mathcal{L}_1 &= \sum_{n=1}^{N}\left( \frac{\sigma_a^2(\mathbf{x}_n;\bm{\phi}) - (y_n - \mu(\mathbf{x}_n;\bm{\theta}))^2}{2\big(\sigma_a^2(\mathbf{x}_n;\bm{\phi})\big)^2}\right).
\end{align}

According to \cref{eq:muder_Lvar}, the gradient with respect to $\sigma_a^2(\mathbf{x};\bm{\phi})$ has a high-order term in the denominator that makes the optimization more challenging and causes an imbalance when minimizing the loss. We also show in \cref{sec:theory_justification_steps1_and_2} that the loss of jointly trained MVE networks is non-convex in its outputs and has no finite stationary point (\cref{lem:mve_no_critical_point}). This explains the above-mentioned optimization challenges that have been empirically observed in the literature and that we also observe in our work.  As a result, joint (or end-to-end) training of MVE networks can lead to unexpected outcomes, for example: (1) if an MVE network overfits the response, i.e., $(y_n - \mu(\mathbf{x}_n; \boldsymbol{\theta}))^2 \to 0$, $\sigma_a^2(\mathbf{x}_n; \boldsymbol{\phi}) \to 0$, the loss function $\mathcal{L}_1$ may become undefined; (2) if the MVE network leads to large variance estimates when quantifying aleatoric uncertainty, then $\sigma_a^2(\mathbf{x}_n; \boldsymbol{\phi}) \to \infty$ and the gradients with respect to $\sigma_a^2(\mathbf{x}_n; \boldsymbol{\phi})$ vanish. This indefinite curvature translates into low-quality parameter point estimates or slower convergence, as training drifts toward degenerate regions (see \Cref{sec:joint_MVE_training_verse_cooperative_training}). While balancing the loss is possible, as proposed in \cite{Seitzer2022} by the $\beta$-NLL loss, this introduces additional hyperparameters that need to be optimized for each problem. Furthermore, conducting Bayesian inference for MVE networks such that we simultaneously predict aleatoric and epistemic uncertainties should only make these issues more acute, as evidenced in the literature \cite{mucsanyi2024benchmarking}.

\subsection{VeBNN: Cooperative Variance Estimation Bayesian Neural Networks}

\begin{algorithm}[h]
\small
\caption{Cooperative VeBNN training}
\label{alg: CUQalgorithm}
\begin{algorithmic}
  \STATE {\bfseries Input:} mean network ${\mu}(\mathbf{x}; \bm{\theta})$, variance network ${\sigma_a^2}(\mathbf{x}; \bm{\phi})$, dataset $\mathcal{D} = \{\mathbf{X}, \mathbf{y} \}$, number of iterations $K$ \\
    {\bf Step 1: mean network training:} Minimize \cref{eq:Lvar}  for constant $\sigma_a^2$ to find point estimate of $\hat{\mu}(\mathbf{x};\hat{\boldsymbol{\theta}})$. \\
   \FOR{$i = 1$ {\bfseries to} $K$}
       \STATE {\bf Step 2: variance network training (Aleatoric uncertainty)}:  Minimize \cref{eq:gamma_loss} for fixed mean from Step 1 or Step 3  to find point estimate $\hat{\sigma}^2_a(\mathbf{x}; \hat{\boldsymbol{\phi}})$ from \cref{eq:aleatoric_variance}. 
        \STATE {\bf Step 3: Bayesian network training (Epistemic uncertainty)}:  Obtain PPD to determine mean and epistemic variance estimates for fixed aleatoric variance found in Step 2; and compute the log marginal likelihood ( $\text{\texttt{LMglk}}[i] = \log \mathbb{E}_{p(\bm{\theta}|\mathcal{D})} \left[ p(\bm{y} \mid \bm{\theta}^{(i)}, \bm{\phi}^{(i)}) \right] $) to obtain the posterior sample set $\boldsymbol{\Theta}$
   \ENDFOR
    \STATE Identify the optimal parameters $\bm{\Theta}^*$ and $\bm{\phi}^*$ by $i^{*}=\arg\max_{i>1}\texttt{LMglk}[i]$
\end{algorithmic}
\end{algorithm}

We propose a sequential training strategy (see \cref{alg: CUQalgorithm}) that starts with training a mean network, then a variance estimation network that predicts aleatoric uncertainty, followed by a BNN that updates both mean and epistemic uncertainty for the previously determined aleatoric uncertainty. By separating the roles of each network and ensuring their cooperative training, we demonstrate that the resulting BNN can learn and improve all three estimates. Importantly, training each network separately is easier, and we show that inference of the BNN is also facilitated due to the presence of a good estimate of aleatoric uncertainty (that is not being learned at that stage). \cref{sec:bias_improvement_by_VeBNNs} includes theoretical arguments in favor of our cooperative training strategy, also discussing the mechanism behind the observed convergence for growing $K$.

The proposed \cref{alg: CUQalgorithm} starts by considering constant aleatoric uncertainty, i.e. $\sigma_a^2(\mathbf{x};\bm{\phi}) = \text{constant}$, and conventionally trains the mean network by finding the maximum a posteriori (MAP) estimate of the parameters $\bm{\theta}$ of \Cref{eq:Lvar}, hence determining only the mean. Because $\sigma_a^2(\mathbf{x};\bm{\phi}) = \text{constant}$, it cancels from the gradient and does not affect the learned parameters.

\subsubsection{Variance Estimation Network Training}
\label{sec:variance_training}

Once the mean is obtained, we then train the variance estimation (Ve) network for this fixed mean. There are, however, multiple important details that facilitate training of this network (Step 2 in \cref{alg: CUQalgorithm}). The variance estimation network does not directly output $\sigma_a^2(\mathbf{x};\bm{\phi})$, and so its parameters are not directly determined by minimizing \cref{eq:Lvar} and keeping the mean fixed. Instead, the variance network models the squared residual $r = \left(\mu(\mathbf{x}; \bm{\theta}) - y\right)^2$, which follows a Gamma distribution because $y$ is Gaussian. 
\begin{assumption}
\label{ass:finite_bias}
$\left(\mu(\mathbf{x}; \bm{\theta}) - f(\mathbf{x})\right)^2$ is finite and converges to zero when $N \rightarrow \infty$. This follows from assuming unbiased or consistent estimations for the ground truth $f(\mathbf{x})$, regardless of noise.
\end{assumption}

\begin{lemma}
Given \cref{ass:finite_bias} and assuming that $y$ follows a Gaussian distribution, the squared residual $r = \left(\mu(\mathbf{x};\bm{\theta}) - y\right)^2$ is asymptotically Gamma-distributed.
\end{lemma}
\begin{proof}
    Since $y \mid \mathbf{x} \sim \mathcal{N}(f(\mathbf{x}), s^2(\mathbf{x}))$, the residual $\mu(\mathbf{x}; \bm{\theta}) - y \sim \mathcal{N}(0,s^2(\mathbf{x}))$ asymptotically, by \cref{ass:finite_bias}. By standardizing, we obtain:
    \begin{equation}
    Z = \frac{\mu(\mathbf{x}; \bm{\theta}) - y}{s(\mathbf{x})} \sim \mathcal{N}(0, 1).
    \end{equation}
   Thus, the squared residual can be obtained: $ r = \left( \mu(\mathbf{x}; \bm{\theta}) - y \right)^2 = s^2(\mathbf{x}) Z^2 $. Since $ Z \sim \mathcal{N}(0, 1) $, we have \( Z^2 \sim \chi^2(1) \), a specific case of Gamma distribution $Z^2 \sim \text{Gamma}\left( \frac{1}{2}, \frac{1}{2} \right)$. Since a Gamma random variable $W \sim \text{Gamma}(\alpha, \lambda)$ scaled by a constant $ c > 0 $ gives $c W \sim \text{Gamma}\left( \alpha, \frac{\lambda}{c} \right)$, then: 
\begin{equation}
     r \sim \text{Gamma}\left( \frac{1}{2}, \frac{1}{2s^2(\mathbf{x})} \right)
\end{equation}
Showing that the mean of the Gamma distribution becomes $\frac{\alpha}{\lambda} =  s^2(\mathbf{x})$, i.e., the aleatoric variance. 
\end{proof}

We propose the Gamma likelihood to model the squared residual $r$, leading to the corresponding NLL loss to train the variance network:
\begin{multline}
\label{eq:gamma_loss}
\mathcal{L}_2(\boldsymbol{\phi}) = \sum_{n=1}^{N}
    \Big[
        \log \Gamma\!\big(\alpha(\mathbf{x}_n; \boldsymbol{\phi})\big)
        - \alpha(\mathbf{x}_n; \boldsymbol{\phi}) 
          \log \lambda(\mathbf{x}_n; \boldsymbol{\phi})
         \\  - \big(\alpha(\mathbf{x}_n; \boldsymbol{\phi}) - 1\big) \log r_n
        + \lambda(\mathbf{x}_n; \boldsymbol{\phi})\, r_n
    \Big].
\end{multline}
where $r$ are the above-mentioned residuals, and with the shape and rate parameters of the Gamma distribution, $\alpha(\mathbf{x}; \bm{\phi}) > 0$ and $\lambda(\mathbf{x}; \bm{\phi}) > 0$, being the outputs of the network. These parameters are obtained by MAP estimate, and their gradients with respect to $\alpha$ and $\lambda$ do not contain high-order terms, as shown below:
\begin{align}
\label{eq:gamma_grads}
\nabla_{\alpha(\mathbf{x})}\mathcal{L}_2 
&= \sum_{n=1}^N 
\Big[
    \psi\big(\alpha(\mathbf{x}_n)\big)
    - \log \lambda(\mathbf{x}_n)
    - \log r_n
\Big],
\notag \\
\nabla_{\lambda(\mathbf{x})}\mathcal{L}_2 
&= \sum_{n=1}^N 
\Big[
    - \frac{\alpha(\mathbf{x}_n)}{\lambda(\mathbf{x}_n)}
    + r_n
\Big].
\end{align}
The expected value of the Gamma distribution becomes the desired heteroscedastic variance:
\begin{equation} 
\label{eq:aleatoric_variance}
     \sigma_a^2(\mathbf{x};\bm{\phi}) = \frac{\alpha(\mathbf{x}; \bm{\phi})}{\lambda(\mathbf{x}; \bm{\phi})}
\end{equation}

\subsubsection{Bayesian Neural Network Training}

Having determined the mean and aleatoric variance estimates, we then train a BNN with a warm-start for the mean, and assuming fixed aleatoric variance that is obtained from \cref{eq:aleatoric_variance}. In principle, the same network architecture can be used for the BNN and the mean network\footnote{Estimating epistemic uncertainty with BNNs can require a network with wider hidden layers when compared to a deterministic network that only estimates the mean. So, choosing a smaller mean network in Step 1 is also possible. However, we like the simplicity of not introducing a new network.}. The posterior of the BNN is determined by Bayes' rule, and it is proportional to the product of likelihood and prior:
\begin{equation}
\label{eq:posterior}
    p \left ( \boldsymbol{\theta} \mid \mathcal{D} \right ) \propto
     p \left ( \mathcal{D} \mid \boldsymbol{\theta} \right) p \left( \boldsymbol{\theta} \right) 
\end{equation}
where $p \left( \boldsymbol{\theta} \right)$ is the prior over the neural network parameters, and $p \left ( \mathcal{D} \mid \boldsymbol{\theta} \right)$ is the likelihood for the observations. In this work, we consider a Gaussian prior with zero mean and unit variance, and the likelihood is given by \cref{eq:Lvar}, i.e., it arises from a Gaussian observation distribution with heteroscedastic noise. The logarithm of the posterior becomes:
\begin{multline}
\label{eq:expansion of posterior}
\log p \left( \boldsymbol{\theta} \mid \mathcal{D} \right) =\\
 \sum_{n=1}^N \left[\log \frac{1}{\sqrt{2 \pi \sigma_a^2(\mathbf{x}_n; \boldsymbol{\phi})}} - \frac{\left( \mathbf{y}_n - \mu(\mathbf{x}_n; \boldsymbol{\theta}) \right)^2}{2 \sigma_a^2(\mathbf{x}_n; \boldsymbol{\phi})} \right] \\
 + m\log \frac{1}{\sqrt{2 \pi / \kappa}} - \frac{\kappa}{2} \|\boldsymbol{\theta}\|^2
\end{multline}
where $\kappa = 1.0$  is the precision of the prior distribution and $m$ is the length of $\bm{\theta}$. Note that \cref{eq:expansion of posterior} is the negative of \cref{eq:Lvar}, but now we explicitly include the prior terms.
\cref{sec:details of BNNs} summarizes common Bayesian inference strategies, including the MCMC-based methods that sample directly from \cref{eq:expansion of posterior}, and VI methods that minimize the evidence lower bound (ELBO) (\Cref{sec:details of MCMC,sec:details of VI}, respectively).  Importantly, since the aleatoric variance $\sigma_a^2(\mathbf{x}_n; \boldsymbol{\phi})$ is fixed from step 2 and is regarded as constant, it converges to the correct mode $\mu(\mathbf{x}_n; \boldsymbol{\theta})$, and has better and faster convergence (no gradient issue).

From the PPD of the BNN, we then estimate the mean, aleatoric, and epistemic variances (disentangled) for any unseen point $\mathbf{x}^\prime$ based on Bayesian model averaging through the obtained $\mathbb{E}_{p(\bm{\theta} \mid \mathcal{D})}$ using the chosen inference method:
\begin{equation}
\mathcal{N}\left( 
    \mathbf{y}^\prime \,\middle|\, 
    \underbrace{\mathbb{E}_{p(\bm{\theta} \mid \mathcal{D})}\big[\mu(\mathbf{x}^\prime; \bm{\theta})\big]}_{\text{Predictive Mean}}, \, 
    \underbrace{\sigma_a^2(\mathbf{x}^\prime; \bm{\phi})}_{\text{Aleatoric}} + 
    \underbrace{\mathbb{V}_{p(\bm{\theta} \mid \mathcal{D})}\big[\mu(\mathbf{x}^\prime; \bm{\theta})\big]}_{\text{Epistemic}}
\right)
\label{eq:bayes_model_average}
\end{equation}

\section{Experiments}
\label{sec:experiments}

\paragraph{Datasets} We consider five distinct sets of datasets: (1) the previously discussed one-dimensional illustrative example \cite{Skafte2019}, with results presented in \cref{sec:synthetic_datasets}; (2) UCI regression datasets \cite{HernandezLobato2015}; (3) 
 large-scale image regression datasets \cite{gustafsson2023reliableregressionmodelsuncertainty}; (4) our own dataset obtained from computer simulations of materials undergoing history-dependent deformations (material plasticity law discovery dataset, readers interested in more details about the dataset are referred to \cref{sec:des_plasticity_law}); and (5) APPA-REAL dataset containing multiple labels for each face image \cite{agustsson2017appareal}. 

\textbf{An important challenge in assessing the performance of methods that disentangle uncertainties is that existing datasets do not provide the ground-truth aleatoric uncertainty to assess the accuracy of its estimation}. Therefore, it is worth mentioning that the UCI and Image regression datasets are mainly used to assess the quality of the mean and total uncertainty estimations despite having \textbf{unknown aleatoric uncertainty}. The last two datasets mentioned above (4 and 5) are the only ones where we \textbf{know the ground-truth of aleatoric uncertainty} for assessing the estimation accuracy of uncertainty disentanglement, and they assume particular importance for this investigation. 

\paragraph{Baselines} We compare against representative methods: (1) \textbf{ME (MSE)}: standard Mean Estimation network;  (2) \textbf{MVE ($\beta\text{-NLL}$)} \cite{Seitzer2022} and (3) \textbf{MVE (Natural)} \cite{Immer2023}: MVE using the $\beta\text{-NLL}$  and natural re-parameterized NLL loss, which are capable of simultaneous prediction of mean and aleatoric uncertainty.  Concerning methods that aim at disentangling uncertainties, we considered (4) \textbf{Evidential}: Deep Evidential regression \cite{Amini2020}; as well as 4 different inference methods that we apply to MVE networks, leading to (5) deep ensembling \cite{Lakshminarayanan2017}, MC-Dropout \cite{Kendall2017}, Bayes By Backpropagation (\textbf{BBB}) \cite{Blundell2015}, and preconditioned Stochastic Gradient Langevin Dynamics (pSGLD) \cite{li2016preconditioned}, which we respectively refer to as \textbf{MVE (Ensembles)} \cite{ValdenegroToro2022,fishkov2025},  \textbf{MVE (MC-Dropout)} \cite{ValdenegroToro2022}, \textbf{MVE (BBB)}, and \textbf{MVE (pSGLD)}. Finally, our method is referred to as \textbf{VeBNN} -- Variance estimation Bayesian Neural Network -- and trained by the proposed cooperative training strategy. For fair comparison, Step 3 of VeBNN training is done for all above-mentioned inference methods; denoted as VeBNN (BBB), VeBNN (Ensembles), etc.

Details about accuracy metrics, additional results, and training are presented in  \cref{sec:performance_metrics},  \cref{sec:additional_experiments}, and \cref{sec:hyperparams_setting}, respectively.

\subsection{Real World Datasets with Unknown Aleatoric Uncertainty}
\label{sec:real_world_datasets}

\paragraph{UCI regression datasets}
\label{sec:uci_regression}

The results for the UCI regression datasets are summarized in  \cref{tab:tll_uci} (and also in \cref{tab:rmse_uci}, \cref{tab:coverage_uci}, and \cref{tab:til_uci} of \Cref{sec:additional_results_UCI_regression}) according to the \textbf{Test log-likelihood (TLL)}, \textbf{Root Mean Square Error (RMSE)}, \textbf{Test coverage (TC)}, \textbf{test interval length (TIL)} metrics, respectively. VeBNN improves TLL and RMSE for the large majority of dataset–inference combinations, showing a clear advantage compared to jointly training a single MVE network. Interestingly, the cooperative training strategy is capable of transforming a ``weak'' inference method (BBB) and achieves competitive performance when compared with other VeBNN inference methods -- see differences in performance from MVE (BBB) to VeBNN (BBB). The test coverage is close to the target value of $0.95$ for nearly all methods, indicating well-calibrated predictive intervals. Interestingly, joint training leads to higher TLL (i.e., better) than the corresponding cooperative training strategy across all inference methods on the \emph{Yacht} problem. However, this improvement comes at the cost of worse mean prediction and a larger test interval length, indicating that jointly trained MVE overfits the noise.  

\begin{table*}[h]

\caption{\textbf{TLL ($\uparrow$) results on the UCI regression datasets.}
Each entry reports the mean TLL with the standard deviation in parentheses. The best method is marked in \textbf{\underline{bold}}, and the second-best in \textbf{bold}. A superscript $*$ indicates that the best method is significantly better than the second-best (Wilcoxon test \cite{demvsar2006statistical}).  Within each inference family, the best-performing variant is italicized; A superscript $+/-$ indicates significantly better or worse performance compared to the other variants in the same family ($p < 0.05$).}
\label{tab:tll_uci}
\centering
\resizebox{\textwidth}{!}{%
\begin{tabular}{lcccccccc}
\toprule
\multirow{2}{*}{Methods} & \multirow{2}{*}{\shortstack{Carbon\\ (10721, 7,1)}} & \multirow{2}{*}{\shortstack{Concrete\\ (1030, 8,1)}} & \multirow{2}{*}{\shortstack{Energy\\ (768,8,2)}} & \multirow{2}{*}{\shortstack{Boston\\ (506,13,1)}}  & \multirow{2}{*}{\shortstack{Power\\ (9568,4,1)}}   & \multirow{2}{*}{\shortstack{Superconduct\\ (21263,81,1)}} & \multirow{2}{*}{\shortstack{Wine-Red\\ (1599, 11,1)}} & \multirow{2}{*}{\shortstack{Yacht\\ (308,6,1)}}\\
& & & & & & & & \\
\midrule
MVE ($\beta\text{-NLL}=1.0$)
& 3.22 (0.50) & -3.11 (0.14) & -1.09 (0.29) & -2.94 (0.55) & -2.79 (0.06) & -3.80 (0.08) & \underline{\textbf{-0.95 (0.06)}} & -2.18 (1.34) \\

MVE (Natural)
& \textbf{3.88 (0.24)} & -3.07 (0.16) & -1.12 (0.15) & -2.74 (0.37) & -2.76 (0.04) & -3.47 (0.12) & -0.96 (0.07) & -1.69 (1.84) \\

Evidential
& -4.98 (13.79) & -3.22 (0.17) & -1.62 (0.31) & -2.84 (0.34) & -2.78 (0.06) & -3.65 (0.05) & -1.14 (0.25) & -1.63 (0.76) \\

\midrule
MVE (Ensembles)
& -7.72 (44.03) & -3.01 (0.24) & \textbf{\textit{-0.85 (0.23)}} & -2.71 (0.29) & -2.78 (0.07) & \textit{-3.43 (0.06)} & -1.05 (0.06) & \textit{-0.92 (0.79)} \\

\textbf{Ours:} VeBNN (Ensembles)
& \textit{3.87 (0.67)} & \textbf{\textit{-2.88 (0.15)}}$^{+}$ & -0.90 (0.08) & \textbf{\textit{-2.54 (0.32)}} & \textbf{\textit{-2.75 (0.04)}} & -3.51 (0.04)$^{-}$  & \underline{\textbf{\textit{-0.95 (0.06)}}}$^{+}$ & -1.13 (0.53)$^{-}$  \\
\midrule

MVE (MC-Dropout)
& 2.76 (0.12) & -3.02 (0.14) & -1.69 (0.22) & -2.69 (0.37) & -2.80 (0.03) & -3.48 (0.19) & \textit{-0.96 (0.12)} & \underline{\textbf{\textit{-0.16 (0.32)}}}* \\

\textbf{Ours:} VeBNN (MC-Dropout)
& \textit{3.19 (0.40)}$^{+}$ & \textit{-2.95 (0.12)} $^{+}$& \textit{-1.33 (0.07)}$^{+}$& \underline{\textbf{\textit{-2.42 (0.23)}}}*$^{+}$ & \textit{-2.79 (0.04)}$^{+}$ & \textit{-3.40 (0.20)}$^{+}$ & -0.99 (0.08)$^{-}$ & \textbf{-0.76 (0.32)}$^{-}$ \\

\midrule

MVE (BBB)
& -0.94 (0.01) & -140.59 (57.47) & -40.19 (12.11) & -47.97 (23.90) & -68.78 (51.33) & -980.85 (896.75) & -2.58 (0.75) & -174.45 (96.93) \\

\textbf{Ours:} VeBNN (BBB)
& \textit{3.23 (0.96)}$^{+}$ & \textit{-3.11 (0.23)}$^{+}$ & \textit{-1.26 (0.32)}$^{+}$ & \textit{-2.67 (0.29)}$^{+}$ & \textit{-2.82 (0.03)}$^{+}$ & \textit{-3.60 (0.07)}$^{+}$ & \textbf{\textit{-0.95 (0.07)}}$^{+}$ & \textit{-1.80 (0.80)}$^{+}$ \\
\midrule

MVE (pSGLD)
& -2.43 (17.66) & -3.07 (0.29) & -1.20 (0.48) & -2.57 (0.26) & -2.75 (0.07) & \textbf{-3.41 (0.20)} & -0.97 (0.11) & \textit{-1.00 (0.35)} \\

\textbf{Ours:} VeBNN (pSGLD)
& \underline{\textbf{\textit{4.04 (0.78)}}}$^{+}$ & \underline{\textbf{\textit{-2.87 (0.18)}}}$^{+}$ & \underline{\textbf{\textit{-0.85 (0.10)}}}$^{+}$ & \textit{-2.57 (0.21)} & \underline{\textbf{\textit{-2.74 (0.04)}}} & \underline{\textbf{\textit{-3.41 (0.09)}}} & \underline{\textbf{\textit{-0.95 (0.06)}}} & -1.27 (0.57) \\

\bottomrule
\end{tabular}
}
\end{table*}

\begin{table*}[hbt!]
\caption{\textbf{TLL ($\uparrow$) results on image regression datasets.} Each entry reports the mean TLL, with the standard deviation in parentheses.}
\label{tab:tll_imgreg}
\centering
\resizebox{\textwidth}{!}{%
\begin{tabular}{lcccccccc}
\toprule
Methods &
  Cells&
   Cells-Gap&
   Cells-Tail
    & ChairAngle
    & ChairAngle-Gap
    & ChairAngle-Tail
    &Skin
    & Aerial \\
& & & & & & & & \\
\midrule
MVE ($\beta\text{-NLL}=0.5$)
& -2.60 (0.31) & -5.21 (1.68) & -16.65 (10.46) & -0.50 (0.47)
& -17.46 (8.27) & -79.55 (71.32) & -8.19 (0.50) & -8.02 (0.52) \\

MVE (Natural) 
& -7.38 (3.49) & -973.22 (2163.47) & -6.53 (0.50) & -4.70 (0.01)
& \textbf{-4.69 (0.01)} &\textbf{ -5.07 (0.05)} & -2345.04 (4671.44) & -8.07 (0.35) \\

Evidential 
& -2.68 (0.17) & -5.73(2.05) & -20.03 (6.60) & -0.51 (0.43)
& -14.25 (4.94) & -62.44 (35.44) & -8.79 (0.63) & -8.65 (0.33) \\
\midrule
MVE (Ensembles) 
& -2.45 (0.06) & -3.89 (0.14) & -6.06 (0.78) & -0.32 (0.23)
& -5.86 (2.37) & -93.40 (32.28) & -7.61 (0.12) & -7.76 (0.27) \\

VeBNN (Ensembles) 
& \underline{\textbf{\textit{-2.31 (0.13)}}} & \textbf{\textit{-3.68 (0.15)}} & \textit{-5.19 (0.28)} & \underline{\textbf{\textit{-0.09 (0.24)}}}
& \textit{-5.36 (3.46)} & \textit{-71.10 (33.66)} & \textit{-7.59 (0.13)} & \textbf{\textit{-7.44 (0.05)}}$^+$ \\
\midrule
MVE (pSGLD) 
& -3.53 (0.56) & -3.84 (0.39) & \underline{\textbf{\textit{-4.84 (0.57)}}} & -2.84 (0.22)
& \underline{\textbf{\textit{-3.45 (0.34)}}}$^*$ & \underline{\textbf{\textit{-4.69 (0.38)}}} & \textbf{-7.42 (0.07)} & -8.94 (0.54) \\

VeBNN (pSGLD) 
& \textbf{\textit{-2.42 (0.11)}}$^+$ & \underline{\textbf{\textit{-3.56 (0.18)}}} & \textbf{-4.97 (1.86)} & \textbf{\textit{-0.14 (0.16)}}$^+$
& -6.02 (3.19)$^-$ & -46.56 (30.92)$^-$ & \underline{\textbf{\textit{-7.38 (0.14)}}} & \underline{\textbf{\textit{-7.42 (0.13)}}}$^+$ \\

\bottomrule
\end{tabular}
}
\end{table*}

\paragraph{Image regression datasets}
\label{sec:image_regression}

We considered 8 large-scale image regression datasets under real-world distribution shifts  \cite{gustafsson2023reliableregressionmodelsuncertainty} that evaluate the mean and total uncertainty estimations trained on ResNet 34 architecture \cite{He_2016_CVPR}.  The TLL results are reported in \Cref{tab:tll_imgreg}, while the RMSE, TC, and TIL results are presented in \Cref{tab:rmse_imgreg}, \Cref{tab:coverage_imgreg}, and \Cref{tab:til_imgreg}, respectively (\cref{sec:image_regression_additional_results}). Results for BBB and MC-Dropout are omitted due to convergence issues. The MVE (Ensembles) baseline is implemented with $\beta\text{-NLL}=0.5$, which provides the strongest competing performance for this dataset. Across the \textit{Cells} and \textit{ChairAngle} datasets, both without distribution shift, VeBNN shows consistent improvements over all metrics for both Ensembles and pSGLD. More interestingly, on the remaining datasets that exhibit distribution shift, VeBNN (Ensembles) achieves higher TLL and better test coverage, demonstrating a stronger ability to deal with shifted data. We also note that VeBNN (pSGLD) performs comparably to MVE (Ensembles), despite the fact that MVE (pSGLD) tends to overfit the aleatoric variance.

In \cref{sec:image_regression_additional_results}, we further show that VeBNN increases epistemic uncertainty in regions affected by distribution shift, indicating the expected reduction in model confidence. However, both aleatoric and epistemic uncertainties increase simultaneously, whereas only epistemic uncertainty is expected to rise under a distribution shift. This indicates that part of the uncertainty increase may be attributed to undesirable aleatoric uncertainty inflation. In addition, MVE (Ensembles) is highly sensitive to the choice of $\beta$. As shown in \cref{sec:image_regression_additional_results}, when using the original NLL loss, MVE (Ensembles) tends to overfit the data noise while underfitting the mean.

\subsection{Datasets with Known Aleatoric Uncertainty}
\subsubsection{Material Plasticity Law Discovery}
\label{sec:plasticity_law_experiment}

\begin{figure*}[h]
\centering
\vspace{-0.10in}
\centerline{\includegraphics[width=0.80\textwidth]{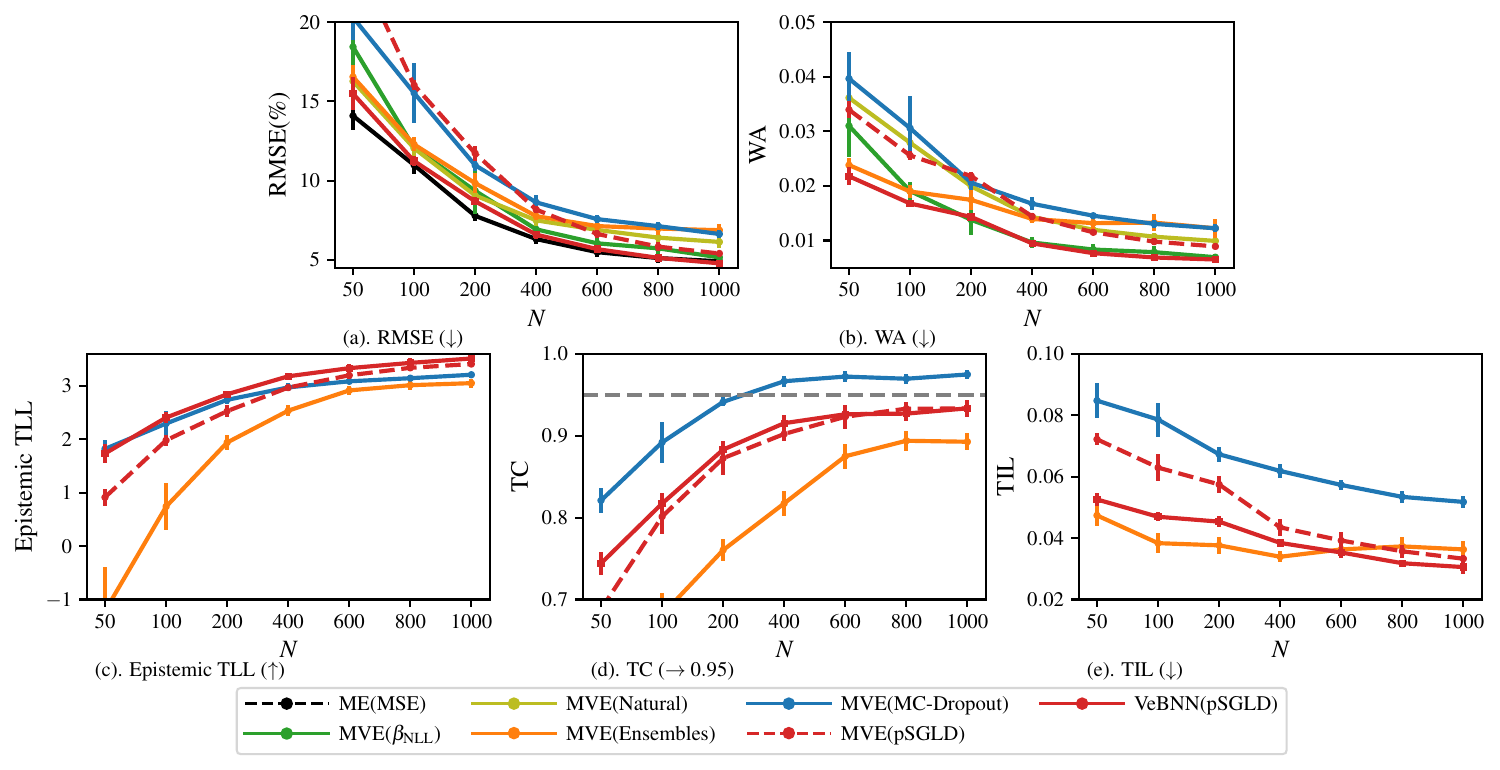}}
\caption{\textbf{Accuracy metrics obtained for the plasticity law discovery dataset considering a training set with different number $N$ of training sequences}. The Wasserstein distance (WA) represents the closeness of the estimated aleatoric uncertainty distribution to the ground truth distribution. All metrics are computed by repeating the training of each method 5 times, randomly resampling points from the training datasets. }
\label{fig:sve_1_case_statistics}

\end{figure*}

\paragraph{Baselines.} This dataset involves history-dependent data, therefore we use a GRU architecture (sequence-to-sequence regression). The results are shown in \cref{fig:sve_1_case_statistics}.\footnote{Note that some lines are absent in some figures because not every method is capable of simultaneously predicting mean, aleatoric, and epistemic uncertainties.} Note that results using BBB inference are not available due to a lack of convergence. 

\begin{figure*}[hbt!]
\centering
\centerline{\includegraphics[width=0.80\textwidth]{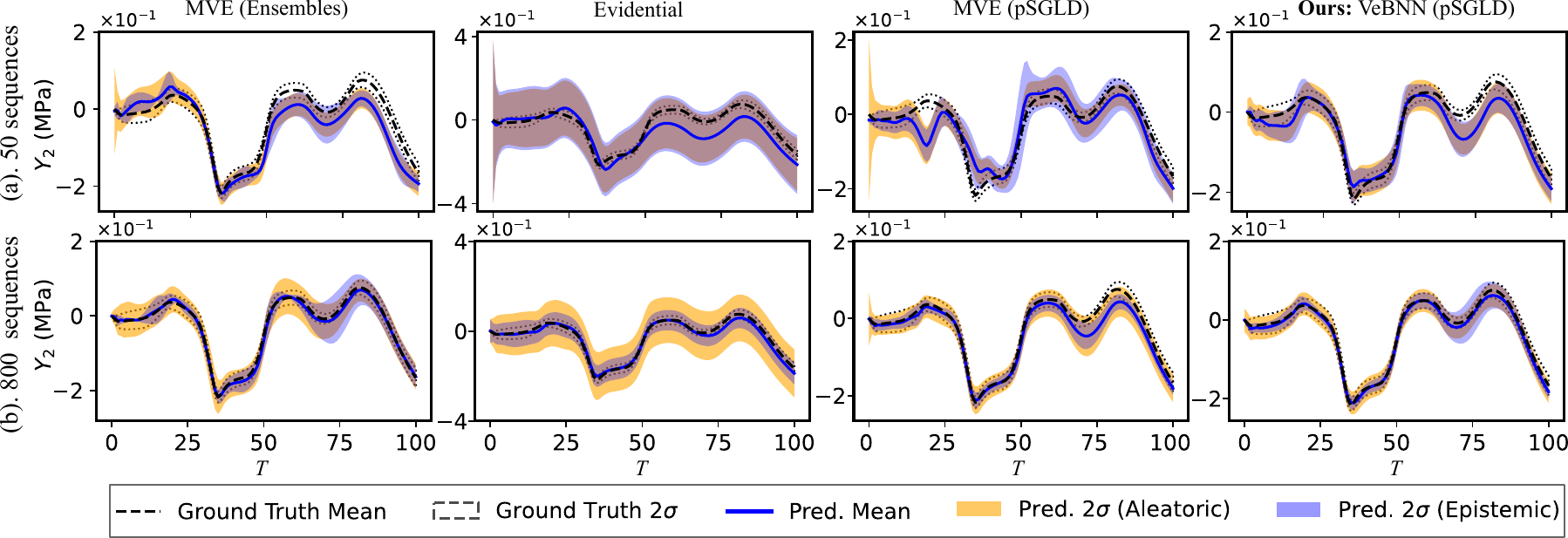}}
\caption{\textbf{Predictions of different methods on plasticity law discovery dataset.} We randomly pick one test point from the dataset and show the entire third component $\mathbf{y}_{2}$ of 100-time steps when considering 50 and 800 training sequences, respectively.}
\vskip -0.10in
\label{fig:plasticity_prediction}
\end{figure*}

\cref{fig:sve_1_case_statistics} shows the accuracy metrics obtained with different training sequences taken from the training dataset where $N$ increases from $50$ to $1000$. As expected, more training data improves accuracy for all methods, but it is clear that the proposed VeBNN (pSGLD) method achieves better predictions for all metrics, including the jointly trained MVE (pSGLD). Note that the proposed VeBNN (pSGLD) performs comparably well to the ME (MSE) when estimating the mean (RMSE metric), while reaching a similar Wasserstein distance to the MVE ($\beta\text{-NLL}=0.5$) network, the best method among jointly trained MVEs with MAP estimation, when estimating the aleatoric uncertainty. Importantly, the improvement of Epistemic TLL, together with the increase in TC, is also accompanied by a decrease in the epistemic TIL. In contrast, jointly trained MVE models with Bayesian inference methods such as MVE (Ensembles), MVE (MC-Dropout), and MVE (pSGLD) exhibit lower performance in both RMSE and WA. Their overestimation of aleatoric uncertainty causes them to inflate the epistemic uncertainty (larger TIL), and results in test coverage and TLL that only become comparable to the VeBNN when the number of samples is large ($N > 600$).

We include the results of Evidential learning in a separate figure (\Cref{fig:deep_evidential_plasticity}) due to its considerably lower performance, and we provide a detailed comparison between VeBNN (Ensembles) and VeBNN (MC-Dropout) with their jointly trained MVE counterparts in \Cref{sec:plasticity_additional_results}, where similar conclusions can be reached.

\cref{fig:plasticity_prediction} shows the predictions of the proposed VeBNN (pSGLD) and its correct identification of the disentangled data uncertainties, which improves as the training data increases (upper column to bottom column). As expected, the proposed VeBNN (pSGLD) outperforms the jointly trained MVE (pSGLD) for the same number of training sequences, and the estimated epistemic uncertainty decreases with increasing training data. Jointly trained MVE (Ensembles) perform well when considering larger training sets (800 training sequences), but performance deteriorates when compared with VeBNN (pSGLD) for 50 training sequences (this is also seen in \cref{fig:sve_1_case_statistics}). On the contrary, Evidential has significant difficulties with this problem. Predictions for other methods are given in \Cref{fig:additional_plot_for_plasiticity}, in which the proposed cooperative training strategy also considerably improves predictions when using MC-Dropout as inference. 

We also observe in \Cref{fig:sve_1_case_statistics} that the aleatoric uncertainty estimated by the VeBNN (pSGLD) converges for $N>400$, since  WA remains stable. Furthermore, the VeBNN correctly disentangles the uncertainties, as shown in \Cref{fig:plasticity_prediction}. We find these results very encouraging, and we hope our dataset will motivate future developments in the field.

\subsubsection{APPA-REAL Dataset}
We also adopt the ResNet-34 backbone for this problem, where the dataset is post-processed into images with a resolution of $128 \times 128$ pixels. The VeBNN (pSGLD) is compared with MVE ($\beta$-NLL) and MVE (pSGLD), where the accuracy metrics are summarized in \Cref{tab:appa_real_results}. VeBNN achieves the best performance across all accuracy metrics, though test coverage stays far from the $0.95$ target for all methods.

\begin{table}[h]
\centering
\caption{\textbf{Performance comparison on the APPA-REAL dataset.} 
The best result in each column is highlighted in bold.}
\label{tab:appa_real_results}
\resizebox{\columnwidth}{!}{
\begin{tabular}{lccccc}
\toprule
Method & RMSE ($\downarrow$) & WA ($\downarrow$) & Epistemic TLL ($\uparrow$) & TC $(\rightarrow 0.95)$ & TIL ($\downarrow$) \\
\midrule
MVE ($\beta$-NLL)
& \makecell{18.29\\(0.16)}
& \makecell{18.56\\(0.43)}
& /
& /
& / \\
MVE (pSGLD)
& \makecell{18.57\\(1.03)}
& \makecell{20.39\\(2.46)}
& \makecell{-22.89\\(26.52)}
& \makecell{0.50\\(0.30)}
& \makecell{36.44\\(42.59)} \\
VeBNN (pSGLD)
& \makecell{\textbf{17.72}\\(0.15)}
& \makecell{\textbf{13.50}\\(0.45)}
& \makecell{\textbf{-8.80}\\(7.19)}
& \makecell{\textbf{0.55}\\(0.09)}
& \makecell{\textbf{22.68}\\(7.19)} \\
\bottomrule
\end{tabular}
}
\end{table}

\section{Discussions}
\label{sec:discussion}

\begin{figure*}[t]
    \centering
    \includegraphics[width=0.75\textwidth]{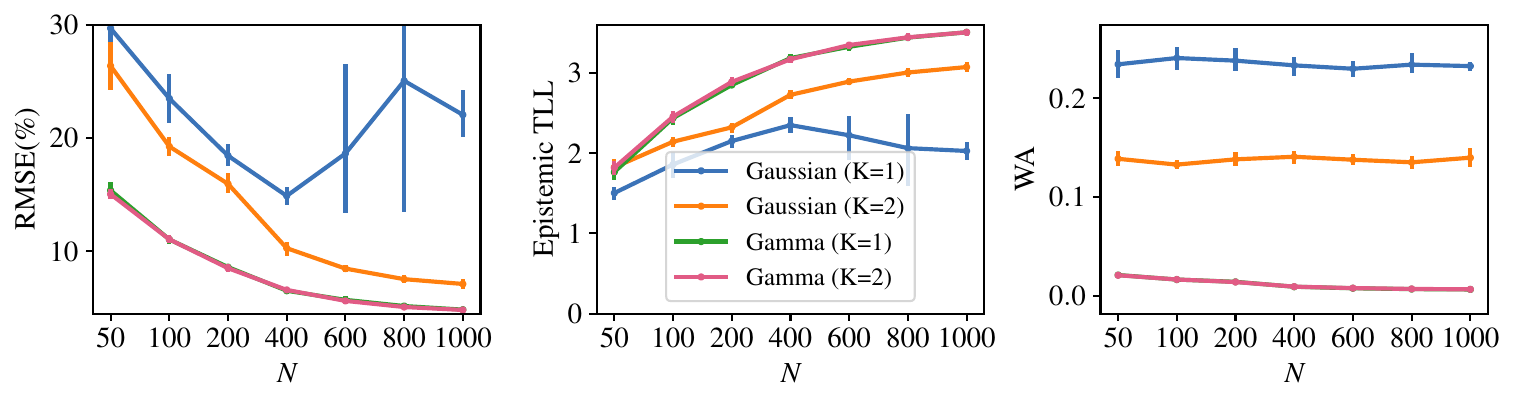}
    \caption{\textbf{Ablation study for the Gamma loss and iteration parameter $K$ using VeBNN (pSGLD).} 
    Curves labeled \emph{Gamma} and \emph{Gaussian} correspond to Step 2 training with the proposed Gamma loss (\Cref{eq:gamma_loss}) and the original Gaussian NLL loss (\Cref{eq:Lvar}), respectively, while fixing $\mu(\mathbf{x};\boldsymbol{\theta})$ from Step 1.}
    \label{fig:ablation_on_gamma_loss}
\end{figure*}

\paragraph{Ablation study for Gamma Loss and iterations $K$.} An ablation study is conducted based on VeBNN (pSGLD) for the material plasticity law discovery dataset because the ground-truth aleatoric uncertainty is known. The results with three performance metrics (RMSE, Epistemic TLL, and WA) are presented in \Cref{fig:ablation_on_gamma_loss}, and the remaining two are presented in \cref{fig:Ablation_Study_Gamma_K_Additional_Metrics}. Finding $\sigma_a^2(\mathbf{x};\bm{\phi})$ with a Gaussian likelihood loss in Step 2 leads to worst performance due to the high-order term in the gradient calculation (\Cref{eq:muder_Lvar}) that can lead to $\sigma_a^2(\mathbf{x}) \to \infty$ and a degenerate local optimum for which the gradient tends to zero (\Cref{eq:muder_Lvar}). In contrast, the proposed Gamma NLL loss addresses this and yields a stable optimization process (\Cref{fig:ablation_on_gamma_loss}).

We also find that the proposed training strategy significantly improves upon the state-of-the-art even when training the networks only once (i.e., $K=1$). Interestingly, we noticed that convergence was achieved for every dataset with only one additional iteration ($K=2$). To further explore this, \Cref{fig:trajectory_of_iteration} and \cref{fig:UCI_validation_loss} show the convergence of VeBNN after $K=5$ for both plasticity law discovery datasets and the \emph{Energy} problem of UCI regression datasets, respectively. These results confirm that $K$ is merely the iteration count until convergence rather than a tunable hyperparameter.

\paragraph{ Ablation study for Gamma vs. $\beta$-NLL vs. Natural Reparameterization losses of Step 2} We conduct an ablation study for the material plasticity law discovery dataset with VeBNN (pSGLD) and compare Step 2 trained with: 
(1) $\beta$-NLL ($\beta = 0.5$); 
(2) Natural re-parameterization; 
(3) Gamma likelihood. The results across varying training sizes ($N = 50$ to $1000$) are very conclusive, as shown in \cref{tab:gamma_results}.

\begin{table}[h]
\centering
\caption{\textbf{Performance comparison under different training sample sizes.}
The best result in each column is highlighted in bold.}
\label{tab:gamma_results}
\scriptsize
\setlength{\tabcolsep}{3pt}
\begin{tabular}{llcccccc}
\toprule
Metric & Method & 50 & 100 & 400 & 600 & 800 & 1000 \\
\midrule
\multirow{3}{*}{RMSE ($\downarrow$)}
& $\beta$-NLL 
& \makecell{24.67\\(1.04)} 
& \makecell{17.47\\(0.44)} 
& \makecell{8.99\\(0.15)} 
& \makecell{7.23\\(0.11)} 
& \makecell{6.35\\(0.06)} 
& \makecell{5.76\\(0.11)} \\
& Natural 
& \makecell{16.26\\(0.26)} 
& \makecell{12.04\\(0.55)} 
& \makecell{7.49\\(0.25)} 
& \makecell{6.89\\(0.15)} 
& \makecell{6.40\\(0.18)} 
& \makecell{6.13\\(0.26)} \\
& \textbf{Gamma} 
& \makecell{\textbf{15.05}\\\textbf{(0.26)}} 
& \makecell{\textbf{11.06}\\\textbf{(0.24)}} 
& \makecell{\textbf{6.59}\\\textbf{(0.12)}} 
& \makecell{\textbf{5.62}\\\textbf{(0.15)}} 
& \makecell{\textbf{5.08}\\\textbf{(0.13)}} 
& \makecell{\textbf{4.82}\\\textbf{(0.06)}} \\
\midrule
\multirow{3}{*}{WA ($\downarrow$)}
& $\beta$-NLL 
& \makecell{0.0426\\(0.0015)} 
& \makecell{0.0317\\(0.0005)} 
& \makecell{0.0173\\(0.0003)} 
& \makecell{0.0136\\(0.0002)} 
& \makecell{0.0117\\(0.0001)} 
& \makecell{0.0104\\(0.0002)} \\
& Natural 
& \makecell{0.0361\\(0.0025)} 
& \makecell{0.0279\\(0.0003)} 
& \makecell{0.0143\\(0.0003)} 
& \makecell{0.0120\\(0.0003)} 
& \makecell{0.0107\\(0.0002)} 
& \makecell{0.0099\\(0.0002)} \\
& \textbf{Gamma} 
& \makecell{\textbf{0.0210}\\\textbf{(0.0005)}} 
& \makecell{\textbf{0.0167}\\\textbf{(0.0005)}} 
& \makecell{\textbf{0.0094}\\\textbf{(0.0002)}} 
& \makecell{\textbf{0.0080}\\\textbf{(0.0002)}} 
& \makecell{\textbf{0.0071}\\\textbf{(0.0002)}} 
& \makecell{\textbf{0.0067}\\\textbf{(0.0002)}} \\
\bottomrule
\end{tabular}
\end{table}

The Gamma likelihood achieves the best RMSE and consistently lower WA across all sample sizes, indicating improved mean estimation and aleatoric uncertainty estimation within the cooperative framework. Combined with the ablation study (\Cref{fig:ablation_on_gamma_loss}), a canonical example based on the illustrative example in \Cref{fig:beta_vs_cuq}, and a theoretical justification in \Cref{sec:theory_justification_steps1_and_2}, we are confident that the Gamma likelihood is clearly advantageous for variance estimation in the cooperative setting.

\paragraph{Size of variance estimation network} The variance estimation network used for estimating aleatoric uncertainty is trained separately from the BNN. However, we believe that this is an advantage, as the architecture required to learn aleatoric uncertainty separately is simpler than when training for everything at once. We considered 8 different variance network configurations and observed robust estimations of aleatoric uncertainty (\cref{fig:barplot_of_different_var_nets} in \Cref{sec:discussion_size_var_net}).

\paragraph{Computational cost} As shown in Experiments, our cooperative training converges quickly: the mean network is trained once, while the variance network and Bayesian inference are performed once or twice. The mean network quickly finds a posterior mode, facilitating Bayesian training later on. Taking pSGLD as an example of inference method, the number of training epochs is comparable to ME and MVE training, as shown in \Cref{sec:hyperparams_setting}. In contrast, considering homoscedastic noise and treating it as a hyperparameter typically requires many full training runs and performs worse. Joint MVE training with Bayesian inference has similar efficiency, but our method consistently achieves better accuracy. Since deterministic ME and VE training is inexpensive relative to BNN training, the additional cost of our strategy is negligible and benefits from the warm start of the mean. Specifically, the wall-clock time on the material plasticity discovery dataset with $N = 1000$ was measured on an NVIDIA GeForce RTX 4070 GPU. For VeBNN (pSGLD): Step~1 $\approx 82$s, Step~2 $\approx 48$s, and Step~3 $\approx 100$s. Thus, VeBNN (pSGLD) takes $230$s ($K = 1$) and $378$s ($K = 2$). Similarly, MVE ($\beta$-NLL) would cost $410$s with selecting a $\beta$ from $5$ values. MVE (pSGLD) would use only around $100$s.  Overall, VeBNN is computationally competitive with MVE ($\beta$-NLL) even when considering a simple hyperparameter selection strategy.

\begin{figure}[h]
    \centering
    \vspace{-0.2in}
    \includegraphics[width=0.8\linewidth]{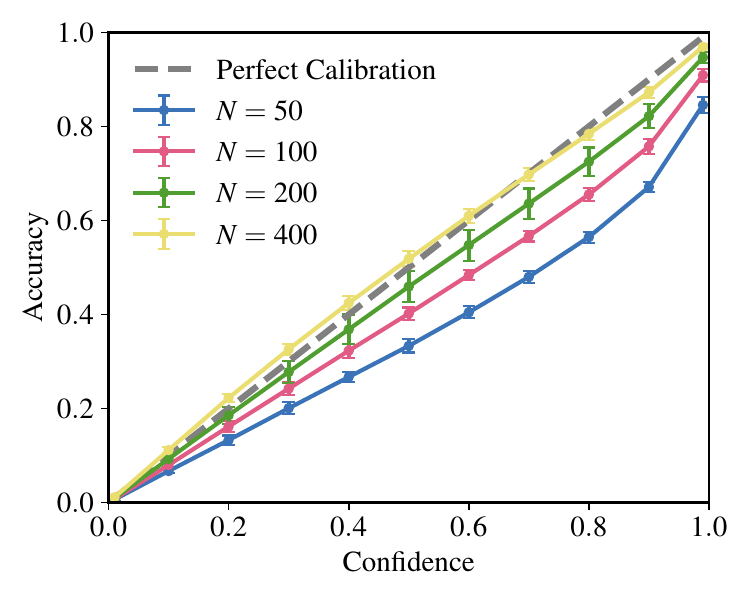}
    \caption{\textbf{Reliability diagrams for VeBNN (pSGLD) on the material plasticity dataset as the number of training sequences $N$ increases.} Calibration improves with more data, approaching the perfect-calibration line by $N=400$.}
    \label{fig:reliability_diagram}
\end{figure}

\paragraph{Epistemic uncertainty calibration} We also report the Expected Calibration Error (ECE) \cite{guo2017calibration} and reliability diagrams in \Cref{tab:epistemic_ece,fig:reliability_diagram}, respectively, to further assess the quality of the predicted epistemic uncertainty under a set of confidence levels ($\alpha=0.1, 0.2, ..., 1.0$).

\begin{table}[h]
\centering
\caption{\textbf{Epistemic calibration error (ECE $\downarrow$) for different training sample sizes.} The best results highlighted in bold.}
\label{tab:epistemic_ece}
\resizebox{\columnwidth}{!}{
\begin{tabular}{lccccccc}
\toprule
Method & 50 & 100 & 200 & 400 & 600 & 800 & 1000 \\
\midrule
MVE (pSGLD)
& \makecell{0.1760\\(0.0079)}
& \makecell{0.1160\\(0.0118)}
& \makecell{0.0577\\(0.0149)}
& \makecell{0.0283\\(0.0082)}
& \makecell{0.0250\\(0.0025)}
& \makecell{0.0241\\(0.0023)}
& \makecell{0.0223\\(0.0070)} \\
VeBNN (pSGLD)
& \makecell{\textbf{0.1352}\\(0.0042)}
& \makecell{\textbf{0.0778}\\(0.0036)}
& \makecell{\textbf{0.0351}\\(0.0074)}
& \makecell{\textbf{0.0261}\\(0.0027)}
& \makecell{\textbf{0.0233}\\(0.0026)}
& \makecell{\textbf{0.0230}\\(0.0035)}
& \makecell{\textbf{0.0201}\\(0.0030)} \\
\bottomrule
\end{tabular}
}
\end{table}

As shown in \Cref{tab:epistemic_ece}, VeBNN (pSGLD) achieves lower ECE than MVE (pSGLD) across all training sizes, indicating better epistemic uncertainty estimation. The reliability diagram in \Cref{fig:reliability_diagram} further confirms this trend: as the number of training samples increases, the predicted epistemic uncertainty becomes increasingly well-calibrated. When $N=400$, the reliability diagram nearly overlaps with the perfect calibration line. These results provide additional evidence that the VeBNN improves epistemic uncertainty estimation.

\paragraph{Extension to multi-modal data noise.} In real-world problems, aleatoric noise may be multi-modal. In these cases, unimodal Gaussian NLL loss (\cref{eq:Lvar}) is insufficient. To address this, VeBNN can be seamlessly extended using Mixture Density Networks (MDNs) \cite{Bishop1994} by replacing the unimodal mean network with an MDN-based mean network and adapting the variance network accordingly. This allows VeBNN to capture a broader class of predictive distributions, as illustrated in \Cref{fig:multi_modal_extension}.

\paragraph{Limitations}\label{sec:discussion_limitations} Although the proposed method facilitates Bayesian inference for BNNs, training BNNs remains more challenging than training deterministic networks. When the posterior is non-smooth or only partially known, inference of epistemic uncertainty under fixed Gaussian prior–likelihood pairs can become unreliable. Recent advances in Imprecise Probabilistic Machine Learning (IPML), such as Credal Bayesian Deep Learning (CBDL) \cite{caprio2023credal, wang2024credal}, provide broader epistemic coverage but at significantly higher computational cost. Future extensions may consider this and other techniques. Our focus was to show that Bayesian inference is easier when aleatoric uncertainty is determined separately.

\section{Conclusion}
\label{sec:conclusion}

We propose a novel Bayesian heteroscedastic regression strategy based on cooperative training of variance estimation and Bayesian neural networks that disentangles epistemic and aleatoric uncertainties. The proposed method is efficient, robust, and straightforward to implement because it is simpler to train each network in isolation, while ensuring complementarity in their training. We believe the method is scalable and applicable to real-world problems that will involve active learning and Bayesian optimization, considering data-scarce and data-rich scenarios, unlike Gaussian process regression.

\newpage

\section*{Acknowledgments}
Jiaxiang Yi would like to acknowledge the generous support of the China Scholarship Council (CSC) under Grant No.~CSC202106160006. Miguel Bessa acknowledges the generous support by the Department of the Navy, Office of Naval Research, awards number N00014-21-1-2670 and N00014-23-1-2688.

\section*{Impact Statement}

The paper proposes a generic method for uncertainty disentanglement, but uncertainty estimates also have error. Practitioners should validate uncertainty estimates on domain-specific data before deploying in safety-critical applications.

\bibliography{reference}

@Unpublished{Bishop1994,
  author    = {Christopher M. Bishop},
  title     = {Mixture density networks},
  year      = {1994},
  address   = {Birmingham},
  publisher = {Aston University},
  type      = {Technical Report},
}

@article{cui2020calibrated,
  title={Calibrated reliable regression using maximum mean discrepancy},
  author={Cui, Peng and Hu, Wenbo and Zhu, Jun},
  journal={Advances in Neural Information Processing Systems},
  volume={33},
  pages={17164--17175},
  year={2020}
}

@article{mohamed2016learning,
  title={Learning in implicit generative models},
  author={Mohamed, Shakir and Lakshminarayanan, Balaji},
  journal={arXiv preprint arXiv:1610.03483},
  year={2016}
}

@article{sluijterman2024optimal,
  title={Optimal training of mean variance estimation neural networks},
  author={Sluijterman, Laurens and Cator, Eric and Heskes, Tom},
  journal={Neurocomputing},
  pages={127929},
  year={2024},
  publisher={Elsevier}
}

@inproceedings{nix1994estimating,
  title={Estimating the mean and variance of the target probability distribution},
  author={Nix, David A and Weigend, Andreas S},
  booktitle={Proceedings of 1994 ieee international conference on neural networks (ICNN'94)},
  volume={1},
  pages={55--60},
  year={1994},
  organization={IEEE}
}

@inproceedings{depeweg2018decomposition,
  title={Decomposition of uncertainty in Bayesian deep learning for efficient and risk-sensitive learning},
  author={Depeweg, Stefan and Hernandez-Lobato, Jose-Miguel and Doshi-Velez, Finale and Udluft, Steffen},
  booktitle={International conference on machine learning},
  pages={1184--1193},
  year={2018},
  organization={PMLR}
}

@inproceedings{guo2017calibration,
  title={On calibration of modern neural networks},
  author={Guo, Chuan and Pleiss, Geoff and Sun, Yu and Weinberger, Kilian Q},
  booktitle={International conference on machine learning},
  pages={1321--1330},
  year={2017},
  organization={PMLR}
}

@article{hullermeier2021aleatoric,
  title={Aleatoric and epistemic uncertainty in machine learning: An introduction to concepts and methods},
  author={H{\"u}llermeier, Eyke and Waegeman, Willem},
  journal={Machine learning},
  volume={110},
  number={3},
  pages={457--506},
  year={2021},
  publisher={Springer}
}

@InProceedings{Skafte2019,
  author = {Detlefsen, Nicki S. and J\o{}rgensen, Martin and Hauberg, S\o{}ren},
title = {Reliable training and estimation of variance networks},
year = {2019},
publisher = {Curran Associates Inc.},
address = {Red Hook, NY, USA},
booktitle = {Proceedings of the 33rd International Conference on Neural Information Processing Systems},
articleno = {568},
numpages = {11}
}

@Misc{Seitzer2022,
  author        = {Maximilian Seitzer and Arash Tavakoli and Dimitrije Antic and Georg Martius},
  title         = {On the Pitfalls of Heteroscedastic Uncertainty Estimation with Probabilistic Neural Networks},
  year          = {2022},
  archiveprefix = {arXiv},
  eprint        = {2203.09168},
  primaryclass  = {cs.LG},
}

@InProceedings{ValdenegroToro2022,
  author = { Valdenegro-Toro, Matias and Mori, Daniel Saromo },
booktitle = { 2022 IEEE/CVF Conference on Computer Vision and Pattern Recognition Workshops (CVPRW) },
title = {{ A Deeper Look into Aleatoric and Epistemic Uncertainty Disentanglement }},
year = {2022},
volume = {},
ISSN = {},
pages = {1508-1516},
doi = {10.1109/CVPRW56347.2022.00157},
url = {https://doi.ieeecomputersociety.org/10.1109/CVPRW56347.2022.00157},
publisher = {IEEE Computer Society},
address = {Los Alamitos, CA, USA},
month =Jun}

@Misc{Stirn2020,
  author        = {Andrew Stirn and David A. Knowles},
  title         = {Variational Variance: Simple, Reliable, Calibrated Heteroscedastic Noise Variance Parameterization},
  year          = {2020},
  archiveprefix = {arXiv},
  eprint        = {2006.04910},
  primaryclass  = {cs.LG},
}

@inproceedings{meyer2020learning,
  title={Learning an uncertainty-aware object detector for autonomous driving},
  author={Meyer, Gregory P and Thakurdesai, Niranjan},
  booktitle={2020 IEEE/RSJ International Conference on Intelligent Robots and Systems (IROS)},
  pages={10521--10527},
  year={2020},
  organization={IEEE}
}

@InProceedings{Amini2020,
  author    = {Amini, Alexander and Schwarting, Wilko and Soleimany, Ava and Rus, Daniela},
  booktitle = {Advances in Neural Information Processing Systems},
  title     = {Deep Evidential Regression},
  year      = {2020},
  editor    = {H. Larochelle and M. Ranzato and R. Hadsell and M.F. Balcan and H. Lin},
  pages     = {14927--14937},
  publisher = {Curran Associates, Inc.},
  volume    = {33},
}

@InProceedings{Immer2023,
  author    = {Alexander Immer and Emanuele Palumbo and Alexander Marx and Julia E Vogt},
  booktitle = {Thirty-seventh Conference on Neural Information Processing Systems},
  title     = {Effective Bayesian Heteroscedastic Regression with Deep Neural Networks},
  year      = {2023},

}

@Book{Rasmussen2005,
  author    = {Rasmussen, Carl Edward and Williams, Christopher K. I.},
  publisher = {The MIT Press},
  title     = {{Gaussian Processes for Machine Learning}},
  year      = {2005},
  isbn      = {9780262256834},
  month     = {11},
  doi       = {10.7551/mitpress/3206.001.0001},
}

@Book{Neal1995,
  author    = {Neal, Radford M},
  publisher = {Springer Science \& Business Media},
  title     = {Bayesian learning for neural networks},
  year      = {1995},
  volume    = {118},
}

@InProceedings{Welling2011,
  author    = {Welling, Max and Teh, Yee W},
  booktitle = {Proceedings of the 28th international conference on machine learning (ICML-11)},
  title     = {Bayesian learning via stochastic gradient Langevin dynamics},
  year      = {2011},
  pages     = {681--688},
}

@inproceedings{mucsanyi2024benchmarking,
title={Benchmarking Uncertainty Disentanglement: Specialized Uncertainties for Specialized Tasks},
author = {Mucs\'{a}nyi, B\'{a}lint and Kirchhof, Michael and Oh, Seong Joon},
booktitle = {Advances in Neural Information Processing Systems},
doi = {10.52202/079017-1614},
editor = {A. Globerson and L. Mackey and D. Belgrave and A. Fan and U. Paquet and J. Tomczak and C. Zhang},
pages = {50972--51038},
publisher = {Curran Associates, Inc.},
title = {Benchmarking Uncertainty Disentanglement: Specialized Uncertainties for Specialized Tasks},
volume = {37},
year = {2024}
}

@inproceedings{li2016preconditioned,
  title={Preconditioned stochastic gradient Langevin dynamics for deep neural networks},
  author={Li, Chunyuan and Chen, Changyou and Carlson, David and Carin, Lawrence},
  booktitle={Proceedings of the AAAI conference on artificial intelligence},
  volume={30},
  number={1},
  year={2016}
}

@InProceedings{Graves2011,
  author    = {Graves, Alex},
  booktitle = {Advances in Neural Information Processing Systems},
  title     = {Practical Variational Inference for Neural Networks},
  year      = {2011},
  editor    = {J. Shawe-Taylor and R. Zemel and P. Bartlett and F. Pereira and K.Q. Weinberger},
  publisher = {Curran Associates, Inc.},
  volume    = {24},
  url       = {https://proceedings.neurips.cc/paper/2011/file/7eb3c8be3d411e8ebfab08eba5f49632-Paper.pdf},
}

@Misc{Blundell2015,
  author    = {Blundell, Charles and Cornebise, Julien and Kavukcuoglu, Koray and Wierstra, Daan},
  title     = {Weight Uncertainty in Neural Networks},
  year      = {2015},
  copyright = {arXiv.org perpetual, non-exclusive license},
  doi       = {10.48550/ARXIV.1505.05424},
  publisher = {arXiv},
  url       = {https://arxiv.org/abs/1505.05424},
}

@InProceedings{Gal2016,
  author    = {Gal, Yarin and Ghahramani, Zoubin},
  booktitle = {Proceedings of The 33rd International Conference on Machine Learning},
  title     = {Dropout as a Bayesian Approximation: Representing Model Uncertainty in Deep Learning},
  year      = {2016},
  address   = {New York, New York, USA},
  editor    = {Balcan, Maria Florina and Weinberger, Kilian Q.},
  month     = {20--22 Jun},
  pages     = {1050--1059},
  publisher = {PMLR},
  series    = {Proceedings of Machine Learning Research},
  volume    = {48},

}

@InProceedings{Lakshminarayanan2017,
  author    = {Lakshminarayanan, Balaji and Pritzel, Alexander and Blundell, Charles},
  booktitle = {Advances in Neural Information Processing Systems},
  title     = {Simple and Scalable Predictive Uncertainty Estimation using Deep Ensembles},
  year      = {2017},
  editor    = {I. Guyon and U. Von Luxburg and S. Bengio and H. Wallach and R. Fergus and S. Vishwanathan and R. Garnett},
  publisher = {Curran Associates, Inc.},
  volume    = {30},
}

@InProceedings{Kendall2017,
  author    = {Kendall, Alex and Gal, Yarin},
  booktitle = {Advances in Neural Information Processing Systems},
  title     = {What Uncertainties Do We Need in Bayesian Deep Learning for Computer Vision?},
  year      = {2017},
  editor    = {I. Guyon and U. Von Luxburg and S. Bengio and H. Wallach and R. Fergus and S. Vishwanathan and R. Garnett},
  publisher = {Curran Associates, Inc.},
  volume    = {30},
}

@Article{Harakeh2023,
  author       = {Harakeh, Ali and Hu, Jordan Sir Kwang and Guan, Naiqing and Waslander, Steven and Paull, Liam},
  journal      = {Proceedings of the AAAI Conference on Artificial Intelligence},
  title        = {Estimating Regression Predictive Distributions with Sample Networks},
  year         = {2023},
  month        = {Jun.},
  number       = {6},
  pages        = {7830-7838},
  volume       = {37},
  doi          = {10.1609/aaai.v37i6.25948},
}

@Misc{izmailov2021bayesian,
  author        = {Pavel Izmailov and Sharad Vikram and Matthew D. Hoffman and Andrew Gordon Wilson},
  title         = {What Are Bayesian Neural Network Posteriors Really Like?},
  year          = {2021},
  archiveprefix = {arXiv},
  eprint        = {2104.14421},
  primaryclass  = {cs.LG},
}

@Misc{Wenzel2020,
  author    = {Wenzel, Florian and Roth, Kevin and Veeling, Bastiaan S. and Świątkowski, Jakub and Tran, Linh and Mandt, Stephan and Snoek, Jasper and Salimans, Tim and Jenatton, Rodolphe and Nowozin, Sebastian},
  title     = {How Good is the Bayes Posterior in Deep Neural Networks Really?},
  year      = {2020},
  copyright = {arXiv.org perpetual, non-exclusive license},
  doi       = {10.48550/ARXIV.2002.02405},
  publisher = {arXiv},
  url       = {https://arxiv.org/abs/2002.02405},
}

@InProceedings{Wilson2020,
  author    = {Wilson, Andrew G and Izmailov, Pavel},
  booktitle = {Advances in Neural Information Processing Systems},
  title     = {Bayesian Deep Learning and a Probabilistic Perspective of Generalization},
  year      = {2020},
  editor    = {H. Larochelle and M. Ranzato and R. Hadsell and M.F. Balcan and H. Lin},
  pages     = {4697--4708},
  publisher = {Curran Associates, Inc.},
  volume    = {33},
  url       = {https://proceedings.neurips.cc/paper/2020/file/322f62469c5e3c7dc3e58f5a4d1ea399-Paper.pdf},
}

@Article{Abdar2021,
  author   = {Moloud Abdar and Farhad Pourpanah and Sadiq Hussain and Dana Rezazadegan and Li Liu and Mohammad Ghavamzadeh and Paul Fieguth and Xiaochun Cao and Abbas Khosravi and U. Rajendra Acharya and Vladimir Makarenkov and Saeid Nahavandi},
  journal  = {Information Fusion},
  title    = {A review of uncertainty quantification in deep learning: Techniques, applications and challenges},
  year     = {2021},
  issn     = {1566-2535},
  pages    = {243-297},
  volume   = {76},
  doi      = {https://doi.org/10.1016/j.inffus.2021.05.008},
}

@inproceedings{Kingma2014,
  author       = {Diederik P. Kingma and
                  Jimmy Ba},
  editor       = {Yoshua Bengio and
                  Yann LeCun},
  title        = {Adam: {A} Method for Stochastic Optimization},
  booktitle    = {3rd International Conference on Learning Representations, {ICLR} 2015,
                  San Diego, CA, USA, May 7-9, 2015, Conference Track Proceedings},
  year         = {2015},
  url          = {http://arxiv.org/abs/1412.6980},
  bibsource    = {dblp computer science bibliography, https://dblp.org}
}

@inproceedings{HernandezLobato2015,
author = {Hern\'{a}ndez-Lobato, Jos\'{e} Miguel and Adams, Ryan P.},
title = {Probabilistic backpropagation for scalable learning of Bayesian neural networks},
year = {2015},
publisher = {JMLR},
booktitle = {Proceedings of the 32nd International Conference on Machine Learning - Volume 37},
pages = {1861–1869},
numpages = {9},
location = {Lille, France},
series = {ICML'15}
}

@Book{Murphy2012,
  author    = {Murphy, Kevin P},
  publisher = {MIT press},
  title     = {Machine learning: a probabilistic perspective},
  year      = {2021},
}

@Article{Melro2008,
  author   = {A.R. Melro and P.P. Camanho and S.T. Pinho},
  journal  = {Composites Science and Technology},
  title    = {Generation of random distribution of fibres in long-fibre reinforced composites},
  year     = {2008},
  issn     = {0266-3538},
  number   = {9},
  pages    = {2092-2102},
  volume   = {68},
  doi      = {https://doi.org/10.1016/j.compscitech.2008.03.013},
}

@misc{ruder2017overviewgradientdescentoptimization,
      title={An overview of gradient descent optimization algorithms}, 
      author={Sebastian Ruder},
      year={2017},
      eprint={1609.04747},
      archivePrefix={arXiv},
      primaryClass={cs.LG},
      url={https://arxiv.org/abs/1609.04747}, 
}

@Article{Kantorovich1960,
  author   = {Kantorovich, L. V.},
  journal  = {Management Science},
  title    = {Mathematical Methods of Organizing and Planning Production},
  year     = {1960},
  number   = {4},
  pages    = {366-422},
  volume   = {6},
  doi      = {10.1287/mnsc.6.4.366},
  eprint   = { https://doi.org/10.1287/mnsc.6.4.366 },
  url      = {
        https://doi.org/10.1287/mnsc.6.4.366
},
}

@article{cho2014learning,
  title={Learning phrase representations using RNN encoder-decoder for statistical machine translation},
  author={Cho, Kyunghyun},
  journal={arXiv preprint arXiv:1406.1078},
  year={2014}
}

@Misc{Gan2017,
  author        = {Zhe Gan and Chunyuan Li and Changyou Chen and Yunchen Pu and Qinliang Su and Lawrence Carin},
  title         = {Scalable Bayesian Learning of Recurrent Neural Networks for Language Modeling},
  year          = {2017},
  archiveprefix = {arXiv},
  eprint        = {1611.08034},
  primaryclass  = {cs.CL},
}

@manual{abaqus2024,
  title        = {Abaqus 2024},
  author       = {{Dassault Systèmes}},
  year         = 2024,
  note         = {Computer software},
  url          = {https://www.3ds.com/products/simulia/abaqus}
}

@Article{Dekhovich2023,
  author   = {Aleksandr Dekhovich and O. Taylan Turan and Jiaxiang Yi and Miguel A. Bessa},
  journal  = {Computer Methods in Applied Mechanics and Engineering},
  title    = {Cooperative data-driven modeling},
  year     = {2023},
  issn     = {0045-7825},
  pages    = {116432},
  volume   = {417},
  doi      = {https://doi.org/10.1016/j.cma.2023.116432},
}

@Article{Mozaffar2019,
  author    = {Mozaffar, M. and Bostanabad, R. and Chen, W. and Ehmann, K. and Cao, J. and Bessa, M. A.},
  journal   = {Proceedings of the National Academy of Sciences},
  title     = {Deep learning predicts path-dependent plasticity},
  year      = {2019},
  issn      = {0027-8424},
  number    = {52},
  pages     = {26414--26420},
  volume    = {116},
  doi       = {10.1073/pnas.1911815116},
  eprint    = {https://www.pnas.org/content/116/52/26414.full.pdf},
  publisher = {National Academy of Sciences},
}

@inproceedings{yi2023rvesimulator,
  title={rvesimulator: An automated representative volume element simulator for data-driven material discovery},
  author={Yi, Jiaxiang and Bessa, Miguel Anibal},
  booktitle={AI for Accelerated Materials Design-NeurIPS 2023 Workshop},
  year={2023}
}

@misc{gustafsson2023reliableregressionmodelsuncertainty,
      title={How Reliable is Your Regression Model's Uncertainty Under Real-World Distribution Shifts?}, 
      author={Fredrik K. Gustafsson and Martin Danelljan and Thomas B. Schön},
      year={2023},
      eprint={2302.03679},
      archivePrefix={arXiv},
      primaryClass={cs.LG},
      url={https://arxiv.org/abs/2302.03679}, 
}

@InProceedings{He_2016_CVPR,
author = {He, Kaiming and Zhang, Xiangyu and Ren, Shaoqing and Sun, Jian},
title = {Deep Residual Learning for Image Recognition},
booktitle = {Proceedings of the IEEE Conference on Computer Vision and Pattern Recognition (CVPR)},
month = {June},
year = {2016}
}

@misc{fort2020deepensembleslosslandscape,
      title={Deep Ensembles: A Loss Landscape Perspective}, 
      author={Stanislav Fort and Huiyi Hu and Balaji Lakshminarayanan},
      year={2020},
      eprint={1912.02757},
      archivePrefix={arXiv},
      primaryClass={stat.ML},
      url={https://arxiv.org/abs/1912.02757}, 
}

@misc{smith2024rethinkingaleatoricepistemicuncertainty,
      title={Rethinking Aleatoric and Epistemic Uncertainty}, 
      author={Freddie Bickford Smith and Jannik Kossen and Eleanor Trollope and Mark van der Wilk and Adam Foster and Tom Rainforth},
      year={2024},
      eprint={2412.20892},
      archivePrefix={arXiv},
      primaryClass={cs.LG},
      url={https://arxiv.org/abs/2412.20892}, 
}

@inproceedings{
wang2025credal,
title={Credal Wrapper of Model Averaging for Uncertainty Estimation in Classification},
author={Kaizheng Wang and Fabio Cuzzolin and Keivan Shariatmadar and David Moens and Hans Hallez},
booktitle={The Thirteenth International Conference on Learning Representations},
year={2025},
url={https://openreview.net/forum?id=cv2iMNWCsh}
}

@Article{caprio2023credal,
	title={Credal Bayesian deep learning},
	author={Caprio, Michele and Dutta, Souradeep and Jang, Kuk Jin and Lin, Vivian and Ivanov, Radoslav and Sokolsky, Oleg and Lee, Insup},
	journal={arXiv preprint arXiv:2302.09656},
	year={2023}
}

@Article{wang2024credal,
	title={Credal deep ensembles for uncertainty quantification},
	author={Wang, Kaizheng and Cuzzolin, Fabio and Shariatmadar, Keivan and Moens, David and Hallez, Hans and others},
	journal={Advances in Neural Information Processing Systems},
	volume={37},
	pages={79540--79572},
	year={2024}
}

@misc{guo2017calibrationmodernneuralnetworks,
      title={On Calibration of Modern Neural Networks}, 
      author={Chuan Guo and Geoff Pleiss and Yu Sun and Kilian Q. Weinberger},
      year={2017},
      eprint={1706.04599},
      archivePrefix={arXiv},
      primaryClass={cs.LG},
      url={https://arxiv.org/abs/1706.04599}, 
}

@article{wilcoxon1945individual,
  title={Individual comparisons by ranking methods},
  author={Wilcoxon, Frank},
  journal={Biometrics Bulletin},
  volume={1},
  number={6},
  pages={80--83},
  year={1945},
  publisher={JSTOR}
}

@article{demvsar2006statistical,
  title={Statistical comparisons of classifiers over multiple data sets},
  author={Dem{\v{s}}ar, Janez},
  journal={Journal of Machine Learning Research},
  volume={7},
  pages={1--30},
  year={2006}
}

@inproceedings{agustsson2017appareal,
  author={Agustsson, Eirikur and Timofte, Radu and Escalera, Sergio and Baro, Xavier and Guyon, Isabelle and Rothe, Rasmus},
  booktitle={2017 12th IEEE International Conference on Automatic Face \& Gesture Recognition (FG 2017)}, 
  title={Apparent and Real Age Estimation in Still Images with Deep Residual Regressors on Appa-Real Database}, 
  year={2017},
  volume={},
  number={},
  pages={87-94},
  doi={10.1109/FG.2017.20}}

@inproceedings{
azizi2026clear,
title={{CLEAR}: Calibrated Learning for Epistemic and Aleatoric Risk},
author={Ilia Azizi and Juraj Bodik and Jakob Heiss and Bin Yu},
booktitle={The Fourteenth International Conference on Learning Representations},
year={2026},
url={https://openreview.net/forum?id=RY4IHaDLik}
}

@misc{fishkov2025,
      title={Uncertainty Quantification for Regression using Proper Scoring Rules}, 
      author={Alexander Fishkov and Kajetan Schweighofer and Mykyta Ielanskyi and Nikita Kotelevskii and Mohsen Guizani and Maxim Panov},
      year={2025},
      eprint={2509.26610},
      archivePrefix={arXiv},
      primaryClass={cs.LG},
      url={https://arxiv.org/abs/2509.26610}, 
}

@inproceedings{TicTac,
author = {Shukla, Megh and Salzmann, Mathieu and Alahi, Alexandre},
title = {{TIC-TAC}: A Framework for Improved Covariance Estimation in Deep Heteroscedastic Regression},
year = {2024},
publisher = {JMLR.org},
booktitle = {Proceedings of the 41st International Conference on Machine Learning},
articleno = {1841},
numpages = {14},
location = {Vienna, Austria},
series = {ICML'24},
url={https://arxiv.org/abs/2310.18953}
}
\bibliographystyle{icml2026}

\newpage
\appendix
\onecolumn
\section{Cooperative Training of Mean Estimation (ME) and Variance Estimation (VE) Networks (Step 1 and Step 2)}
\label{sec:joint_MVE_training_verse_cooperative_training}

Across all the datasets considered in this article, we consistently found the proposed cooperative strategy (Step 1 and Step 2) to lead to improved mean and aleatoric variance estimates. Due to space limitations, \cref{sec:aleatoric uncertainty} only provides a short explanation about the challenges involved in jointly training mean variance estimation (MVE) networks. Therefore, this appendix further elaborates on this and provides an illustrative example in \cref{sec:losses_comparison}, as well as a theoretical justification in \cref{sec:theory_justification_steps1_and_2} that supports our empirical findings.

\subsection{Canonical Example: Joint Training of MVE Network vs. Sequential Training of ME and VE Networks}
\label{sec:losses_comparison}

\begin{figure}[h]
\centering
\centerline{\includegraphics[width=0.9\columnwidth]{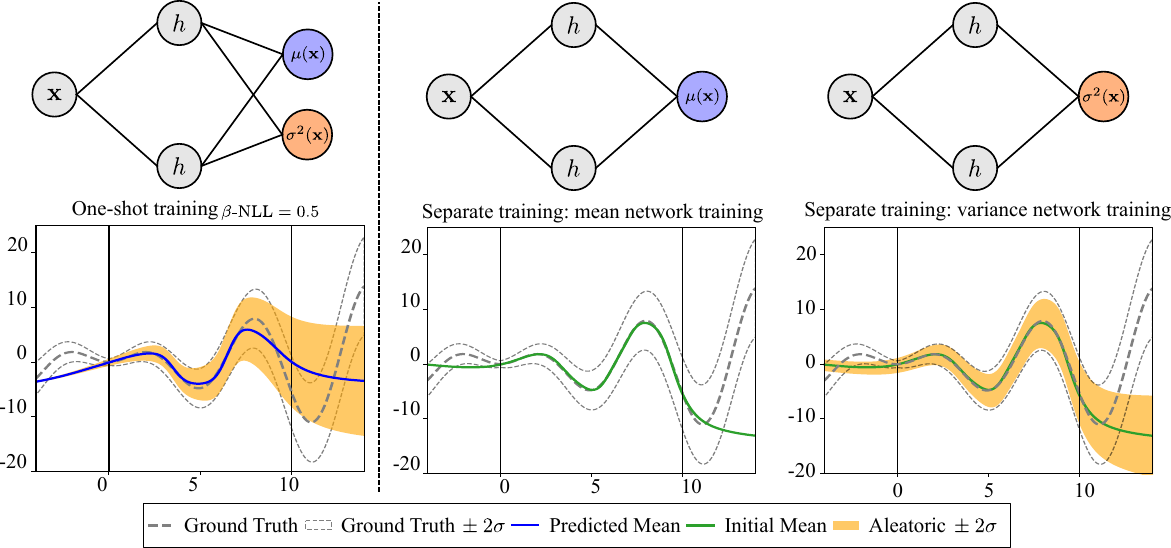}}
\caption{\textbf{Comparison between MVE network training (left figure) and separate training of ME network and variance estimation network (right figure).} The MVE network outputs $\mu(\mathbf{x})$ and $\sigma_a^2(\mathbf{x})$ and is trained by minimizing \cref{eq:Lvar}. The ME network is trained first using \cref{eq:Lvar} assuming constant aleatoric uncertainty, and then the VE network is trained using  \cref{eq:gamma_loss} and assuming a fixed mean obtained from the ME network.}
\label{fig:beta_vs_cuq}
\end{figure}

\cref{fig:beta_vs_cuq} contains the results of a canonical example for one-dimensional data. As discussed in the main text, we can jointly train mean variance estimation (MVE) networks that estimate both mean and variance simultaneously, or we can train two separate networks sequentially: an ME network and a VE network.

MVE achieves worse predictions (left in \cref{fig:beta_vs_cuq}) in the region ($x>7$) with higher aleatoric variance. As we elaborate in the next section, this occurs due to optimization challenges arising from the high-order variance term that appears in the denominator of the gradient of \cref{eq:Lvar}. In contrast, the figure in the middle shows the result of training only the mean network (i.e., assuming constant aleatoric uncertainty), which leads to a good mean estimate -- this is not surprising in light of the vast empirical evidence throughout the literature on the successes of training deterministic models that only provide a mean estimate. Furthermore, the right figure also illustrates that the subsequent training of the variance estimation network leads to an improved variance prediction that aligns well with the ground truth. This simplified example illustrates well the results obtained for all other datasets.

\subsection{Theoretical Justification of Challenges in MVE Network Training}
\label{sec:theory_justification_steps1_and_2}

We start by restating the heteroscedastic regression problem introduced in the main section:
\begin{equation} 
\label{eq:problem_set_up2}
     y = f(\mathbf{x}) + \epsilon(\mathbf{x})
\end{equation}
where $f(\mathbf{x})$ denotes the underlying noiseless ground truth function (expected mean), and $\epsilon(\mathbf{x})$ is the corresponding heteroscedastic Gaussian noise: $\epsilon(\mathbf{x})\sim \mathcal{N}(0, s^2(\mathbf{x}))$ where $s^2(\mathbf{x})$ represents its ground truth input-dependent aleatoric variance.

The goal in this article is to learn:
\begin{enumerate}
	\item The predictive mean: $\mu(\mathbf{x}) \approx f(\mathbf{x})$
	\item Aleatoric uncertainty: $\sigma_a^2(\mathbf{x}) \approx s^2(\mathbf{x})$
	\item Epistemic uncertainty: $\sigma_e^2(\mathbf{x})$ reflecting model uncertainty
\end{enumerate}

\subsubsection{Loss Functions of MVE, ME and VE Networks}

Training a Mean-Variance Estimation (MVE) network implies joint optimization of the NLL loss written in \cref{eq:Lvar}. We rewrite it here for completeness:
\begin{equation}
\label{eq:loss_MVE}
\mathcal{L}_{\text{MVE}}(\boldsymbol{\theta}) = \sum_{n=1}^N \left[\frac{(y_n - \mu(\mathbf{x}_n; \boldsymbol{\theta}))^2}{2 \sigma_a^2(\mathbf{x}_n;\boldsymbol{\phi})} + \frac{1}{2}\log\big(\sigma_a^2(\mathbf{x}_n;\boldsymbol{\phi})\big)\right]
\end{equation}
The typical MVE architecture \cite{nix1994estimating} shown on the left of \cref{fig:beta_vs_cuq} is a single network that outputs $\mu(\mathbf{x}; \boldsymbol{\theta}_{\text{joint}})$ and $\sigma_a^2(\mathbf{x}; \boldsymbol{\theta}_{\text{joint}})$, where a single set of parameters is shared $\boldsymbol{\theta} = \boldsymbol{\phi} \equiv \boldsymbol{\theta}_{\text{joint}}$.\footnote{Alternatively, two separate networks can also be trained \textbf{simultaneously}, where one network outputs $\mu(\mathbf{x}; \boldsymbol{\theta})$ and the other $\sigma_a^2(\mathbf{x}; \boldsymbol{\phi})$, in which case each network has its set of parameters). Another option is to have a single network with a common trunk (with shared parameters) and then two heads with independent parameters that output the mean and variance estimates. Nevertheless, these architectural variations are not relevant to our analysis because in all cases there is a coupled optimization problem that needs to be solved to find the mean and variance estimates simultaneously, as discussed in this appendix.}

As presented in the main text, the gradient of $\mathcal{L}_{\text{MVE}}$ with respect to the mean $\mu$ is
\begin{align}
\label{eq:MVE_grad_mean}
    \frac{\partial \mathcal{L}_{\text{MVE}}}{\partial \mu}= \sum_{n=1}^{N}\left( \frac{\mu(\mathbf{x}_n;\bm{\theta}) - y_n}{\sigma_a^2(\mathbf{x}_n;\bm{\phi})}\right)
\end{align}
and the gradient of $\mathcal{L}_{\text{MVE}}$ with respect to the variance $\sigma_a^2$ is
\begin{equation}
\label{eq:MVE_grad_var}
\frac{\partial \mathcal{L}_{\text{MVE}}}{\partial \sigma_a^2} = \sum_{n=1}^{N}\left( \frac{\sigma_a^2(\mathbf{x}_n;\bm{\phi}) - (y_n - \mu(\mathbf{x}_n;\bm{\theta}))^2}{2\big(\sigma_a^2(\mathbf{x}_n;\bm{\phi})\big)^2}\right)
\end{equation}
This creates a coupling where the gradient with respect to the variance depends on the mean's estimation error.

Contrary to joint training, the cooperative training strategy decouples the optimization problem. As presented in \cref{alg: CUQalgorithm}, training starts by considering a mean estimation (ME) network with constant variance. Therefore, the NLL loss for Step 1 is simplified because the variance does not contribute to the estimation of the mean.

\begin{itemize}
	\item \textbf{Step 1 (Mean network training)}: The point estimate of this network is given by (omitting the regularization term):
	\begin{equation}
    \label{eq:point_estimate_MEnet}
        \hat{\boldsymbol{\theta}} = \arg\min_{\boldsymbol{\theta}} \left[ \sum_{n=1}^{N} (y_n - \mu(\mathbf{x}_n; \boldsymbol{\theta}))^2 \right]
	\end{equation}
	
	\item \textbf{Step 2 (Variance network training)}: The VE network $\sigma_a^2(\mathbf{x}; \boldsymbol{\phi})$ is trained on the residuals $r_n = (y_n - \mu(\mathbf{x}_n; \hat{\boldsymbol{\theta}}))^2$ using the point estimate for the weights $\hat{\boldsymbol{\theta}}$ that was obtained from the ME network training in Step 1. If we use \cref{eq:loss_MVE}, i.e. a Gaussian NLL, then the point estimate for the VE network parameters is (omitting the regularization term):
	\begin{equation}
        \hat{\boldsymbol{\phi}} = \arg\min_{\boldsymbol{\phi}} \left\{ \sum_{n=1}^{N} \left[\frac{r_n}{2 \sigma_a^2(\mathbf{x}_n;\boldsymbol{\phi})} + \frac{1}{2}\log\big(\sigma_a^2(\mathbf{x}_n;\boldsymbol{\phi})\big)\right] \right\}
    \end{equation}
    Alternatively, as we propose in the main text, we can also use the Gamma NLL to obtain the variance estimate, as the Gamma loss does not have higher-order terms in its gradient. In that case, we minimize the loss given by \cref{eq:gamma_loss}, instead of \cref{eq:loss_MVE}, and estimate the variance by \cref{eq:aleatoric_variance}.
\end{itemize}

\subsubsection{Loss Function Convexity Analysis}

Training neural networks involves the minimization of loss functions that are non-convex with respect to the network parameters. However, we demonstrate that the \textbf{loss of MVE networks is non-convex with respect to the mean and variance outputs (non-convexity of the last layer)}. In contrast, the ME network and the VE network (trained separately) have convex losses with respect to their outputs (mean and variance, respectively).

\textbf{Loss convexity in mean estimation network (Step 1)}: Let the network output be $u$, estimating the mean $\mu$. Then, the NLL is given by the Mean Squared Error,
    \begin{equation}
        \mathcal{L}_{ME}(u) = (y-u)^2
    \end{equation}
and its Hessian with respect to $u$ is
    \begin{equation}
        \frac{d^2\mathcal{L}_{ME}}{du^2} = 2 > 0
    \end{equation}
concluding that $\mathcal{L}_{ME}(u)$ is strictly convex in $u$.

\textbf{Loss convexity in variance estimation network (Step 2)}: Let the network output be $v=\log\sigma^2$, and let the residual from the fixed mean estimate be $r = y - \hat\mu$. The NLL (with $\mu$ fixed) is
    \begin{equation}
        \mathcal{L}_{VE}(v) = \frac{r^2}{2}e^{-v} + \frac{1}{2}v
    \end{equation}
and its Hessian with respect to $v$ is
    \begin{equation}
        \frac{d^2\mathcal{L}_{VE}}{dv^2} = \frac{r^2}{2} e^{-v} \ge 0
    \end{equation}
concluding that $\mathcal{L}_{VE}(v)$ is convex in $v$ (strictly convex when $r\neq 0$).

\textbf{Loss non-convexity in MVE network}: The MVE network has a joint output loss,
    \begin{equation}
        \mathcal{L}_{\text{MVE}}(u,v) = \tfrac{1}{2}(y-u)^2 e^{-v} + \tfrac{1}{2}v
    \end{equation}
and its second derivatives with respect to $u$ and $v$ are
    \begin{equation}
        \frac{\partial^2 \mathcal{L}_{\text{MVE}}}{\partial u^2} = e^{-v},\qquad
    \frac{\partial^2 \mathcal{L}}{\partial v^2} = \tfrac{1}{2}(y-u)^2 e^{-v}
    \end{equation}
and the mixed derivative is
    \begin{equation}
        \frac{\partial^2 \mathcal{L}_{\text{MVE}}}{\partial u\partial v}
    = \frac{\partial}{\partial v}\big(-(y-u)e^{-v}\big) = (y-u)e^{-v}
    \end{equation}
Therefore, the Hessian is
    \begin{equation}
        H \;=\; \begin{bmatrix}
        	e^{-v} & (y-u)e^{-v}\\[4pt]
        	(y-u)e^{-v} & \tfrac{1}{2}(y-u)^2 e^{-v}
        \end{bmatrix}
    \end{equation}
Its determinant is
    \begin{equation}
        \det(H) = e^{-v}\cdot \tfrac{1}{2}(y-u)^2 e^{-v} - \big((y-u)e^{-v}\big)^2
    = -\tfrac{1}{2}(y-u)^2 e^{-2v}
    \end{equation}

Hence for any point with nonzero residual $(y-u)\neq 0$, $\det(H)<0$, so $H$ is indefinite (it has eigenvalues of opposite sign). Therefore, the joint loss is \emph{not} convex in $(u,v)$. The Hessian is only not indefinite when the exact solution $y=u$ occurs.

In fact, a stronger statement holds: the MVE output-space loss has \emph{no stationary point at all}.

\begin{lemma}[No finite critical point of the MVE loss]
\label{lem:mve_no_critical_point}
For a single data point, the MVE output-space loss $\mathcal{L}_{\mathrm{MVE}}(u,v)=\tfrac{1}{2}(y-u)^2 e^{-v}+\tfrac{1}{2}v$ has no finite critical point: there is no $(u,v)\in\mathbb{R}^2$ at which both partial derivatives vanish.
\end{lemma}
\begin{proof}
The stationarity conditions are
\begin{equation*}
\frac{\partial \mathcal{L}_{\mathrm{MVE}}}{\partial u} = -(y-u)\,e^{-v}=0,
\qquad
\frac{\partial \mathcal{L}_{\mathrm{MVE}}}{\partial v} = -\tfrac{1}{2}(y-u)^2 e^{-v}+\tfrac{1}{2}=0 .
\end{equation*}
Since $e^{-v}>0$, the first condition forces $u=y$, hence $(y-u)^2=0$; substituting into the second yields $\tfrac{1}{2}=0$, a contradiction. The two conditions are therefore jointly inconsistent, so no finite stationary point exists.
\end{proof}

We conclude that training an MVE network is unstable in a strong sense: the output-space loss is everywhere non-convex (indefinite Hessian) and, by \cref{lem:mve_no_critical_point}, has no stationary point to converge to. Gradient-based optimization therefore cannot settle at a critical point and instead drifts toward the degenerate limit $u\to y$, $v\to-\infty$ (an overfit mean with a vanishing predicted variance). This explains the empirical observations reported in the literature \cite{Skafte2019, Seitzer2022,sluijterman2024optimal} and also the results we obtain for the 18 datasets considered in this article. Interestingly, the optimization challenges are still observed when the MVE network is trained with a modified loss function that includes a gradient stopping operation, as proposed in \cite{Seitzer2022}, and summarized in \cref{sec:seitzer_beta_loss}. In contrast, the cooperative strategy is stable and does not exhibit this behavior.

\subsection{Finite-Sample Bias Improvement by Cooperative Training}
\label{sec:bias_improvement_by_VeBNNs}

The MVE network requires the solution of a more challenging optimization problem, as described in the previous section. This has important practical implications due to finite-sample bias.

\begin{theorem}[Finite-Sample Bias]
	\label{thm:finite_sample_bias}
	In a finite-sample regime ($N < \infty$), the trained mean is imperfect:
	conditioned on the learned parameters $\hat{\boldsymbol{\theta}}$, its deviation from
	the ground truth, $b(\mathbf{x}) := f^*(\mathbf{x}) - \hat{\mu}(\mathbf{x}; \hat{\boldsymbol{\theta}})$,
	is generally non-zero. The optimal variance estimate then \textbf{absorbs this
	squared deviation}, making it a biased estimator of the true aleatoric
	variance $s^2(\mathbf{x})$:
	$$\mathbb{E}\!\left[\hat{\sigma}^2_a \mid \hat{\boldsymbol{\theta}}\right] = s^2(\mathbf{x}) + b(\mathbf{x})^2 .$$
\end{theorem}

\begin{proof}
	Conditioned on the trained parameters $\hat{\boldsymbol{\theta}}$, the mean
	$\hat{\mu}(\mathbf{x};\hat{\boldsymbol{\theta}})$ is fixed. The optimal aleatoric variance
	minimizes the expected NLL of \cref{eq:loss_MVE}, hence equals the expected
	squared residual at $\mathbf{x}$ over a fresh observation $y = f^*(\mathbf{x}) + \epsilon$
	with $\epsilon \sim \mathcal{N}(0, s^2(\mathbf{x}))$ independent of $\hat{\boldsymbol{\theta}}$:
	\begin{align*}
		\mathbb{E}\!\left[\hat{\sigma}_a^2 \mid \hat{\boldsymbol{\theta}}\right]
		&= \mathbb{E}\!\left[\left(y-\hat{\mu}(\mathbf{x}; \hat{\boldsymbol{\theta}})\right)^2 \,\middle|\, \hat{\boldsymbol{\theta}}\right] \\
		&= \mathbb{E}\!\left[\left( b(\mathbf{x}) + \epsilon \right)^2 \,\middle|\, \hat{\boldsymbol{\theta}}\right] \\
		&= b(\mathbf{x})^2 + \mathbb{E}[\epsilon^2] + 2\,b(\mathbf{x})\,\mathbb{E}[\epsilon]
		 = s^2(\mathbf{x}) + b(\mathbf{x})^2 ,
	\end{align*}
	where the cross-term vanishes because $\mathbb{E}[\epsilon]=0$ and $\epsilon$ is
	independent of the fixed trained parameters $\hat{\boldsymbol{\theta}}$. Since $N<\infty$,
	the deviation $b(\mathbf{x})$ is generally non-zero. This holds whether the variance
	estimate comes from a jointly trained MVE network or a cooperatively trained VE network.
\end{proof}

While we do not prove this in general, the mean network (Step 1) optimizes a convex, saddle-free output-space objective that avoids the degenerate dynamics of \cref{lem:mve_no_critical_point}, so its mean estimate is expected to be better behaved than that of the jointly trained MVE network. Consistent with this, we empirically observe a lower mean-estimation bias for cooperative training than for joint MVE training across all datasets considered in this article:
$$\text{Bias}(\hat{\mu}_{ME}(\mathbf{x},\hat{\boldsymbol{\theta}}_{\text{ME}}))^2 \;<\; \text{Bias}(\hat{\mu}_{\text{MVE}}(\mathbf{x},\hat{\boldsymbol{\theta}}_{\text{MVE}}))^2 \qquad \text{(observed empirically).}$$

Furthermore, iterating the ME and VE training is expected to further reduce the bias of the mean estimator while improving the variance estimate, through the following mechanism: the updated variance estimate of the VE network is used to reweight the samples when training the ME network in the next iteration. We do not prove monotonic improvement, but we observe this behavior empirically, with convergence typically within $K=2$ iterations (\cref{sec:disccussion_iteration}). To see the mechanism, observe what happens after the first iteration:

\begin{itemize}
	\item \textbf{Step 1 (Mean network training) at iteration $k$}: The point estimate of this network is updated to (omitting the regularization term):
	\begin{equation}
    \label{eq:point_estimate_MEnet_iteration_i}
        \hat{\boldsymbol{\theta}}_k = \arg\min_{\boldsymbol{\theta}}\left[ \sum_{n=1}^{N} \frac{\left(\mu(\mathbf{x}_n;\boldsymbol{\theta}) - y_n\right)^2}{\sigma_a^2(\mathbf{x}_n;\hat{\boldsymbol{\phi}}_{k-1})} \right]
	\end{equation}
	where it is important to note that the aleatoric variance $\sigma_a^2(\mathbf{x}; \hat{\boldsymbol{\phi}}_{k-1})$ is fixed from the previous iteration ($k-1$) when training the VE network, but \textbf{it is not constant} (only the log term in \cref{eq:loss_MVE} vanishes). Therefore, finding the point estimate of the ME network in this iteration $k$ remains trivial, but the mean is updated because each sample is associated to a different variance. Samples with low variance will have a large ``weight'' and the network will be forced to fit these points tightly; while samples with high variance will have a small weight and the network will mostly ignore them. Therefore, \textbf{the mean estimate is improved, lowering the bias of the mean estimator}.
	\item \textbf{Step 2 (Variance network training) at iteration $k$}: The VE network estimation will also be updated in this iteration because it is trained on the new residuals $r_{n,k} = (y_n - \hat{\mu}(\mathbf{x}_n; \hat{\boldsymbol{\theta}}_k))^2$, leading to:
	\begin{equation}
        \hat{\boldsymbol{\phi}}_k = \arg\min_{\boldsymbol{\phi}} \left\{ \sum_{n=1}^{N} \left[\frac{r_{n,k}}{2 \sigma_a^2(\mathbf{x}_n;\boldsymbol{\phi})} + \frac{1}{2}\log\big(\sigma_a^2(\mathbf{x}_n;\boldsymbol{\phi})\big)\right] \right\}
    \end{equation}
    Or, equivalently, updated by minimizing the Gamma loss, as previously discussed. In any case, the variance estimation improves because the mean was updated in the beginning of this iteration.
\end{itemize}

In summary, we propose the cooperative training of an ME and VE network to avoid the coupled optimization problem that imposes a non-convex loss and causes a saddle landscape in the last layer. All our experiments validate this theoretical argument. Evidently, the disadvantage of cooperative training is the introduction of an iterative procedure, but surprisingly we observe convergence after one iteration (train initial ME network with constant variance, then train VE network, and then train the ME network with the updated variance). For cases where we are interested in also estimating the epistemic uncertainty, then we do not even train the deterministic ME network again. Instead, we do Step 3 and the BNN network updates the mean and estimates the epistemic uncertainty. If a convergence test is desired, we recommend to iterate between Step 1 and Step 2, and then do the more costly Step 3 (BNN training) only once. Alternatively, it is also possible to iterate between Step 2 and Step 3 -- this is a more principled iterative procedure because all estimates are being updated at each iteration, but it is also the most costly.

\subsection{MVE Network with Loss Function with Gradient Stopping Operation}
\label{sec:seitzer_beta_loss}

$\beta$-NLL loss \cite{Seitzer2022} was proposed by introducing an additional variance-weighting term, which is given as follows:
\begin{equation}
    \label{eq:beta-nll}
     \mathcal{L}_{\beta\text{-NLL}} = \frac{1}{N} \sum_{n=1}^N \lfloor \sigma_a^{2\beta}(\mathbf{x}_n) \rfloor \left[\frac{(y_n - \mu(\mathbf{x}_n))^2}{2 \sigma_a^2(\mathbf{x}_n)} + \frac{\log(\sigma_a^2(\mathbf{x}_n))}{2}\right]
\end{equation}
where $\lfloor \cdot \rfloor $ represents the stop gradient operation. Under this setup, the gradient of the $\beta-$NLL loss becomes:
\begin{align}
\label{eq:L_beta_nll_der}
    \nabla_\mu\mathcal{L}_{\beta\text{-NLL}} &= \frac{1}{N}\sum_{n=1}^{N}\left( \frac{\mu(\mathbf{x}_n) - y_n }{\sigma_a^{2-2\beta}(\mathbf{x}_n)}\right)  \\ \nabla_{\sigma_a^2}\mathcal{L}_{\beta\text{-NLL}} &= \frac{1}{2N}\sum_{n=1}^{N}\left( \frac{\sigma_a^2(\mathbf{x}_n) - (y_n - \mu(\mathbf{x}_n))^2 }{\sigma_a^{4-2\beta}(\mathbf{x}_n)}\right)
\end{align}

According to \cref{eq:beta-nll}, the variance-weighting term $\beta$ can interpolate between the original NLL loss and equivalent MSE, recovering the original NLL loss when $\beta=0$. The gradient of \cref{eq:L_beta_nll_der} is equivalent to MSE for $\beta=1$.

\newpage

\section{Bayesian Neural Networks}
\label{sec:details of BNNs}
In the Bayesian formalism, the posterior parameter distribution is defined as:
\begin{equation}
\label{eq:parameter_posterior}
    p \left ( \boldsymbol{\theta} \mid  \mathcal{D} \right ) = 
    \frac{  p \left ( \mathcal{D}  \mid \boldsymbol{\theta} \right) p \left( \boldsymbol{\theta} \right) }
    { p \left ( \mathcal{D}  \right )}, \, 
    p \left ( \mathcal{D} \right ) = \int p \left ( \mathcal{D} \mid \boldsymbol{\theta}\right) p \left ( \boldsymbol{\theta} \right ) \mathrm{d} \boldsymbol{\theta}
\end{equation}

where $p \left( \boldsymbol{\theta} \right)$ is the prior, typically a Gaussian distribution, $p \left ( \mathcal{D}  \mid \boldsymbol{\theta} \right)$ is the likelihood,  $ p \left ( \mathcal{D} \right )$ is the marginal likelihood (or evidence), which integrates the likelihood over $\boldsymbol{\theta}$, 
and $p \left ( \boldsymbol{\theta} \mid  \mathcal{D}  \right )$ is the posterior distribution of the parameters $\boldsymbol{\theta}$.

The BNN posterior predictive distribution (PPD) for a new data point is computed as follows:
\begin{equation}
\label{eq:ppd_mcmc}
    p \left (y^\prime \mid \mathbf{x}^\prime, \mathcal{D} \right )
    = \int p \left ( y^\prime \mid \mathbf{x}^\prime, \boldsymbol{\theta}\right)  
    p \left ( \boldsymbol{\theta} \mid \mathcal{D} \right ) \mathrm{d} \boldsymbol{\theta}
\end{equation}
where $\mathbf{x}^\prime$ denotes the features of the unknown point, and $y^\prime$ the corresponding target.

\subsection{MCMC Sampling for BNNs}
\label{sec:details of MCMC}
Markov Chain Monte Carlo (MCMC) methods remain the gold standard in Bayesian inference. Usually, the prior is enforced on the weights and biases of the neural network. The most common choice is a multivariate Gaussian prior:
\begin{equation}
    \label{eq:prior_expression}
     p(\bm{\theta}) = \mathcal{N}(\bm{\theta} \mid \mathbf{0}, \kappa^{-1}\mathbf{I}) 
\end{equation}
where $\kappa$ is the precision (inverse variance) of the Gaussian prior, which in deterministic networks is referred to as regularization, and $\mathbf{I}$ is the identity matrix.

The likelihood function $p \left ( \mathcal{D}  \mid \boldsymbol{\theta} \right)$ quantifies how well the neural network output $\mu(\mathbf{x;\boldsymbol{\theta}})$ explains the observations $\mathbf{y}$. Commonly, the observation distribution is assumed to be Gaussian:
\begin{equation}
    \label{eq:bnn_gaussian_likelihood}
    p \left ( \mathcal{D}  \mid \boldsymbol{\theta} \right) = \prod_{n=1}^N \mathcal{N}(\mathbf{y}_n \mid \mu(\mathbf{x}_n; \bm{\theta}), \sigma_a^2(\mathbf{x}_n; \boldsymbol{\phi}))
\end{equation}
where $\mu(\mathbf{x}; \boldsymbol{\theta})$ is the predictive mean of the neural network and $\sigma_a^2(\mathbf{x}; \boldsymbol{\phi})$ is the data noise variance (assumed constant when the noise is homoscedastic, or learned by another neural network parameterized by $\boldsymbol{\phi}$ as described in \cref{sec:variance_training}). 

As shown in \Cref{eq:posterior,eq:parameter_posterior}, it is not possible to get the analytical solution because it requires integrating the posterior and marginal likelihood. However, we can obtain the following relation by omitting the denominator: 
\begin{equation}
    p \left ( \boldsymbol{\theta} \mid  \mathcal{D} \right ) \propto 
     p \left ( \mathbf{y} \mid \boldsymbol{\theta}, \mathbf{x} \right) p \left( \boldsymbol{\theta} \right) 
\end{equation}

Substituting the prior (\cref{eq:prior_expression}) and likelihood (\cref{eq:bnn_gaussian_likelihood}), we obtain:
\begin{align}
     p \left ( \boldsymbol{\theta} \mid  \mathcal{D}  \right )  &\propto \prod_{n=1}^N \mathcal{N}\left(\mathbf{y}_n \mid \mu(\mathbf{x}_n; \bm{\theta}), \sigma_a^2(\mathbf{x}_n; \boldsymbol{\phi})\right) \cdot \mathcal{N}(\bm{\theta} \mid \mathbf{0}, \kappa^{-1}\mathbf{I})\\
    &= \prod_{n=1}^N \frac{1}{\sqrt{2 \pi \sigma_a^2(\mathbf{x}_n; \boldsymbol{\phi})}} \exp\left(-\frac{\left(\mathbf{y}_n - \mu(\mathbf{x}_n; \boldsymbol{\theta})\right)^2}{2 \sigma_a^2(\mathbf{x}_n; \boldsymbol{\phi})}\right) \cdot \frac{\kappa^{m/2}}{(2 \pi)^{m/2}} \exp\left(-\frac{\kappa}{2} \|\boldsymbol{\theta}\|^2\right)
\end{align}
where $m$ is the number of parameters within $\boldsymbol{\theta}$. By taking the logarithmic form, we obtain 
\begin{align}
\label{eq:expansion of pdd}
\log p \left( \boldsymbol{\theta} \mid \mathcal{D}  \right) &=
 \sum_{n=1}^N \left[\log \frac{1}{\sqrt{2 \pi \sigma_a^2(\mathbf{x}_n; \boldsymbol{\phi})}}- \frac{1}{2}\frac{\left( \mathbf{y}_n - \mu(\mathbf{x}_n; \boldsymbol{\theta}) \right)^2}{\sigma_a^2(\mathbf{x}_n; \boldsymbol{\phi})} \right]+ m \log \frac{1}{\sqrt{2 \pi / \kappa}}- \frac{\kappa}{2} \|\boldsymbol{\theta}\|^2
\end{align}
where the first two terms relate to the likelihood and the final two terms only relate to the neural network parameters. In the deterministic setting, the last two often act as regularization or weight decay.

\begin{figure}[h]
    \centering
    \includegraphics[width=0.9\textwidth]{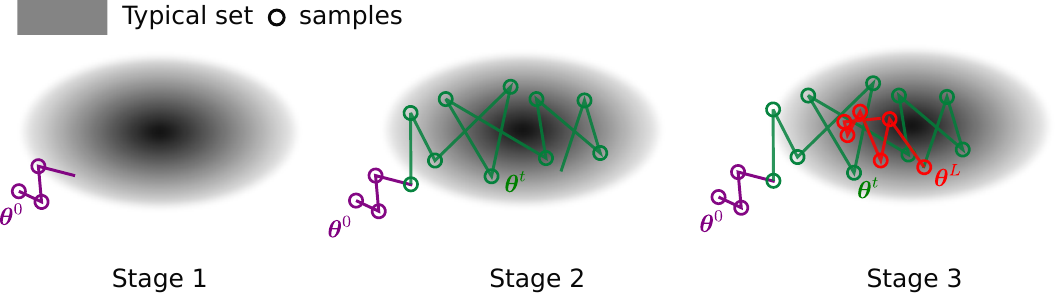}
    \caption{\textbf{Schematic of MCMC-based Bayesian inference.} The shaded area is the typical set, which is the region that covers the high probabilities of the posterior distribution. The samplers start with a random initialization and try to integrate the typical set. It has three stages: the first stage is to search for the typical set; the second stage is the most effective phase, exploring the typical set rapidly; and in the third stage, the sampler might converge to a single mode.  }
    \label{fig:mcmc_schematic}
\end{figure}

MCMC is a class of algorithms that can be used to sample from a probability distribution \cite{Murphy2012}. The most straightforward approach to get the posterior distribution is the random walk Metropolis-Hasting algorithm \cite{Neal1995}, although it suffers from high rejection rates for high-dimensional problems. To accelerate the mixing rate of the MCMC methods, the Hamiltonian Monte Carlo (HMC) was developed by Neal et al. \cite{Neal1995}, which leverages the concept of Hamiltonian mechanics, where the \textit{gradient} of the target probability density function is properly utilized. Therefore, the mixing rate can be improved tremendously compared with the random walk Metropolis-Hasting algorithm. To address the scalability bottleneck of HMC, Welling et al. \cite{Welling2011} developed a novel Bayesian inference approach called Stochastic Gradient Langevin Dynamics (SGLD) leveraging Stochastic Gradient Decent (SGD) \cite{ruder2017overviewgradientdescentoptimization} and Langevin Monte Carlo (LMC) \cite{Murphy2012}. \Cref{fig:mcmc_schematic} illustrates the typical behavior of MCMC-based inference. 

In essence, SGLD starts with a random guess for the unknown neural network parameters $\boldsymbol{\theta}^{(0)}$ in \cref{eq:expansion of pdd}  and then updates the position according to the following rule \cite{Welling2011}:
\begin{equation} \label{eq:sgld_update}
    \Delta \boldsymbol{\theta}_t = \frac{\eta_t}{2} \left(\nabla \log p \left( \boldsymbol{\theta}_t \right)  + \nabla\frac{N}{M} \sum_{n=1}^M\log p(\mathcal{D}_n|\bm{\theta}_t)\right) + \bm{\zeta}_t
\end{equation}
where $\bm{\zeta}_t \sim \mathcal{N}(\bm{0}, \bm{\eta}_t)$, $\eta$ is the step size (learning rate), $N$ is the number of the total data points, $M$ is the batch size. 

However, SGLD assumes that all parameters $\bm{\theta}$ have the same step size, which can lead to slow convergence or even divergence in cases where the components of $\bm{\theta}$ have different curvatures. A refined version of SGLD called preconditioned SGLD (pSGLD) \cite{li2016preconditioned}, was proposed to address this issue. In pSGLD \cite{li2016preconditioned}, the update rule incorporates a user-defined preconditioning matrix $G(\bm{\theta}_t)$, which adjusts the gradient updates and the noise term adaptively:

\begin{equation} \label{eq:sgld}
    \Delta \theta_t = \frac{\eta_t}{2} \left[G(\bm{\theta}_t) \left(\nabla \log p(\bm{\theta}_t )  +\nabla \frac{N}{M} \sum_{n=1}^M p( \mathcal{D}_n | \bm{\theta}_t)\right)  + \chi(\bm{\theta}_t) \right ] + \bm{\zeta}_tG(\bm{\theta}_t)
\end{equation}
where  $\chi_n = \sum_j \frac{\partial G_{n,j}(\bm{\theta}) }{\partial \theta_j};  \, j=0, \cdots, d$ describes how the preconditioner changes with respect to $\bm{\theta}$.

\subsection{Variational Inference for BNNs}
\label{sec:details of VI}
\begin{figure}[h]
    \centering
    \includegraphics[width=0.4\textwidth]{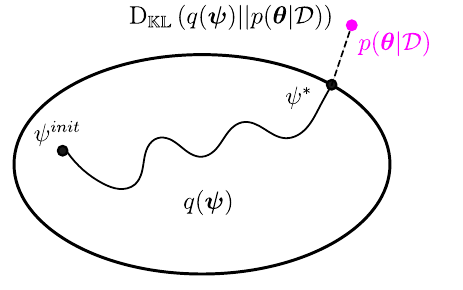}
    \caption{\textbf{Schematic of Variational Inference \cite{Murphy2012}.} The ellipse represents the search space of the proposed distribution, $p(\bm{\theta}|\mathcal{D})$ is the true posterior distribution, the KL divergence is the distance between the two distributions defined as $\mathrm{D}_\mathbb{KL} \left(q(\boldsymbol{\psi})||p(\boldsymbol{\theta}|\mathcal{D}) \right)$, and the goal is to minimize it.}
    \label{fig:vi}
\end{figure}

VI is another type of Bayesian inference method that approximates the posterior distribution by a proposed distribution from a parametric family \cite{Blundell2015, Murphy2012}. The goal of VI is to minimize the Kullback-Leibler divergence between the true posterior distribution $p(\bm{\theta}|\mathcal{D})$ and the proposed distribution $q(\bm{\theta})$ as illustrated in \cref{fig:vi}.

Practically, $\boldsymbol{\psi}$ is regarded as the variational parameter from a parametric family $\boldsymbol\Omega$; therefore, the optimal $\boldsymbol{\psi}^*$ can be obtained by minimizing the KL divergence as follows \cite{Murphy2012}:
\begin{align} \label{eq:vi_inference_equation}
    \boldsymbol{\psi}^* & = \arg \min_{\boldsymbol{\psi}} \mathrm{D}_\mathbb{KL} \left(q(\boldsymbol{\psi}) || p(\boldsymbol{\theta}|\mathcal{D}) \right)  \notag \\
           & =  \arg \min_{\boldsymbol{\psi}} \mathbb{E}_{q(\boldsymbol{\psi})} \left[\log q(\boldsymbol{\psi}) - \log p(\boldsymbol{\theta}|\mathcal{D}) \right]                                                          \notag \\
           & =  \arg \min_{\boldsymbol{\psi}} \mathbb{E}_{q(\boldsymbol{\psi})} \left[\log q(\boldsymbol{\psi}) - \log
    \left(\frac{p(\mathcal{D}| \boldsymbol{\theta})p(\boldsymbol{\theta})}{p(\mathcal{D})} \right) \right]               \notag  \\
           & =  \arg \min_{\boldsymbol{\psi}} \mathbb{E}_{q(\boldsymbol{\psi})} \left[\log q(\boldsymbol{\psi}) - \log p(\mathcal{D}|\boldsymbol{\theta}) - \log p(\boldsymbol{\theta}) +\log p(\mathcal{D}) \right] \notag \\
           & =  \arg \min_{\boldsymbol{\psi}} \mathbb{E}_{q(\boldsymbol{\psi})} \left[\log q(\boldsymbol{\psi}) - \log p(\mathcal{D}| \boldsymbol{\theta}) - \log p(\boldsymbol{\theta})  \right] +\log p(\mathcal{D})
\end{align}
According to \cref{eq:vi_inference_equation}, the optimization problem can be decomposed into two terms: the first term is the negative evidence lower bound (ELBO), and the second term is the log evidence. The log evidence is a constant term that does not depend on $\boldsymbol{\psi}$; therefore, we only need to optimize the first term:
\begin{align}
    \boldsymbol{\psi}^* & = \arg \min_{\boldsymbol{\psi}} \mathbb{E}_{q(\boldsymbol{\psi})} \left[\log q(\boldsymbol{\psi}) - \log p(\mathcal{D}|\boldsymbol{\theta}) - \log p(\boldsymbol{\theta})  \right]                  \notag    \\
           & = \arg \min_{\boldsymbol{\psi}} \mathbb{E}_{q(\boldsymbol{\psi})} \left[ - \log p(\mathcal{D}|\boldsymbol{\theta})\right] + \mathrm{D}_{\mathbb{KL}}(q(\boldsymbol{\psi})|| p(\boldsymbol{\theta})) \
\end{align}
Similarly as \cref{eq:expansion of pdd}, the first term is the negative log-likelihood of the given dataset, and the second term is the KL divergence between the proposed distribution and the prior distribution, which again can be regarded as a regularizer. The optimization problem can be solved by any optimization algorithm, such as SGD \cite{ruder2017overviewgradientdescentoptimization} or Adam \cite{Kingma2014}. After obtaining the best $\boldsymbol{\psi}$, the variational distribution $q(\boldsymbol{\psi})$ can be used as the posterior distribution.

\newpage

\section{Performance Metrics}
\label{sec:performance_metrics}

 No single metric can thoroughly evaluate an algorithm capable of uncertainty quantification and disentanglement \cite{azizi2026clear}. Alternative metrics have been proposed to assess the quality of the estimated covariance when ground-truth uncertainty is unavailable, such as the Task Agnostic Correlations (TAC) metric \cite{TicTac}; however, since our material plasticity dataset provides ground-truth aleatoric uncertainty, we adopt the Wasserstein distance as a more direct measure. Therefore, in this paper, we employ 5 metrics that aim to give a fair comparison of all selected methods. Specifically, we use Root Mean Square Error (RMSE), Test Log-Likelihood (TLL), Wasserstein distance \cite{Kantorovich1960}, Test Coverage (TC), and Test Interval Length (TIL), as detailed below:

\begin{itemize}
    \item \textbf{Root Mean Square Error (RMSE)} \\
- For MLP:
\begin{equation}
\label{eq:rmse}
    \text{RMSE} = \sqrt{\frac{1}{N_{test}} \sum_{i=1}^{N_{test}} \left( \bar{\boldsymbol{\mu}}_i - \mathbf{y}_i \right)^T \left( \bar{\boldsymbol{\mu}}_i - \mathbf{y}_i \right)} 
\end{equation}
where $N_{test}$ is the number of test data points, $\Bar{\boldsymbol{\mu}}$ is the predictive mean, and $\mathbf{y}$ is the observation values of the test points.\\
- For RNN:
\begin{equation}
\label{eq:error}
    \text{RMSE} = \frac{1}{N_{test}\,D}
\sum_{i=1}^{N_{test}}\sum_{d=1}^{D}
\frac{\bigl\lVert \bar{\boldsymbol{\mu}}_{i,:,d}-\bar{\mathbf{y}}_{i,:,d}\bigr\rVert_{2}}
     {\bigl\lVert \bar{\mathbf{y}}_{i,:,d}\bigr\rVert_{2}}
\times 100\%, 
\end{equation}
where $N_{test}$ is the number of testing points, $D$ is the number of output components, $\norm{\cdot}_2$ is the Euclidean norm taken over the time dimension, and $\Bar{\mathbf{y}}_{i,:,d}$ and $\Bar{\bm{\mu}}_{i,:,d}$ are the ground truth and the predictive mean of the $i^{th}$ test sample for the $d^{th}$ output component, correspondingly. Note that it is a normalized percentage for the RNN to eliminate the influence of the magnitude of different stress outputs, since different outputs would have different magnitudes. 

    \item \textbf{Test Log-Likelihood (TLL)}
          \begin{equation} \label{eq:TLL}
              \text{TLL} = \frac{1}{N_{test} T} \sum_{i=1}^{N_{test}} \sum_{j=1}^{T} \log p(\Bar{\mathbf{y}}_{i,j}| \bm{\theta})
          \end{equation}
    We also clarify that we use the ground truth $\Bar{\mathbf{y}}$ for the plasticity law dataset. Observation values $\mathbf{y}$ are used for synthetic and UCI regression datasets and $T=1$. 

    \item \textbf{Wasserstein distance (WA)} 
          \begin{equation} \label{eq:general wassertein distance}
              \text{WA} = \frac{1}{N_{\text{test}} T} \sum_{i=1}^{N_{\text{test}}} \sum_{j=1}^{T} \text{W}_2 \left( 
                p\left(\bar{\boldsymbol{\mu}}_{i,j}, \boldsymbol{{\sigma}}^2_{a_{i,j}} \mathbf{I} \mid \bm{\theta}, \bm{\phi}\right), 
                q\left(\bar{\mathbf{y}}_{i,j} \right) 
                \right)
          \end{equation}
          where $p\left(\bar{\boldsymbol{\mu}}_{i,j}, \boldsymbol{{\sigma}}^2_{a_{i,j}} \mathbf{I} \mid \bm{\theta}, \bm{\phi}\right)$ and $q\left(\bar{\mathbf{y}}_{i,j} \right)$ are the predictive and ground truth distributions, respectively. Since the Gaussian assumption is applied, we can rewrite \cref{eq:general wassertein distance} into:
          \begin{equation}
              \text{WA} = \frac{1}{N_{\text{test}} T} \sum_{i=1}^{N_{\text{test}}} \sum_{j=1}^{T} \sqrt{\left(\bar{\mathbf{y}}_{i,j} - \bar{\boldsymbol{\mu}}_{i,j}\right)^T \left(\bar{\mathbf{y}}_{i,j} - \bar{\boldsymbol{\mu}}_{i,j}\right) + \left( \mathbf{s}_{i,j} - \boldsymbol{\sigma}_{a_{i,j}} \right)^T \left( \mathbf{s}_{i,j} - \boldsymbol{\sigma}_{a_{i,j}} \right) }
          \end{equation}
          where $\mathbf{s}_{i,j}$ and $\boldsymbol{\sigma}_{a_{i,j}}$ are, respectively, the ground-truth and predicted aleatoric standard deviations (note that the $2$-Wasserstein distance between Gaussians involves standard deviations, not variances).

    \item \textbf{Test Coverage (TC)} \\
    Given a confidence level $\alpha$, we can get the predictive confidence interval for the test dataset calibrated by the validation dataset.
     \begin{equation}
   \hat{C}_\alpha= [\mu(\mathbf{x};\bm{\theta}) - l_\alpha \sigma(\mathbf{x}),\mu(\mathbf{x};\bm{\theta}) + l_\alpha \sigma(\mathbf{x})],
    \end{equation}
    
    With the predictive confidence interval, we can obtain the test coverage with the following formula:
    \begin{equation}
        \label{eq:test_coverage} 
        \text{Test coverage} = \frac{1}{N_{test}} \sum_{i=1}^{N_{test}} \mathbb{I}\{y_i^\prime \in \hat{C}_\alpha(\mathbf{x}, \mathcal{D}_{test})\}  .
    \end{equation}
    \item \textbf{Test Interval Length (TIL)}
    The test interval length with a given confidence interval $\alpha$ is given by 
    \begin{equation}
   \text{TIL}=2l_\alpha \sigma(\mathbf{x}),
    \end{equation}

\end{itemize}

\paragraph{Uncertainty calibration} We calibrate the predicted uncertainty using a validation dataset such that every method has the same test coverage (set to be 0.95) on the validation dataset by finding a constant $c$ to adjust the total variance \cite{guo2017calibrationmodernneuralnetworks}. Then, we report the final accuracy metrics with the calibrated total uncertainty for all compared methods. We also note that the calibration is executed for both the UCI regression and the image regression datasets. 

Since the plasticity-law discovery dataset provides test datasets with ground-truth aleatoric uncertainty via 100 repeated simulations per input, it allows a direct assessment of aleatoric accuracy using the WA. Therefore, using calibration would artificially distort WA and eliminate its interpretability. In addition, we wish to preserve the intrinsic behavior of the epistemic uncertainty, where it tightens as more data are observed. Therefore, we keep the calibration factor $c=1$ for all methods, ensuring that WA and Epistemic TLL are calculated based on ground-truth mean and predicted epistemic uncertainty reflect the raw predictive behavior of each model.

\newpage

\section{Additional Experiment Results}
\label{sec:additional_experiments}
\subsection{Illustrative Example: One-Dimensional Dataset}
\label{sec:synthetic_datasets}
\paragraph{Comparison of different inference methods} The predictions for the one-dimensional example by our method with pSGLD inference -- VeBNN (pSGLD) -- are shown in \cref{fig:illustration}. A direct comparison with MVE ($\beta\text{-NLL}$) assuming the best value for the $\beta$ hyperparameter that we found, $\beta=0.5$, is shown in \cref{fig:beta_vs_cuq}. It is clear that our strategy of separately training the mean and aleatoric variance leads to better results, and avoids the need for an extra hyperparameter ($\beta$).

\begin{figure}[h]
\centering
\centerline{\includegraphics[width=0.95\columnwidth]{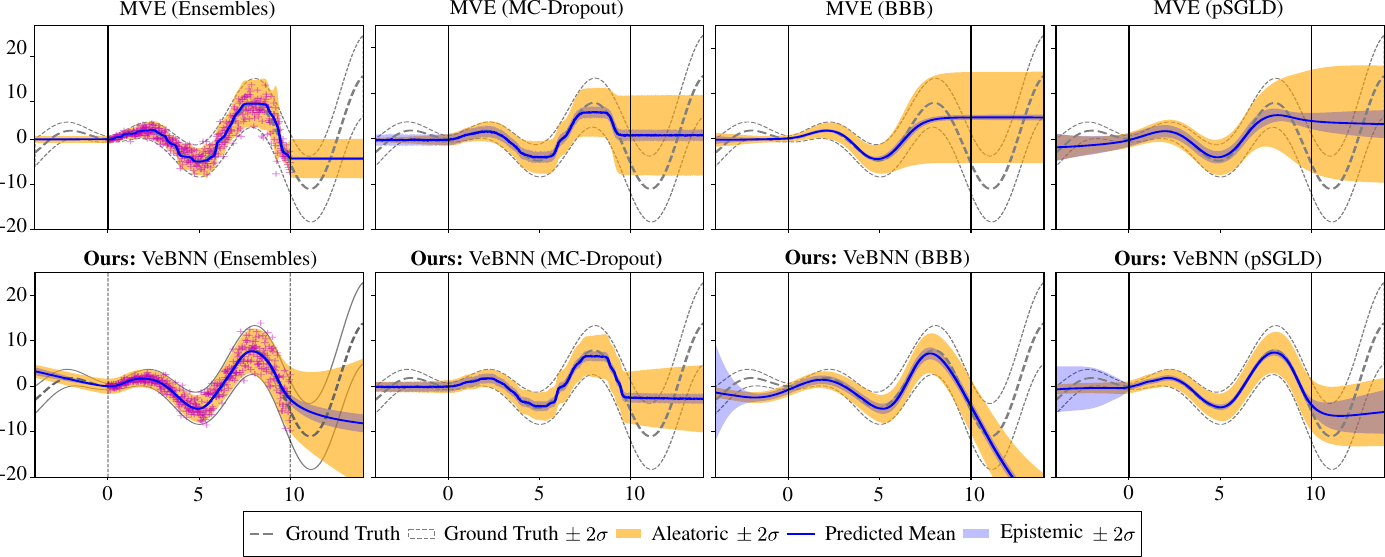}}
\caption{\textbf{Heteroscedastic regression by our method (bottom) compared to jointly trained MVE.}}
\label{fig:illustrative_example}
\end{figure}

\Cref{fig:illustrative_example} also shows the same example when compared to other Bayesian methods with joint MVE training that are capable of predicting both uncertainties. We see a consistent improvement in the predictions with the cooperative training strategy we proposed, independently of the inference method that is chosen.  Joint MVE training strategy with Bayesian inference starts to overestimate aleatoric uncertainty approximately after $x>8$, regardless of the Bayesian inference method. This effect becomes more pronounced in the extrapolation region ($x>10$), which reflects the difficulties that arise from the joint training process. There is a clear tendency to overestimate the aleatoric uncertainty by the joint training strategy that also affects the mean estimation. In contrast, the proposed strategy improves predictions according to all metrics. The pSGLD shows its advantage on this problem, while other inference methods tend to give confident predictions even in the extrapolation regions ($x<0$ and $x>10$).

\begin{figure}[h]
\centering
\centerline{\includegraphics[width=\columnwidth]{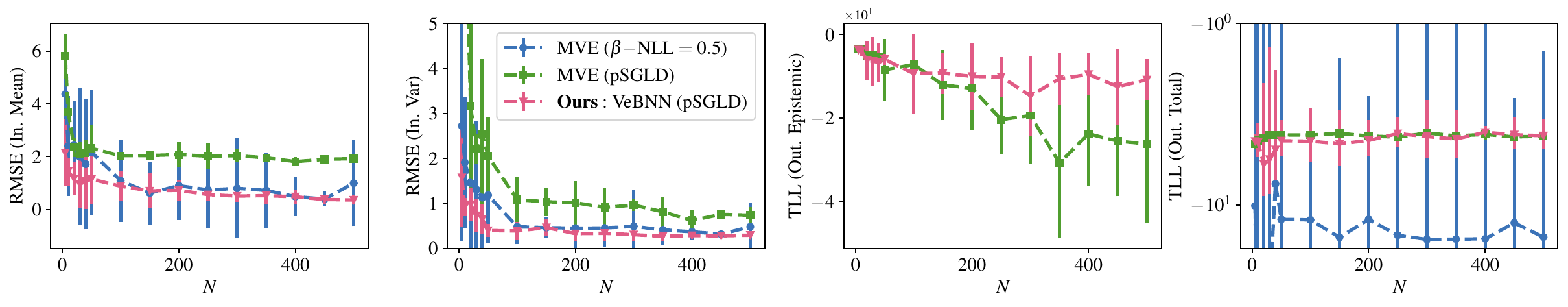}}
\caption{\textbf{RMSE and TLL convergence curves under 5 different seeds for heteroscedastic data generation.} The first figure shows the RMSE between the data values and predictive mean within the interpolation region; the second figure shows the RMSE between the noise standard deviation and the predictive one in the interpolation region since the ground truth of aleatoric uncertainty is known; and the third and fourth figures show the Epistemic TLL and Total TLL values for extrapolation test points.}
\label{fig:rmse_tll_heter_data_points}
\end{figure}

\begin{figure}[h]
\centering
\centerline{\includegraphics[width=\columnwidth]{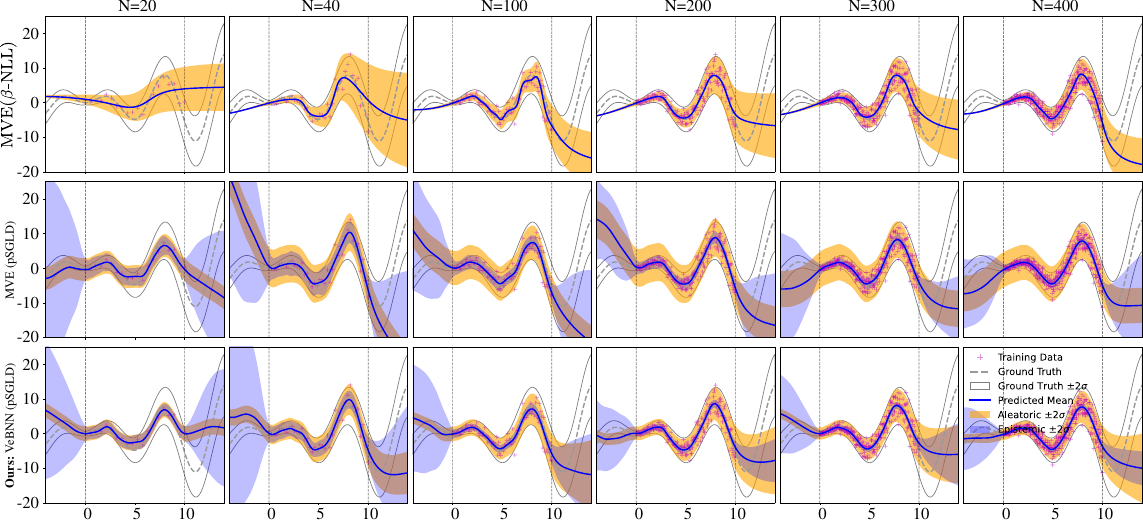}}
\caption{\textbf{Predictions of MVE ($\beta\text{-NLL} =0.5$),  MVE (pSGLD), and \textbf{Ours:} VeBNN (pSGLD) with different data points under heteroscedastic data noise.}}
\label{fig:heter_data_points}
\end{figure}

\paragraph{Heteroscedastic noise}
\label{sec:heter_noise_point} We show the regression results for increasing the number of training data points from $5$ to $500$ based on VeBNN (pSGLD). The accuracy metrics obtained by running each case 5 times for randomly sampled training sets are shown in \cref{fig:rmse_tll_heter_data_points}, and we also visualize the fitting performance of selected methods under an arbitrary run in \Cref{fig:heter_data_points}.

The optimal value for the hyperparameter $\beta$ was 0.5. The experimental results reveal that MVE with $\beta=0.5$ converges to the same mean but slower than the proposed VeBNN (pSGLD). Meanwhile, MVE (pSGLD) has difficulty in converging to either mean or aleatoric uncertainty. In contrast, the proposed strategy successfully combines the strengths of MVE and BNN to effectively model aleatoric uncertainty, mean predictions, and epistemic uncertainty. The results highlight the effectiveness of cooperative training instead of joint MVE training, especially in the case of a data-scarce scenario. Notably, the proposed method maintains a stable TLL value in the extrapolation region, outperforming alternative approaches in this critical area.

\begin{figure}[h]

\centering
\centerline{\includegraphics[width=\columnwidth]{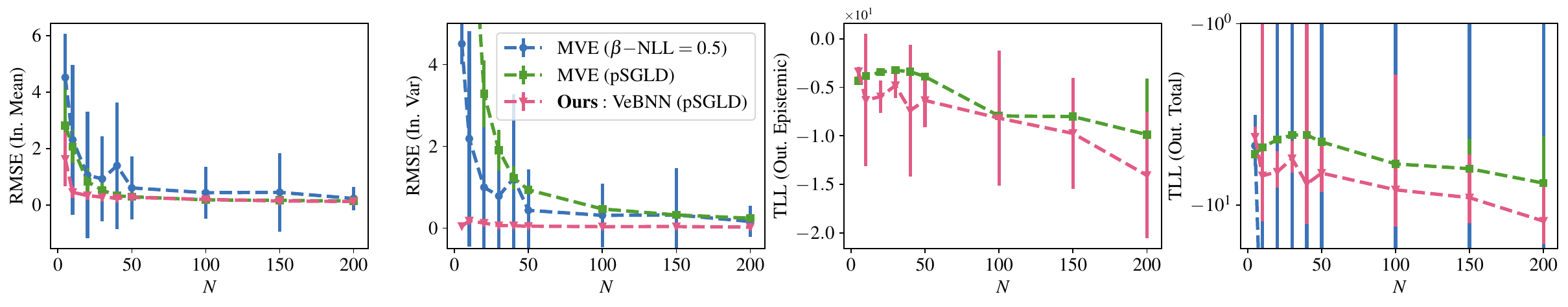}}
\caption{\textbf{RMSE and TLL convergence curves under 5 different seeds for homoscedastic data generation.}}
\label{fig:rmse_tll_homo_data_points}
\end{figure}

\begin{figure}[h]

\centering
\centerline{\includegraphics[width=\columnwidth]{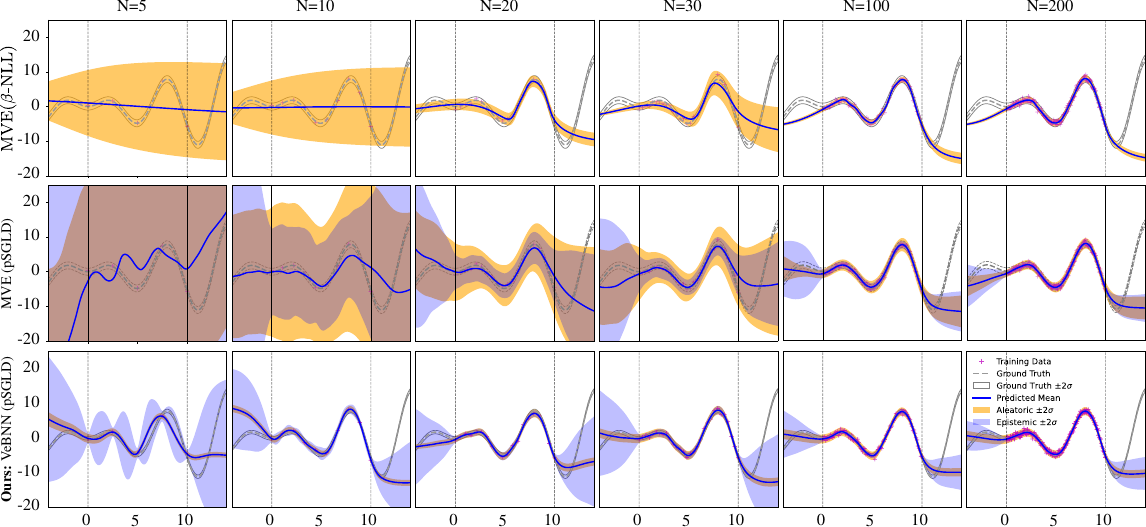}}
\caption{\textbf{Predictions of MVE ($\beta\text{-NLL}=0.5$), BNN-Homo (pSGLD) and \textbf{Ours:} VeBNN (pSGLD) with different data points under homoscedastic data noise.}}
\label{fig:homo_data_points}

\end{figure}

\paragraph{Homoscedastic noise}
\label{sec:homo_data_points}
We execute a similar experiment to test homoscedastic noise cases where the data size changes from $5$ to $200$. \cref{fig:rmse_tll_homo_data_points} and \cref{fig:homo_data_points} highlight similar conclusions as \cref{sec:heter_noise_point}. Again, the MVE training experiences great difficulty when the training data is scarce. MVE (pSGLD) gives really large uncertainties in the cases of  $N<30$, suggesting that it tends to give overconfident uncertainty prediction instead of converging to the mean. However, the proposed cooperative learning strategy has excellent performance even in the case where 5 points are used for training, which is encouraging.

\paragraph{Extension to multi-modal aleatoric noise.} In cases where the outputs exhibit multi-modal aleatoric uncertainty (data noise), a unimodal Gaussian distribution becomes insufficient to properly capture the underlying noise structure. Therefore, in this section, we adopt the illustrative problem from \cite{Harakeh2023} to demonstrate how the proposed VeBNN handles such problems. In short, this toy problem has two separate modes for each input $x$, and their variance changes with different $x$ values. Therefore, it is difficult to use MVE to handle this problem with both MAP and Bayesian inference setups because they would learn a compromised mean with large aleatoric uncertainty, as shown in \cite{Harakeh2023}.

\begin{figure}[h]
    \centering
    \includegraphics[width=0.9\textwidth]{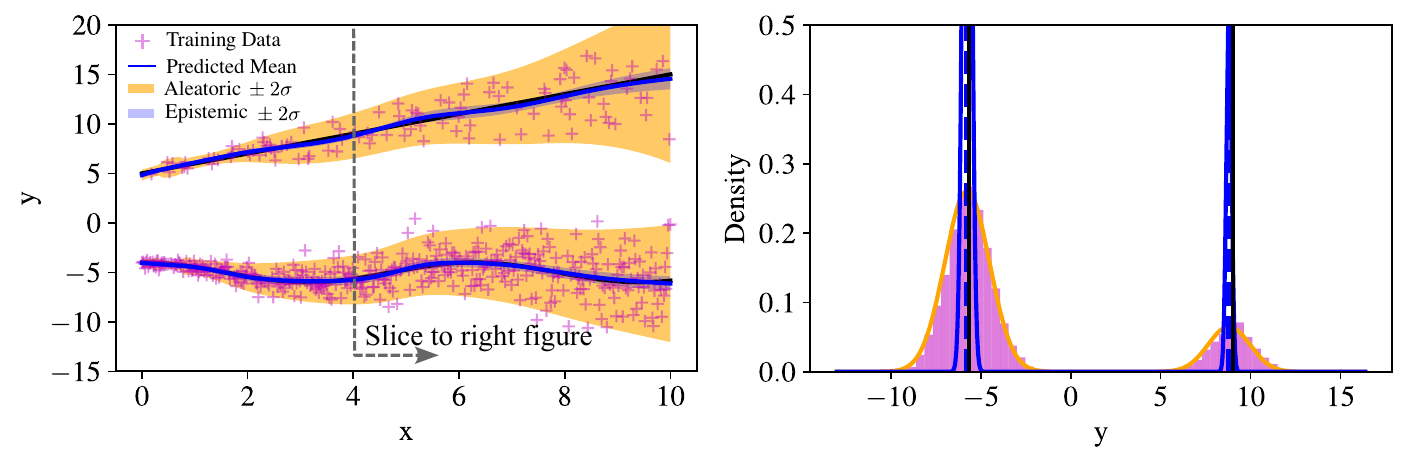}
    \caption{\textbf{Illustration of extension to multi-modal noise based on VeBNN (pSGLD) using Mixture density neural (MDN) network head.} The left figure shows the training data, and predictions with a Gaussian distribution of MDN to be 2, and the right figure shows the uncertainty identification at a test point $x=4$.
    }
    \label{fig:multi_modal_extension}
\end{figure}

In order to deal with the multi-modal data noise, the results in \Cref{fig:multi_modal_extension} employ a Mixture Density Network (MDN) \cite{Bishop1994} as the mean network, which outputs both the mode mean and its corresponding coefficient of each Gaussian distribution. After adjusting the variance network accordingly, VeBNN can be seamlessly trained under this setup. As shown in the left panel of \Cref{fig:multi_modal_extension}, VeBNN (pSGLD) achieves excellent predictive performance. Moreover, the slice at $x=4$ in the right panel further highlights this behavior: the predicted means and aleatoric uncertainties at each mode closely match the ground truth, with negligible epistemic uncertainty.

\subsection{Additional Results for UCI Regression Datasets}
\label{sec:additional_results_UCI_regression}

In \Cref{sec:uci_regression}, due to article length constraints, we only presented the TLL performance where the proposed VeBNN shows clear improvements for each Bayesian inference method considered. Recall that this is demonstrated in every table by the \textit{italic} values that highlight if cooperative training (VeBNN) or joint training (MVE) performs better for each inference method (Ensemble, MC-Dropout, BBB, and pSGLD). In this section, we summarize the results concerning RMSE, TC, and TIL in \Cref{tab:rmse_uci}, \Cref{tab:coverage_uci}, and \Cref{tab:til_uci}, respectively.

\begin{table*}[h]
\caption{\textbf{RMSE ($\downarrow$) results on UCI Regression Datasets.}  Each entry reports the mean RMSE, with the standard deviation shown in parentheses. For each dataset, the best method is marked in \textbf{\underline{bold}}, and the second-best in \textbf{bold}. A superscript $*$ indicates that the best method is significantly better than the second-best (Wilcoxon test \cite{wilcoxon1945individual, demvsar2006statistical}).  Within each inference family, the best-performing variant is italicized; A superscript $+/-$ indicates significantly better or worse performance compared to the other variants in the same family ($p < 0.05$).}

\label{tab:rmse_uci}
\centering
\resizebox{\textwidth}{!}{%
\begin{tabular}{lcccccccc}
\toprule
\multirow{2}{*}{Methods} & \multirow{2}{*}{\shortstack{Carbon\\ (10721, 7,1)}} & \multirow{2}{*}{\shortstack{Concrete\\ (1030, 8,1)}} & \multirow{2}{*}{\shortstack{Energy\\ (768,8,2)}} & \multirow{2}{*}{\shortstack{Boston\\ (506,13,1)}}  & \multirow{2}{*}{\shortstack{Power\\ (9568,4,1)}}   & \multirow{2}{*}{\shortstack{Superconduct\\ (21263,81,1)}} & \multirow{2}{*}{\shortstack{Wine-Red\\ (1599, 11,1)}} & \multirow{2}{*}{\shortstack{Yacht\\ (308,6,1)}}\\
& & & & & & & & \\
\midrule
ME (MSE) & 0.0069 (0.0028) & 5.01 (0.79) & 0.83 (0.10) & 3.89 (0.93)  & 3.88 (0.13) & \textbf{12.19 (0.44)} & \underline{\textbf{0.63 (0.05)}} & 0.90 (0.36) \\
MVE ($\beta\text{-NLL}=1.0$) 
& 0.0084 (0.0024) & 5.64 (0.73) & 0.94 (0.29) & 4.10 (1.15) & 3.91 (0.13) & 13.58 (0.39) &  \underline{\textbf{0.63 (0.05)}} & 2.20 (1.23) \\
MVE (Natural) 
& 0.0072 (0.0026) & 5.24 (0.68) & 1.06 (0.32) & 3.85 (1.48) & 3.84 (0.15) & 13.94 (5.41) & 0.64 (0.05) & 1.39 (1.58) \\

Evidential 
& \underline{\textbf{0.0062 (0.0028)}} & 6.12 (0.79) & 2.42 (0.56) & 4.10 (1.02) & 3.91 (0.15) & 13.93 (0.52) & 0.65 (0.06) & 3.49 (2.84) \\

\midrule
MVE (Ensembles) 
& 0.0066 (1.1927) & 5.07 (1.19) & \textbf{\textit{0.81 (0.15)}} & 3.92 (0.90) & 3.87 (0.12) & \underline{\textbf{\textit{11.85 (0.48)}}}$^{*}$ & 0.74 (0.09) & 1.10 (0.58) \\

\textbf{Ours:} VeBNN (Ensembles) 
& \textit{0.0065 (0.0030)} & \textbf{\textit{4.83 (0.70)}} & 0.83 (0.11) & \textbf{\textit{3.50 (0.96)}} & \textbf{\textit{3.83 (0.14)}} & 12.36 (0.40)$^{-}$ &  \underline{\textbf{\textit{0.63 (0.05)}}}$^{+}$ & \textbf{\textit{0.84 (0.36)}} \\

\midrule
MVE (MC-Dropout) 
& 0.0158 (0.0017) & 5.44 (0.77) & 2.27 (0.39) & 3.94 (1.18) & 4.03 (0.12) & 13.24 (0.40) &  \underline{\textbf{\textit{0.63 (0.05)}}} & 0.96 (0.42) \\

\textbf{Ours:} VeBNN (MC-Dropout) 
& \textit{0.0092 (0.0019)}$^{+}$ & \textit{5.00 (0.70)}$^{+}$ & \textit{1.28 (0.13)}$^{+}$ & \underline{\textbf{\textit{3.49 (1.60)}}} & \textit{3.99 (0.13)} & \textit{12.26 (0.40)}$^{+}$ & 0.64 (0.06) &  \underline{\textbf{\textit{0.79 (0.27)}}} \\

\midrule

MVE (BBB) 
& 0.1748 (0.0594) & 73.61 (54.76) & 63.73 (48.89) & 530.37 (225.64) & 12.85 (8.79) & 335.87 (217.57) & 4.19 (3.72) & 572.11 (188.76) \\

\textbf{Ours:} VeBNN (BBB) 
& \textit{0.0070 (0.0027)}$^{+}$ & \textit{5.62 (0.87)}$^{+}$ & \textit{1.37 (0.28)}$^{+}$ & \textit{3.85 (1.05)}$^{+}$ &\textit{ 4.10 (0.12)}$^{+}$ & \textit{13.96 (0.46)}$^{+}$ &  \underline{\textbf{\textit{0.63 (0.05)}}}$^{+}$ & \textit{1.61 (0.55)} $^{+}$\\

\midrule

MVE (pSGLD) 
& 0.0066 (0.0029) & 5.76 (0.78) & 2.09 (0.32) & 3.88 (0.97) & 3.84 (0.15) & 13.63 (0.38) & 0.64 (0.06) & 1.89 (0.88) \\

\textbf{Ours:} VeBNN (pSGLD) 
& \textbf{\textit{0.0064 (0.0030)}} & \underline{\textbf{\textit{4.70 (0.77)}}}$^{+}$ & \underline{\textbf{\textit{0.77 (0.12)}}}$^{+}$ & \textit{3.64 (0.82)} & \underline{\textbf{\textit{3.79 (0.14)}}}$^{*+}$ & \textit{12.23 (0.45)}$^{+}$ &  \underline{\textbf{\textit{0.63 (0.05)}}} & \textit{0.87 (0.35)}$^{+}$ \\

\bottomrule
\end{tabular}
}
\end{table*}

\begin{table*}[h]
\caption{\textbf{TC ($\rightarrow 0.95$) results on UCI regression datasets.} Each entry reports the mean, with the standard deviation in parentheses. For each dataset, the best overall performance is underlined and in bold, while the second-best is bold. Within each inference method, the best-performing variant is additionally italicized.}

\label{tab:coverage_uci}
\centering
\resizebox{\textwidth}{!}{%
\begin{tabular}{lcccccccc}
\toprule
\multirow{2}{*}{Methods} 
    & \multirow{2}{*}{\shortstack{Carbon\\ (10721, 7,1)}} 
    & \multirow{2}{*}{\shortstack{Concrete\\ (1030, 8,1)}} 
    & \multirow{2}{*}{\shortstack{Energy\\ (768,8,2)}} 
    & \multirow{2}{*}{\shortstack{Boston\\ (506,13,1)}} 
    & \multirow{2}{*}{\shortstack{Power\\ (9568,4,1)}}   
    & \multirow{2}{*}{\shortstack{Superconduct\\ (21263,81,1)}} 
    & \multirow{2}{*}{\shortstack{Wine-Red\\ (1599, 11,1)}} 
    & \multirow{2}{*}{\shortstack{Yacht\\ (308,6,1)}} \\
& & & & & & & & \\
\midrule
MVE ($\beta\text{-NLL}=1.0$) 
& \textbf{0.9496 (0.0074)} 
& \textbf{0.9422 (0.0385)} 
& \textbf{0.9490 (0.0277)} 
& \underline{\textbf{0.9451 (0.0439)}} 
& 0.9448 (0.0083) 
& 0.9508 (0.0065) 
& 0.9556 (0.0252) 
& 0.9081 (0.0757) \\

MVE (Natural)
& \underline{\textbf{0.9497 (0.0063)}} 
& 0.9379 (0.0411) 
& 0.9396 (0.0262) 
& 0.9324 (0.0553) 
& 0.9463 (0.0079) 
& 0.9506 (0.0076) 
& 0.9459 (0.0312) 
& 0.9306 (0.0680) \\

Evidential
& 0.7459 (0.4419) 
& \underline{\textbf{0.9500 (0.0284)}} 
& \underline{\textbf{0.9497 (0.0234)}} 
& 0.9363 (0.0524) 
& 0.9508 (0.0087) 
& 0.9513 (0.0087) 
& 0.9572 (0.0238) 
& \underline{\textbf{0.9565 (0.0395)}} \\

\midrule
MVE (Ensembles)
& \textit{0.9491 (0.0069)} 
& 0.9291 (0.0434) 
& 0.9438 (0.0297) 
& 0.9343 (0.0437)
& \textit{0.9493 (0.0078)} 
& \textbf{\textit{0.9494 (0.0068)}} 
& \textbf{0.9491 (0.0301)} 
& \textit{0.9274 (0.0636)} \\

\textbf{Ours:} VeBNN (Ensembles)
& 0.9478 (0.0057) 
& \textit{0.9325 (0.0414)} 
& \textit{0.9532 (0.0153)} 
& \textbf{\textit{0.9441 (0.0388)}}
& 0.9457 (0.0085) 
& 0.9513 (0.0061) 
& \underline{\textbf{\textit{0.9506 (0.0258)}}} 
& 0.9258 (0.0503) \\

\midrule
MVE (MC-Dropout)
& 0.9601 (0.0160) 
& 0.9301 (0.0426) 
& 0.9373 (0.0257) 
& 0.9402 (0.0395)
& 0.9479 (0.0089) 
& \underline{\textbf{\textit{0.9495 (0.0069)}}} 
& 0.9547 (0.0290) 
& 0.9210 (0.0568) \\

\textbf{Ours:} VeBNN (MC-Dropout)
& 0.9794 (0.0073) 
& \textit{0.9354 (0.0360)} 
& \textit{0.9523 (0.0196)} 
& \textit{0.9431 (0.0376)}
& 0.9462 (0.0084) 
& 0.9512 (0.0059) 
& 0.9569 (0.0247) 
& 0.9177 (0.0682) \\

\midrule
MVE (BBB)
& 0.9998 (0.0003) 
& 0.1015 (0.0473) 
& 0.2075 (0.0707) 
& 0.1598 (0.0677)
& 0.1574 (0.0706) 
& 0.0590 (0.0219) 
& 0.9816 (0.0151) 
& 0.1048 (0.0947) \\

\textbf{Ours:} VeBNN (BBB)
& \textit{0.9474 (0.0068)} 
& \textit{0.9248 (0.0499)} 
& \textit{0.9477 (0.0277)} 
& \textit{0.9324 (0.0578)}
& \textit{0.9477 (0.0068)} 
& \textit{0.9512 (0.0061)} 
& \textit{0.9528 (0.0267)} 
& \textit{0.8903 (0.0853)} \\

\midrule
MVE (pSGLD)
& \textit{0.9489 (0.0058)} 
& 0.9340 (0.0369) 
& 0.9412 (0.0295) 
& \textit{0.9382 (0.0372)}
& 0.9476 (0.0096) 
& \textit{0.9492 (0.0062)} 
& \textit{0.9512 (0.0266)} 
& 0.9018 (0.0664) \\

\textbf{Ours:} VeBNN (pSGLD)
& 0.9473 (0.0064) 
& \textit{0.9398 (0.0388)} 
& \textit{0.9532 (0.0184)} 
& \textit{0.9382 (0.0552)}
& \textbf{\textit{0.9492 (0.0078)}} 
& 0.9511 (0.0068) 
& 0.9528 (0.0284) 
& \textbf{\textit{0.9323 (0.0644)}} \\

\bottomrule
\end{tabular}
}
\end{table*}

\begin{table*}[hbt!]
\caption{\textbf{TIL ($\downarrow$) results on UCI regression datasets.} Each entry reports the mean, with the standard deviation in parentheses. For each dataset, the best overall performance is underlined and in bold, while the second-best is bold. Within each inference method, the best-performing variant is additionally italicized.}

\label{tab:til_uci}
\centering
\resizebox{\textwidth}{!}{%
\begin{tabular}{lcccccccc}
\toprule
\multirow{2}{*}{Methods} 
    & \multirow{2}{*}{\shortstack{Carbon\\ (10721, 7,1)}} 
    & \multirow{2}{*}{\shortstack{Concrete\\ (1030, 8,1)}} 
    & \multirow{2}{*}{\shortstack{Energy\\ (768,8,2)}} 
    & \multirow{2}{*}{\shortstack{Boston\\ (506,13,1)}} 
    & \multirow{2}{*}{\shortstack{Power\\ (9568,4,1)}}   
    & \multirow{2}{*}{\shortstack{Superconduct\\ (21263,81,1)}} 
    & \multirow{2}{*}{\shortstack{Wine-Red\\ (1599, 11,1)}} 
    & \multirow{2}{*}{\shortstack{Yacht\\ (308,6,1)}} \\
& & & & & & & & \\
\midrule
MVE ($\beta\text{-NLL}=1.0$)
& 0.0366 (0.0041) 
& 22.22 (4.15) 
& 3.13 (0.55) 
& 18.62 (9.42)
& 14.91 (0.53) 
& 51.27 (2.97) 
& \textbf{2.56 (0.28)} 
& 7.35 (3.56) \\

MVE (Natural)
& 0.0213 (0.0016) 
& 20.31 (3.24) 
& 3.13 (0.68) 
& 13.07 (5.27)
& 14.59 (0.54) 
& 39.27 (1.92) 
& \underline{\textbf{2.51 (0.24)}} 
& 2.43 (1.16) \\

Evidential
& 0.0292 (0.0079) 
& 28.61 (9.31) 
& 9.07 (4.12) 
& 17.01 (6.60)
& 15.68 (0.77) 
& 46.55 (3.58) 
& 3.40 (0.86) 
& 13.70 (14.01) \\

\midrule
MVE (Ensembles)
& 0.0172 (0.0056) 
& 18.85 (3.35) 
& \textbf{\textit{2.51 (0.39)}} 
& 13.97 (2.43)
& 18.93 (0.55) 
& \textbf{\textit{36.91 (1.14)}} 
& 3.05 (0.36) 
& 3.63 (1.50) \\

VeBNN (Ensembles)
& \textit{0.0143 (0.0009)} 
& \underline{\textbf{\textit{17.24 (2.98)}}} 
& 2.60 (0.28) 
& \textbf{\textit{12.27 (2.86)}}
& \textbf{\textit{14.40 (0.46)}} 
& 38.26 (1.27) 
& \textit{2.65 (0.24)} 
& \textbf{\textit{2.32 (0.82)}} \\

\midrule
MVE (MC-Dropout)
& 0.0496 (0.0042) 
& \textit{18.99 (2.61)} 
& 7.22 (0.79) 
& 14.38 (5.54)
& 15.14 (0.53) 
& \textit{40.11 (1.33)} 
& \textit{2.62 (0.33)} 
& 2.80 (0.96) \\

VeBNN (MC-Dropout)
& \textit{0.0304 (0.0004)} 
& 19.15 (2.71) 
& \textit{4.13 (0.48)} 
& \underline{\textbf{\textit{11.43 (1.99)}}}
& \textit{14.86 (0.59)} 
& 41.56 (1.59) 
& 2.88 (0.33) 
& \underline{\textbf{\textit{1.89 (0.51)}}} \\

\midrule
MVE (BBB)
& 3.93 (0.02) 
& \textit{18.23 (14.56)} 
& 31.05 (33.95) 
& 249.92 (114.79)
& \textit{4.37 (1.32)} 
& \underline{\textbf{\textit{24.75 (11.51)}}} 
& 20.88 (18.59) 
& 136.95 (44.57) \\

VeBNN (BBB)
& \textit{0.0146 (0.0010)} 
& 19.84 (3.11) 
& \textit{4.20 (1.49)} 
& \textit{13.60 (4.76)}
& 15.16 (0.58) 
& 43.83 (1.60) 
& \textit{2.63 (0.27)} 
& \textit{4.46 (0.99)} \\

\midrule
MVE (pSGLD)
& \underline{\textbf{\textit{0.0112 (0.0009)}}} 
& 20.12 (2.46) 
& 4.88 (0.92) 
& 15.22 (4.21)
& \underline{\textbf{\textit{14.31 (0.44)}}} 
& 42.88 (1.61) 
& 2.79 (0.30) 
& 4.80 (2.13) \\

VeBNN (pSGLD)
& \textbf{0.0141 (0.0011)} 
& \textbf{\textit{17.71 (2.93)}} 
& \underline{\textbf{\textit{2.45 (0.27)}}} 
& \textit{12.60 (3.44)}
& 14.66 (0.53) 
& \textit{39.94 (2.00)} 
& \textit{2.66 (0.26)} 
& \textit{2.71 (0.68)} \\

\bottomrule
\end{tabular}
}
\end{table*}

Overall, the cooperative strategy (labeled as VeBNN) performs better for every inference method across almost all datasets. It should be mentioned that the deterministic network, ME (MSE), is competitive because \Cref{tab:rmse_uci} only evaluates the error in estimating the mean. However, it is still slightly worse than VeBNN (pSGLD), which is the best for the UCI regression datasets. We raise awareness that the proposed cooperative training strategy improves over joint MVE training for the large majority of dataset–inference combinations across all four Bayesian inference methods (Deep Ensembles, MC-Dropout, BBB, and pSGLD). This reinforces that the proposed strategy is clearly beneficial, independent of the Bayesian inference method of choice. It is worth noting that BBB has great difficulty converging when adopting joint training for MVE, as shown in \Cref{tab:rmse_uci}. However, by adopting the cooperative training strategy, VeBNN (BBB) reports comparable performance to state-of-the-art methods. Nevertheless, as we discuss in the main text, we recommend the use of pSGLD for the types of distributions considered herein.

\subsection{Additional Results for Image Regression Datasets} 
\label{sec:image_regression_additional_results}

\subsubsection{Complete Results Compared to Baselines}
\label{sec:image_regression_compared_to_baseline}
As stated in \cite{gustafsson2023reliableregressionmodelsuncertainty}, the purpose of this dataset is to evaluate the uncertainty quantification capabilities of deep learning models under distribution shift. MVE (Ensembles) consistently serves as the strongest baseline across all problems, aligning with findings from \cite{gustafsson2023reliableregressionmodelsuncertainty}. \textit{Cells} and \textit{ChairAngle} are the two baseline tasks without any distribution shift between training and test sets.  In these cases, the VeBNNs achieve the targeted test coverage of $95\%$ (\Cref{tab:coverage_imgreg}), while performing better by reaching larger TLL (\Cref{tab:tll_imgreg}), smaller RMSE (\Cref{tab:rmse_imgreg}), and test interval length (\Cref{tab:til_imgreg}) for both inference methods: Ensembles and pSGLD. This is relevant because, independently of the inference method that is chosen, the benefits of cooperative training are noteworthy for all accuracy metrics, and training is more robust (less sensitive to the hyperparameters).

For tasks with distribution shift (Cells-Gap, Cells-Tail, ChairAngle-Gap, ChairAngle-Tail, Skin and Aerial), the VeBNN (Ensembles) reach a test coverage closer to the $95\%$ target than other methods, while achieving better TLL. We note that VeBNN does not improve every case: on ChairAngle-Tail and ChairAngle-Gap, VeBNN (pSGLD) underperforms its jointly trained MVE (pSGLD) counterpart in TLL. This reflects a known difficulty of pSGLD under strong distribution shift, where the posterior is explored less reliably; notably, VeBNN (Ensembles) remains the strongest method on these shifted tasks. This is expected; the model is tested on a region without nearby training points, therefore, its mean predictions are less trustworthy, and there is an increase in predicted epistemic uncertainty. Therefore, this would be advantageous in decision-making scenarios.  

\begin{figure}[h]
\centering
\centerline{\includegraphics[width=0.9\textwidth]{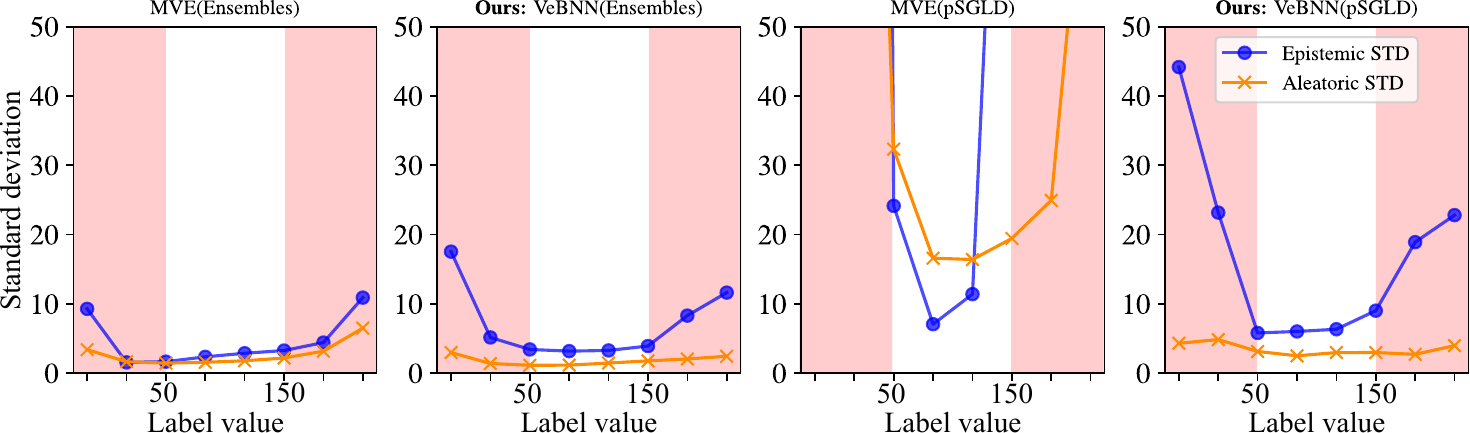}}
\caption{\textbf{Uncertainty predictions and decompositions for the \textit{Cells-Tail} problem.} The white area represents the in-distribution labels, and the light pink area represents out-of-distribution labels.}
\label{fig:image_regression_uncertainty_decomposition}
\end{figure}

To further demonstrate this effect, we plot the predicted aleatoric and epistemic uncertainties of jointly trained MVE and VeBNN with both Ensembles and pSGLD inference in \Cref{fig:image_regression_uncertainty_decomposition}. Although these datasets cannot be used to evaluate the quality of aleatoric uncertainty estimation, we noticed that the proposed VeBNN methods tend to have higher epistemic uncertainty in out-of-distribution regions (datasets ending with ``-Tail''), while the predicted aleatoric uncertainty remains stable even in these regions, regardless of whether Ensembles or pSGLD are used. This suggests that VeBNN may disentangle aleatoric uncertainty more effectively than existing approaches. In contrast, for both MVE (Ensembles) and MVE (pSGLD), the two uncertainty types behave similarly: they remain small within the in-distribution region and increase simultaneously when entering out-of-distribution regions.

\begin{table*}[h]
\caption{\textbf{RMSE ($\downarrow$) results on image regression datasets.} Each entry reports the mean RMSE, with the standard deviation shown in parentheses. A superscript $*$ indicates that the best method is significantly better than the second-best (Wilcoxon test \cite{wilcoxon1945individual, demvsar2006statistical}).  Within each inference family, the best-performing variant is italicized; A superscript $+/-$ indicates significantly better or worse performance compared to the other variants in the same family ($p < 0.05$).}

\label{tab:rmse_imgreg}
\centering
\resizebox{\textwidth}{!}{%
\begin{tabular}{lcccccccc}
\toprule
Methods
    & Cells
    & Cells-Gap
    & Cells-Tail
    & ChairAngle
    & ChairAngle-Gap
    & ChairAngle-Tail
    &Skin
    &Aerial\\
& & & & & & & & \\
\midrule
MVE ($\beta\text{-NLL}=0.5$)
& 5.37 (2.24) & 9.79 (2.03) & 24.97 (4.79) & 0.44 (0.22)
& \textbf{2.70 (0.47)} & 6.56 (0.33) & 497.01 (49.91) & 534.29 (48.11) \\

MVE (Natural) 
& 1111.10 (2302.96) & 278836.40 (623369.05) & 58.32 (1.52) & 25.97 (0.18)
& 26.04 (0.16) & 25.85 (0.36) & 5667.62 (8615.82) & 1011.35 (529.46) \\

Evidential 
& 5.56 (1.44) & \underline{\textbf{7.29 (1.63)}} & 29.11 (2.79) & 0.45 (0.22)
& 3.46 (0.46) &\underline{\textbf{5.68 (0.26)}}$^*$ & 485.41 (23.23) & 565.96 (29.64) \\
\midrule
MVE (Ensembles) 
& 4.82 (0.46) & \textbf{\textit{7.88 (0.84)}} & \textbf{24.28 (1.55)} & 0.34 (0.08)
& \underline{\textbf{\textit{2.37 (0.19)}}} & \textbf{\textit{6.46 (0.13)}} & \underline{\textbf{\textit{446.29 (9.67)}}} & \underline{\textbf{\textit{456.06 (22.18)}}} \\

VeBNN (Ensembles) 
& \underline{\textbf{\textit{3.67 (0.52)}}} & 8.22 (0.23) & \underline{\textbf{\textit{22.96 (1.63)}}} & \underline{\textbf{\textit{0.24 (0.07)}}}
& 5.53 (2.60)$^-$ & 6.66 (0.05)$^-$ & 454.71 (6.55) & 572.47 (18.85)$^-$ \\
\midrule
MVE (pSGLD) 
& 30.74 (34.90) & 21.91 (6.92) & 414.41 (698.37) & 35.71 (31.92)
& 34.45 (30.23) & 35.90 (20.74) & 463.60 (10.60) & 1003.62 (139.64) \\

VeBNN (pSGLD) 
& \textbf{\textit{4.44 (1.01)}}$^+$ & \textit{8.06 (0.46)}$^+$ & \textit{39.03 (26.79)}$^+$ & \textbf{\textit{0.27 (0.08)}}$^+$
& \textit{3.88 (0.61)} $^+$& \textit{8.98 (1.33)}$^+$ & \textbf{\textit{452.89 (23.66)}} & \textbf{\textit{479.29 (46.35)}}$^+$ \\

\bottomrule
\end{tabular}
}
\end{table*}

\begin{table*}[h]
\caption{\textbf{TC ($\rightarrow 0.95$) results on image regression datasets.} Each entry reports the mean coverage, with the standard deviation in parentheses. For each dataset, the best overall performance is underlined and in bold, while the second-best is bold. Within each inference method, the best-performing variant is additionally italicized.}

\label{tab:coverage_imgreg}
\centering
\resizebox{\textwidth}{!}{%
\begin{tabular}{lcccccccc}
\toprule
Methods
    & Cells
    & Cells-Gap
    & Cells-Tail
    & ChairAngle
    & ChairAngle-Gap
    & ChairAngle-Tail
    & Skin
    & Aerial \\
& & & & & & & & \\
\midrule
MVE ($\beta\text{-NLL}=0.5$)
& 0.9479 (0.0028) 
& 0.7224 (0.1039) 
& 0.5769 (0.0447) 
& 0.9540 (0.0026)
& 0.7019 (0.0146) 
& 0.6638 (0.0215) 
& 0.8344 (0.0233) 
& 0.8472 (0.0426) \\

MVE (Natural) 
& 0.9128 (0.0845) 
& \underline{\textbf{0.9749 (0.0039)}} 
& 0.5436 (0.0748) 
& 0.9529 (0.0020)
& \underline{\textbf{0.9679 (0.0006)}} 
& 0.6812 (0.0279) 
& 0.5413 (0.3497) 
& \textbf{0.9325 (0.1146)} \\

Evidential 
& \underline{\textbf{0.9502 (0.0022)}} 
& 0.6939 (0.0482) 
& 0.5581 (0.0180) 
& 0.9815 (0.0126)
& 0.7168 (0.0151) 
& 0.6837 (0.0081) 
& 0.8381 (0.0144) 
& 0.7694 (0.0307) \\
\midrule
MVE (Ensembles) 
& 0.9442 (0.0056) 
& 0.7452 (0.0385)
& 0.5604 (0.0378)
&0.9518 (0.0033)
& 0.7494 (0.0160) 
& 0.6471 (0.0075) 
& 0.8371 (0.0125) 
& 0.8550 (0.0280) \\

VeBNN (Ensembles) 
& \textit{0.9487 (0.0019)} 
& \textit{0.8198 (0.0351)} 
& \textit{0.5869 (0.0207)} 
& \textit{0.9518 (0.0016)}
& \textit{0.7971 (0.0370)} 
& \textit{0.6507 (0.0067)} 
& \textbf{\textit{0.8612 (0.0038)}} 
& \textit{0.9019 (0.0146)} \\
\midrule
MVE (pSGLD) 
& 0.9450 (0.0048) 
& \textbf{\textit{0.9114 (0.0485)}} 
&  \underline{\textbf{\textit{0.8492 (0.1065)}}} 
& \textbf{0.9491 (0.0036)}
& \textbf{\textit{0.9278 (0.0303)}}
& \underline{\textbf{\textit{0.7797 (0.0536)}}}
& \underline{\textbf{\textit{0.8625 (0.0116)}}} 
& 0.6724 (0.0600) \\

VeBNN (pSGLD) 
& \underline{\textbf{\textit{0.9508 (0.0048)}}} 
& 0.8416 (0.0755) 
& \textbf{0.7827 (0.1695)} 
& \underline{\textbf{\textit{0.9503 (0.0059)}}}
& 0.7654 (0.0255) 
& \textbf{0.6938 (0.0325)}
& 0.8561 (0.0164) 
& \textbf{\textit{0.9024 (0.0243)}} \\

\bottomrule
\end{tabular}
}
\end{table*}

\begin{table*}[h]
\caption{\textbf{TIL ($\downarrow$) results on image regression datasets.} Each entry reports the mean interval length, with the standard deviation in parentheses. For each dataset, the best overall performance is underlined and in bold, while the second-best is bold. Within each inference method, the best-performing variant is additionally italicized.}

\label{tab:til_imgreg}
\centering
\resizebox{\textwidth}{!}{%
\begin{tabular}{lcccccccc}
\toprule
Methods 
    & Cells
    & Cells-Gap
    & Cells-Tail 
    & ChairAngle
    & ChairAngle-Gap
    & ChairAngle-Tail
    & Skin
    & Aerial \\
& & & & & & & & \\
\midrule
MVE ($\beta\text{-NLL}=0.5$) 
& 14.65 (6.29) & 18.18 (7.43) & \textbf{18.23 (4.68)} & 1.57 (0.55)
& \underline{\textbf{1.83 (0.36)}} & 2.33 (0.93) & 886.30 (163.04) & 1546.45 (216.47) \\

MVE (Natural) 
& 196.27 (7.23) & 210.25 (21.55) & 108.23 (14.06) & 85.97 (1.58)
& 89.09 (1.88) & 60.05 (2.65) & 1886.45 (1281.29) & 2539.68 (238.92) \\

Evidential 
& 13.05 (3.39) & \underline{\textbf{11.49 (3.25)}} & \underline{\textbf{14.84 (2.72)}} & 1.88 (0.66)
& 2.58 (0.51) & 2.06 (0.53) & \textbf{748.02 (40.17)} &\underline{\textbf{ 1030.07 (31.59)}} \\
\midrule
MVE (Ensembles) 
& \textbf{\textit{12.48 (1.19)}} & \textbf{\textit{16.74 (1.98)}} & 22.02 (2.22) & \textbf{1.20 (0.12)}
& \textbf{\textit{2.46 (0.30)}} & \underline{\textbf{\textit{1.39 (0.17)}}} & \underline{\textbf{\textit{727.95 (27.48)}}} & \textbf{\textit{1486.05 (62.13)}}\\

VeBNN (Ensembles) 
& 13.52 (1.59) & 25.57 (3.32) & \textit{19.01 (4.33)} & \underline{\textbf{\textit{1.05 (0.16)}}}
& 10.08 (6.27)& \textbf{1.52 (0.42)} & 933.98 (45.17) & 1568.95 (175.70) \\

\midrule
MVE (pSGLD) 
& 45.73 (24.38) & 49.59 (20.10) & 439.95 (691.24) & 40.80 (15.63)
& 42.06 (16.22) & 47.65 (20.68) & 917.17 (66.14) & 1562.00 (92.58) \\

VeBNN (pSGLD) 
& \underline{\textbf{\textit{11.87 (1.33)}}} & \textit{21.58 (5.70)} & \textit{66.25 (53.73)} & \textit{1.31 (0.19)}
& \textit{5.17 (1.52)} & \textit{2.97 (1.30)} & \textit{885.89 (68.61)} & \textit{1510.50 (97.49)} \\

\bottomrule
\end{tabular}
}
\end{table*}

\subsubsection{Results of MVE (Ensembles) with Original NLL Loss}
\label{sec:mve_ensemble_orginal_nll_loss}

\begin{table*}[hbt!]
\centering
\caption{\textbf{Comparison of the performance of MVE (Ensembles) trained with $\beta\text{-NLL}=0.5$ and Original NLL across image regression datasets.} Each entry reports the mean interval length, with the standard deviation in parentheses.}

\begin{scriptsize}
\label{tab:beta_vs_nll_imgreg}
\begin{tabular}{l l c c c c}
\toprule
\textbf{Problem} & \textbf{Method} & \textbf{TLL}($\uparrow$) & \textbf{RMSE} ($\downarrow$)& \textbf{TC} ($\rightarrow 0.95$) & \textbf{TIL} ($\downarrow$) \\
\midrule
\multirow{2}{*}{Cells} 
    & $\beta\text{-NLL}=0.5$      & -2.45 (0.06) & 4.82 (0.46)   & 0.9442 (0.0056)  & 12.48 (1.19) \\
    & Original NLL    & -2.05 (0.09) &18.23 (8.39)  &  0.9473 (0.0058)  &12.62 (1.81)  \\
\midrule
\multirow{2}{*}{Cells-Gap} 
    & $\beta\text{-NLL}=0.5$    & -3.89 (0.14) &  7.88 (0.84)&  0.7452 (0.0385)& 16.74 (1.98) \\
    & Original NLL    & -5.24 (1.24) &8.20 (0.96)  &0.6843 (0.1021)  & 14.43 (6.69) \\
\midrule
\multirow{2}{*}{Cells-Tail} 
    & $\beta\text{-NLL}=0.5$      & -6.06 (0.78)  & 24.28 (1.55) & 0.5604 (0.0378) & 22.02 (2.22) \\
    & Original NLL    & -10.23 (3.44) &  23.04 (2.00) & 0.5399 (0.0115) & 14.53 (2.33) \\
\midrule
\multirow{2}{*}{ChairAngle} 
    & $\beta\text{-NLL}=0.5$       &-0.32 (0.23)  & 0.34 (0.08) &  0.9518 (0.0033)& 1.20 (0.12) \\
    & Original NLL    & -3.80 (0.21) &11.25 (1.70)  &   0.9556 (0.0012) & 47.85 (11.68) \\
\midrule
\multirow{2}{*}{ChairAngle-Gap} 
    &$\beta\text{-NLL}=0.5$      &-5.86 (2.37)  &2.37 (0.19)  &  0.7494 (0.0160) & 2.46 (0.30) \\
    & Original NLL    &-3.69 (0.19)& 10.30 (1.57) &  0.9522 (0.0190)&   40.16 (10.35)\\
\midrule
\multirow{2}{*}{ChairAngle-Tail} 
    &$\beta\text{-NLL}=0.5$       &  -93.40 (32.28) & 6.46 (0.13) & 0.6471 (0.0075)  &  1.39 (0.17)\\
    & Original NLL    & -5.08 (0.40) &  34.50 (19.52) & 0.7456 (0.0357) & 64.21 (36.42) \\
\midrule
\multirow{2}{*}{Skin} 
    & $\beta\text{-NLL}=0.5$      & -7.61 (0.12) &446.29 (9.67)  & 0.8371 (0.0125)  & 727.95 (27.48)  \\
    & Original NLL    & -7.56 (0.24)  &  479.44 (43.51) & 0.8734 (0.0136) & 1013.12 (133.83) \\
\midrule
\multirow{2}{*}{Aerial} 
    & $\beta\text{-NLL}=0.5$      & -7.76 (0.27) & 456.06 (22.18) &  0.8550 (0.0280) & 1486.05 (62.13) \\
    & Original NLL    &  -7.76 (0.27) &  726.16 (148.76) &   0.9473 (0.0575)&  2374.48 (184.77)\\
\bottomrule
\end{tabular}
\end{scriptsize}
\end{table*}

Although MVE (Ensembles) is the strongest baseline on the image regression datasets, we find that it is highly sensitive to the choice of $\beta$. We present the results of MVE (Ensembles) trained with the original NLL loss and compare them to those obtained with the $\beta$-NLL formulation in \cref{tab:beta_vs_nll_imgreg}. As shown, the performance degrades substantially when using the original NLL loss, due to its tendency to overfit the data noise while underfitting the mean.

\subsection{Plasticity Laws Datasets} 
\label{sec:plasticity_additional_results}

\begin{figure*}[h]
\centering
\centerline{\includegraphics[width=0.95\textwidth]{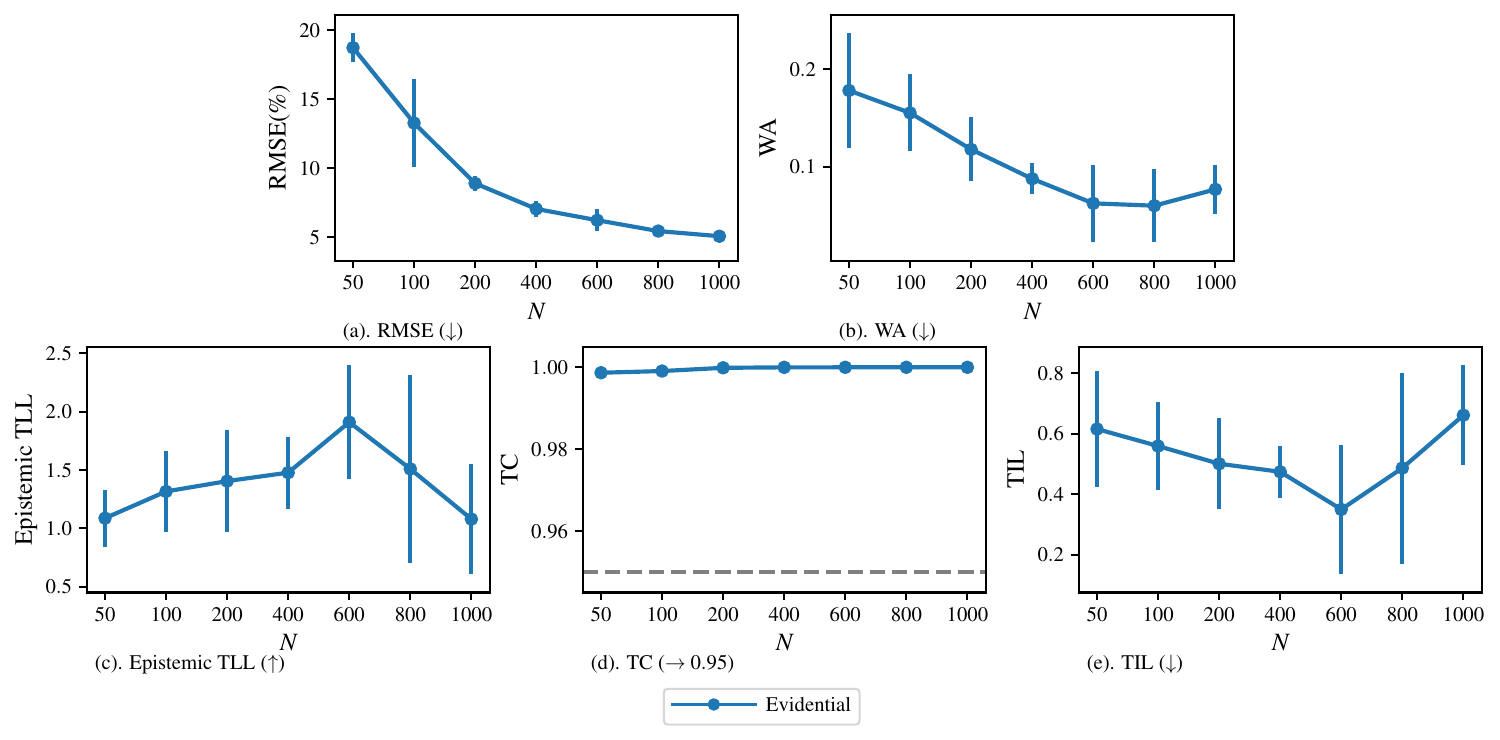}}
\caption{\textbf{Accuracy metrics obtained for the plasticity law discovery dataset considering a training set with different number $N$ of training sequences with Evidential method}.}
\label{fig:deep_evidential_plasticity}

\end{figure*}

\paragraph{Performance of Evidential method} \Cref{fig:deep_evidential_plasticity} gives the performance metrics of the Evidential method for the plasticity laws datasets. It is observed that Evidential can only predict a comparable mean, while its aleatoric and epistemic uncertainties predictions are far worse than the methods reported in \Cref{fig:sve_1_case_statistics}. It starts by overfitting the aleatoric uncertainty, leading to large WA values and an inflated visualization in \Cref{fig:plasticity_prediction}. Then, the corresponding epistemic uncertainty prediction leads to an undesirable  TC close to 1 as shown in \Cref{fig:deep_evidential_plasticity}d).

\begin{figure*}[hbt!]
\centering
\centerline{\includegraphics[width=0.95\textwidth]{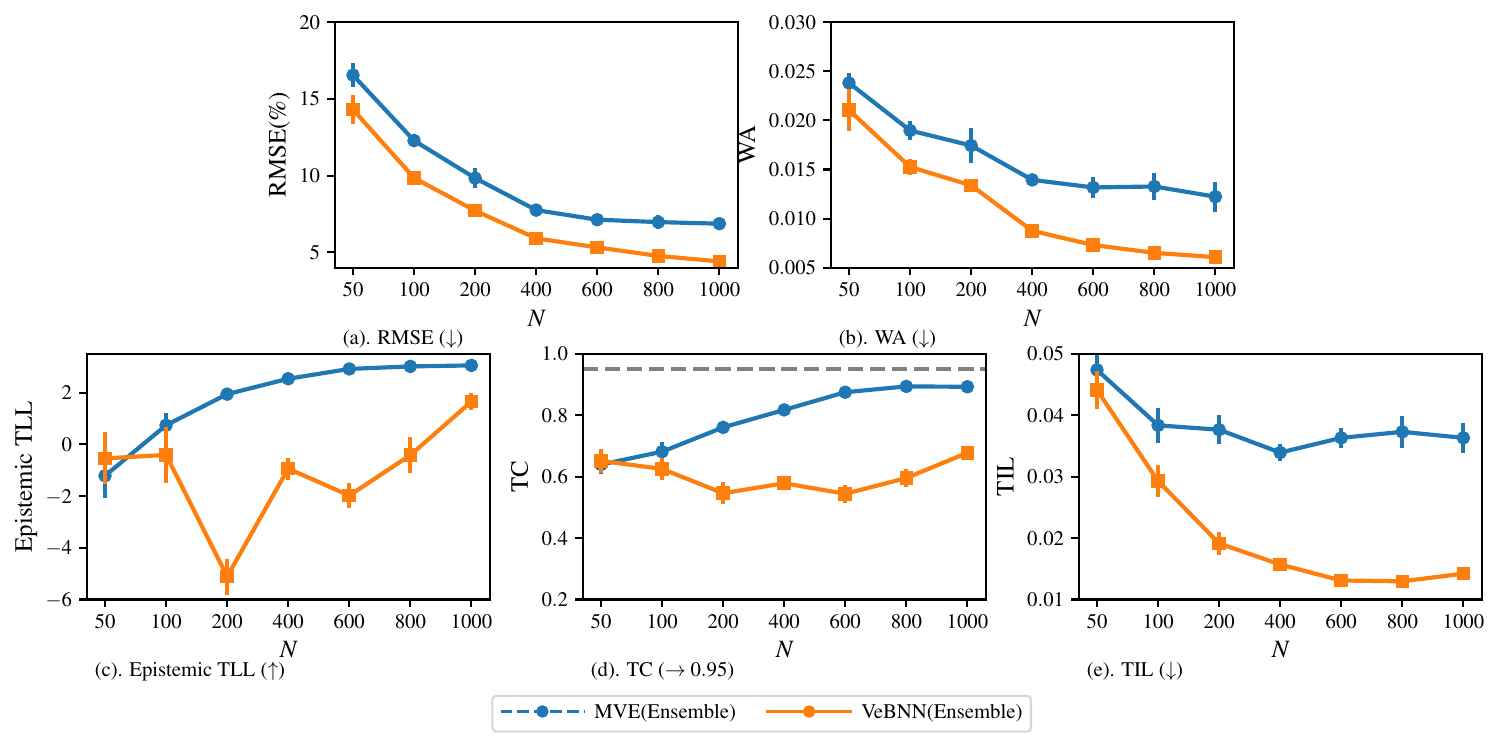}}
\caption{\textbf{Comparison between MVE (Ensembles) and VeBNN (Ensembles) for the plasticity law discovery dataset considering a training set with different number $N$ of training sequences}. }
\label{fig:comparision_ensemble_plasticity}

\end{figure*}

\paragraph{Comparison between MVE (Ensembles) and VeBNN (Ensembles).} The comparison results are presented in \Cref{fig:comparision_ensemble_plasticity}. It is clear that VeBNN (Ensembles) achieves substantially better mean and aleatoric uncertainty predictions, as evidenced by the clear performance gap in RMSE and WA. However, in this case, VeBNN (Ensembles) produces overconfident epistemic uncertainty estimates. This occurs because Deep Ensembles is not a principled Bayesian inference method; its epistemic uncertainty primarily arises from independent restarts of the training procedure. When the loss landscape is well-behaved, like step 3 of VeBNN, the ensemble members are likely to converge to very similar solutions across different runs, which significantly reduces the variability among models and therefore leads to underestimated epistemic uncertainty. Therefore, one should be cautious when using Deep Ensembles as the inference method, though it may contain enough variability in large language models due to the enormous parameter space.

\begin{figure*}[h]
\centering
\centerline{\includegraphics[width=0.95\textwidth]{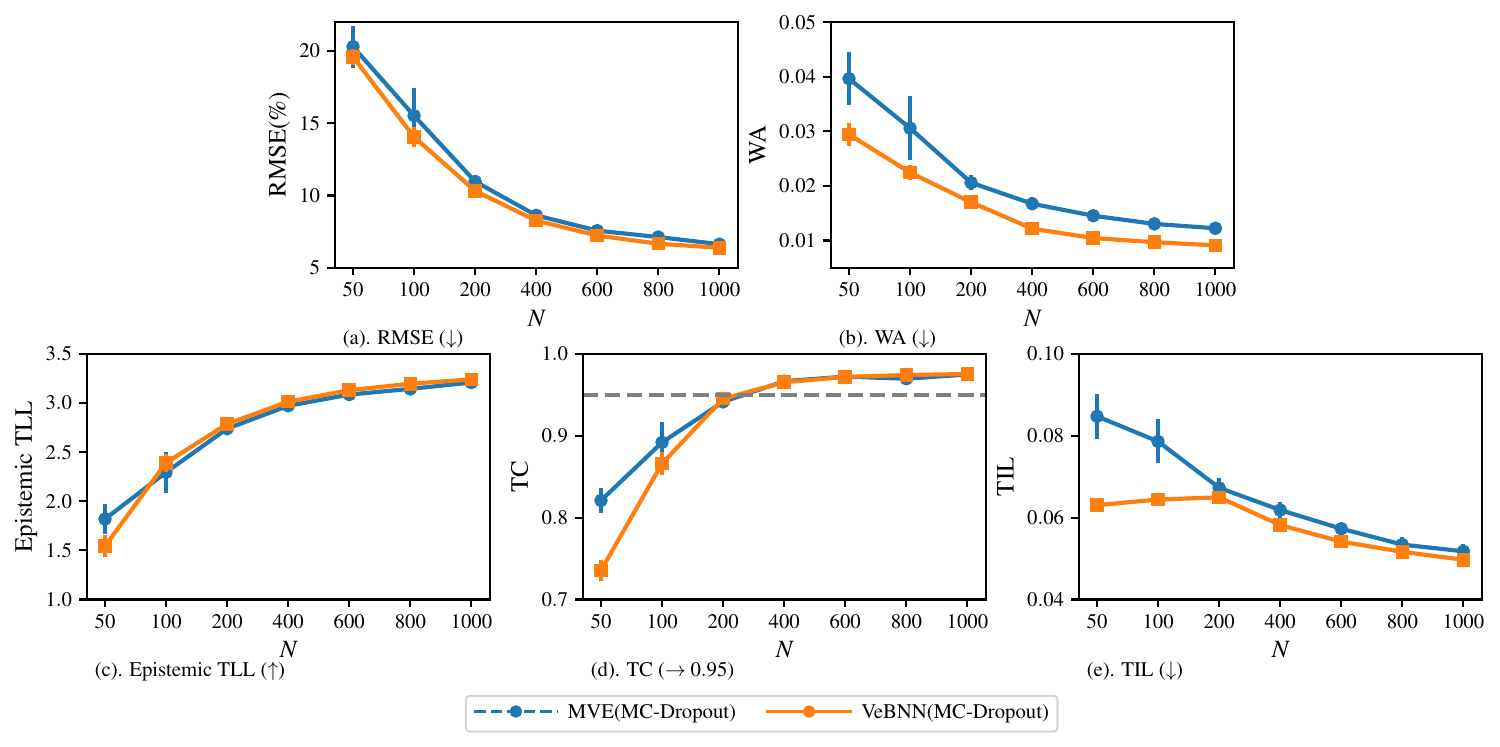}}
\caption{\textbf{Comparison between MVE (MC-Dropout) and VeBNN (MC-Dropout) for the plasticity law discovery dataset considering a training set with different number $N$ of training sequences}.}
\label{fig:comparision_mc_dropout_plasticity}
\end{figure*}

\paragraph{Comparison between MVE (MC-Dropout) and VeBNN (MC-Dropout).} As it shows in \Cref{fig:comparision_mc_dropout_plasticity}, the observation is broadly the same: VeBNN (MC-Dropout) can improve both the mean and aleatoric uncertainty. As for the epistemic uncertainty, the VeBNN (MC-Dropout) gives a small TIL, which leads to slightly smaller Epistemic TLL and TC. With more samples, VeBNN (MC-Dropout) gets better performance as expected.

\begin{figure*}[h]
\centering
\centerline{\includegraphics[width=0.95\textwidth]{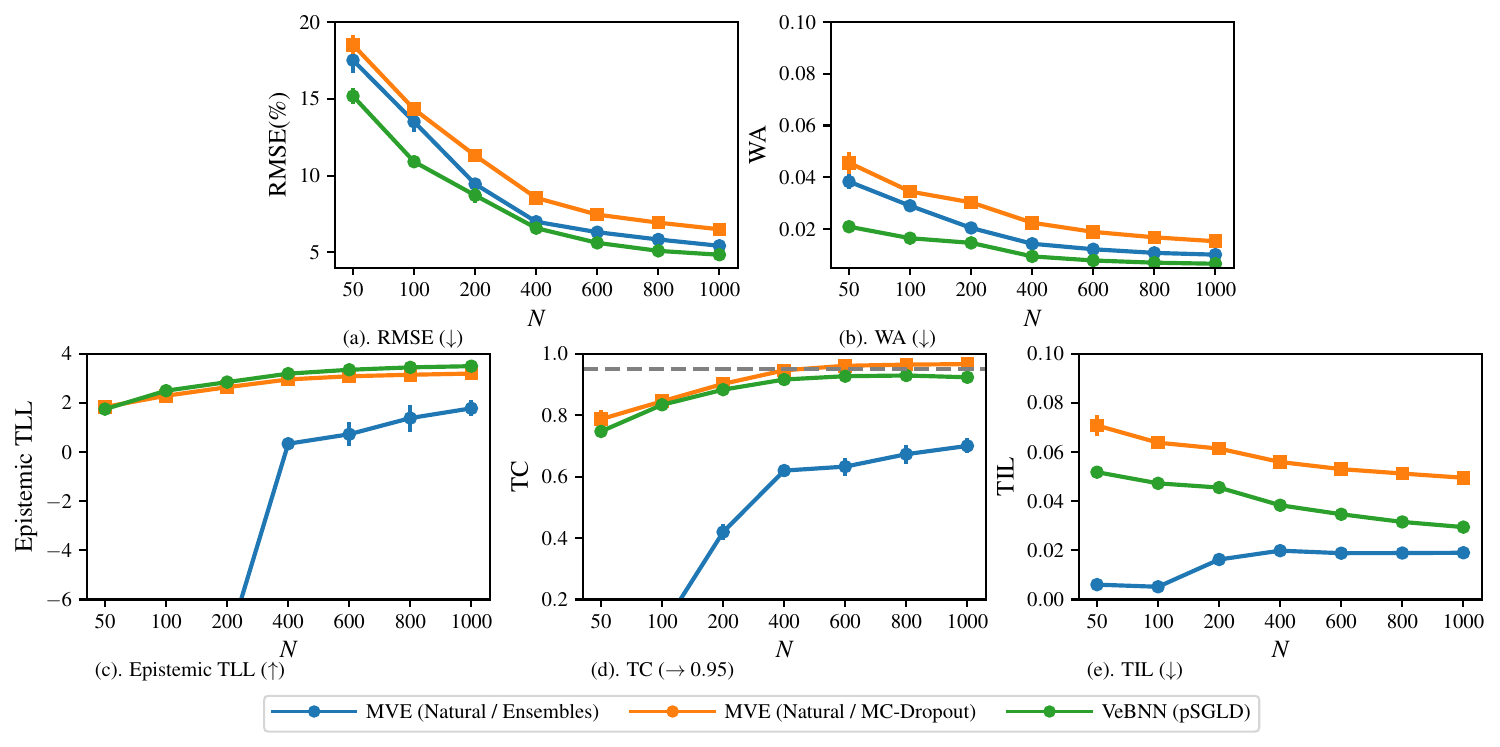}}
\caption{\textbf{Accuracy metrics obtained for the plasticity law discovery dataset considering a training set with different number $N$ of training sequences, comparing with ensemble methods with natural reparameterization loss}.}
\label{fig:plasticity_additional_results}
\end{figure*}

\paragraph{Comparison to additional baselines: MVE (Natural / Ensembles) and MVE (Natural / MC-Dropout).} When using the ensemble methods like deep ensembles and MC-Dropout, we can also adopt natural reparameterization loss to mitigate the gradient imbalance issue as indicated in \Cref{sec:theory_justification_steps1_and_2}. In this paragraph, we additionally provide the results of MVE (Natural / Ensembles) and MVE (Natural / MC-Dropout) \cite{fishkov2025} compared with VeBNN (pSGLD) in \Cref{fig:plasticity_additional_results}. VeBNN outperforms both baselines across all metrics and sample sizes, with the largest differences observed at small sample sizes where natural reparameterization-based methods are less stable.

\begin{figure}[hbt!]
    \centering
    \includegraphics[width=0.8\textwidth]{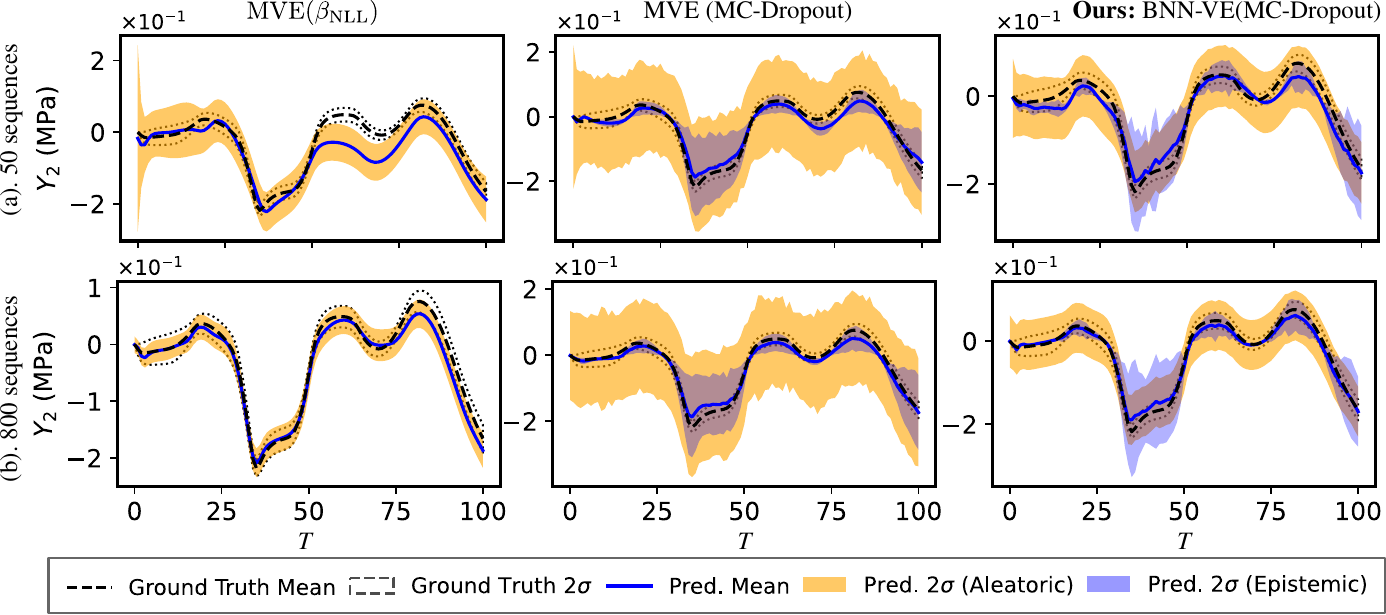}
    \caption{\textbf{Predictions of MVE ($\beta$-NLL), MVE (MC-Dropout), and VeBNN (MC-Dropout) methods on plasticity law discovery dataset.} The upper and bottom rows correspond to 50 and 800 training sequences, respectively.}
    \label{fig:additional_plot_for_plasiticity}
\end{figure}

\paragraph{Uncertainty identification}
Given the limited space in the main text, we present the predictions of MVE ($\beta$-NLL), MVE (MC-Dropout), and VeBNN (MC-Dropout) on the plasticity law discovery dataset in \Cref{fig:additional_plot_for_plasiticity}. We first observe that MVE ($\beta$-NLL) provides reliable aleatoric uncertainty estimates when sufficient training data is available. However, its predictions—both in terms of the mean and aleatoric uncertainty—deteriorate when the training size is reduced to $N=50$. For MVE (MC-Dropout), the predictions remain suboptimal, even with $N=800$ training sequences. In contrast, the proposed VeBNN (MC-Dropout) improves prediction quality, particularly in estimating aleatoric uncertainty.

It is important to note that predictions for plasticity laws are expected to be smooth. The inherent bumpiness of MC-Dropout-based approaches poses challenges for deployment in Finite Element Analysis in real-world applications.

\subsection{Ablation Study on Iteration $K$}
\label{sec:disccussion_iteration}

\begin{figure}[h]
\centering
\centerline{\includegraphics[width=0.8\columnwidth]{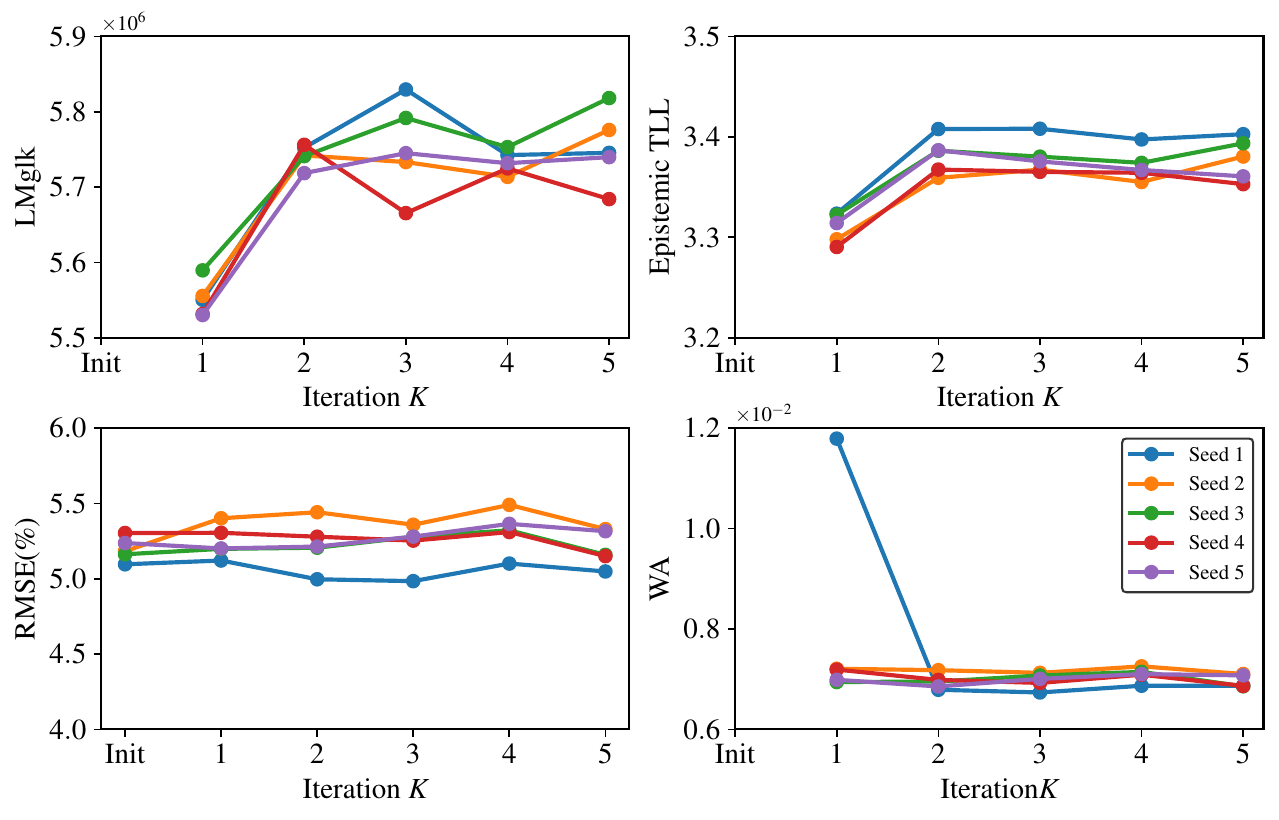}}
\caption{\textbf{Trajectory of performance metrics with respect to the iteration $K$ for 800 training sequences in the plasticity law discovery problem.} In the figures, ``Init'' represents the initialization of training the mean network only as described in \cref{sec:mean_training}, and different curves are realizations under different seeds to restart the procedure, as well as using new samples of training sequences in the dataset.}
\label{fig:trajectory_of_iteration}

\end{figure}

We claim in \Cref{sec:discussion} that the iteration $K$ is not a hyperparameter but a parameter, where we present an experiment conducted based on the plasticity law discovery dataset with $800$ training sequences. The results are shown in \Cref{fig:trajectory_of_iteration}.

According to \cref{fig:trajectory_of_iteration}, the proposed method exhibits rapid convergence, typically requiring only a single iteration to achieve stable performance, highlighting the effectiveness of the cooperative training strategy.
Specifically, the initial warm-up of the mean network provides a favorable starting point for both \textbf{Step 2} (aleatoric modeling) and \textbf{Step 3} (Bayesian training), facilitating accurate uncertainty decomposition in subsequent stages.
However, as observed in the case of seed 1, an unfavorable initialization may lead the warm-up stage to overfit noise. In such cases, additional iterations can effectively mitigate the issue and restore model performance.
Based on empirical evidence, convergence occurs within 2 iterations.

\begin{figure}[h]
\centering
\centerline{\includegraphics[width=0.8\textwidth]{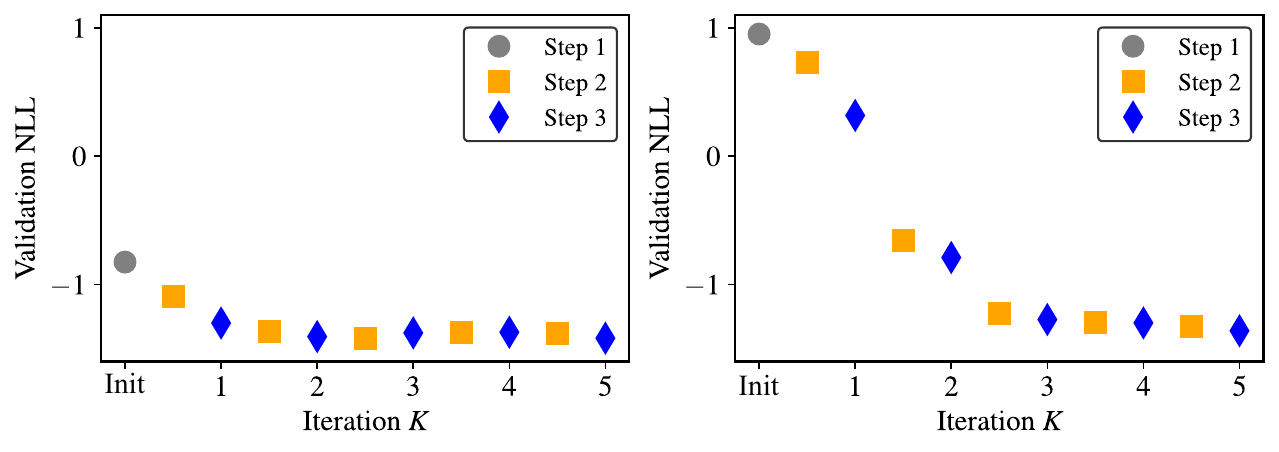}}
\caption{\textbf{Trajectory of validation loss with respect to the iteration $K$ for \emph{Energy} problem randomly picked from UCI regression dataset.} For each iteration, the validation NLL of each step is plotted using different colors. The left subfigure corresponds to a properly trained mean network of step 1, whereas the right subfigure shows the case where the mean network of step 1 is not trained. }
\label{fig:UCI_validation_loss}

\end{figure}

In addition, we randomly select the \emph{Energy} problem from the UCI regression datasets and track its validation NLL over 5 iterations. The results based on VeBNN (pSGLD) are shown in \Cref{fig:UCI_validation_loss}. We observe that VeBNN converges to a similar validation NLL regardless of whether the mean network in \textbf{Step 1} is properly trained. However, starting from an untrained mean network requires more iterations ($K$) to reach convergence.

\paragraph{Additional figure of test coverage and interval length for \cref{fig:ablation_on_gamma_loss}.} It further enhances the robustness of VeBNN (pSGLD) with the additional test coverage and test internal length being almost the same for different $K$ values.
\begin{figure}[h]
\centering
\centerline{\includegraphics[width=0.75\columnwidth]{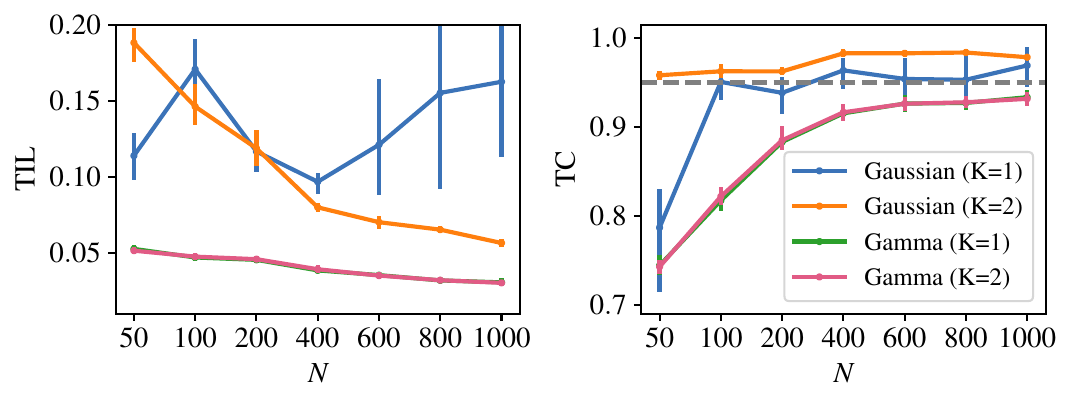}}
\caption{\textbf{TC and TIL metric performance for the ablation study for the Gamma loss and iteration parameter $K$ using VeBNN (pSGLD).}  Curves labeled \emph{Gamma} and \emph{Gaussian} correspond to Step 2 training with the proposed Gamma loss (\Cref{eq:gamma_loss}) and the original Gaussian NLL loss (\Cref{eq:Lvar}), respectively, while fixing $\mu(\mathbf{x};\boldsymbol{\theta})$ from Step 1. }
\label{fig:Ablation_Study_Gamma_K_Additional_Metrics}

\end{figure}

\subsection{Ablation Study on the Size of Variance Estimation Network}
\label{sec:discussion_size_var_net} 

\begin{figure}[hbt!]
\begin{center}
\centerline{\includegraphics[width=\columnwidth]{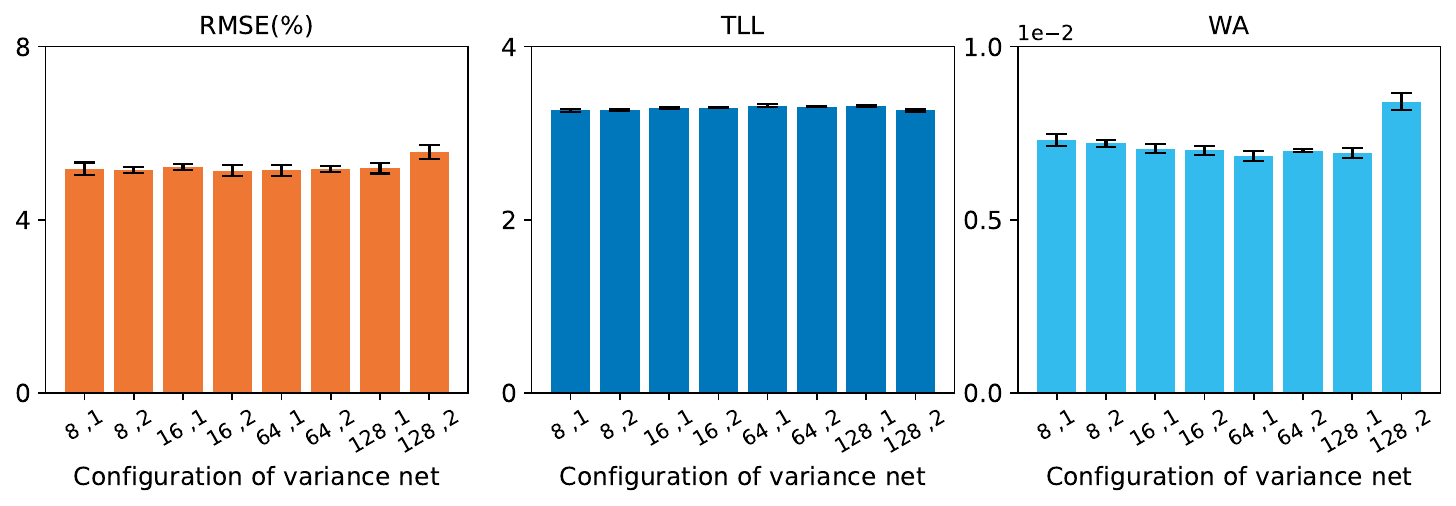}}
\caption{Barplot of all performance metrics with different variance network configurations when considering 800 training points for plasticity law discovery problem. }
\label{fig:barplot_of_different_var_nets}
\end{center}
\end{figure}

In \Cref{sec:discussion}, we also comment on the impact of the size of the variance network, and we provide detailed results of varying the variance of the architecture configurations in \Cref{fig:barplot_of_different_var_nets}. As observed, the predictions do not change significantly when choosing different network architectures.

\newpage
\section{Hyperparameter Settings}
\label{sec:hyperparams_setting}

\subsection{Illustrative Datasets}
\label{sec:hyperparams_synthetic}

\paragraph{Heteroscedastic noise}
\label{sec:hyperparams_heter}
In \cref{sec:synthetic_datasets}, we consider the following one-dimensional example \cite{Skafte2019}:
\begin{equation}
\label{eq:skafte_function}
    y = x \sin{x} + 0.3 \cdot x \cdot \varepsilon_1 + 0.3 \cdot \varepsilon_2
\end{equation}
where $\varepsilon_1, \varepsilon_2 \sim \mathcal{N}(0, 1)$. In the experiments of \cref{sec:synthetic_datasets} and \cref{sec:heter_noise_point}, we sample points uniformly from $\left[0, 10\right]$ for training. We generate 1000 points in $\left[0, 10\right]$ and 1000 points  $\left[-4, 0\right] \cup \left[10,14\right]$ to test the performance of different methods. 

We depict the results of the illustrative example in \cref{fig:illustration}, \cref{fig:illustrative_example},  and \cref{fig:beta_vs_cuq}, for which the hyperparameters are summarized as follows. 
\begin{itemize}
    \item \textbf{Architectures:} We use a two-layer multi-layer perception (MLP) with 256 neurons for the methods that only require one neural network, except BNN (BBB), where a one-layer with 50 neurons is adopted \footnote{We also try the same architecture with other approaches; however, BBB has difficulty with such a large MLP architecture}. \textit{Tanh} function is used as the activation function. We note that only the mean is outputted for the ME network, but there are two outputs, mean and aleatoric variance, for the MVE network. For the proposed cooperative learning strategy, we employ the same architecture for the BNN as the ME network.  An additional one-layer MLP with 5 neurons is employed as the variance neural network to learn data uncertainty.  
    \item \textbf{Optimizer/Inference:} We use \textit{Adam} optimizer with a learning rate of $0.001$ for $20000$ epochs to optimize \cref{eq:Lvar} for all deterministic methods and MC-Dropout (Dropout rate is 0.1 for each layer).  \textit{Adam} is also used to optimize the ELBO of BBB for $10000$ epochs with a learning rate of $0.01$. As for pSGLD, we set the burn-in epoch to be 10000 and collect 100 posterior samples every 100 epochs. We also optimize the variance network for $5000$ epochs with a learning rate of $0.001$ and early stopping of 100 epochs for the proposed cooperative learning strategy.
    \item \textbf{Hyperparameter selection:} We select 70$\%$ data points for training, and the remaining data is used for finding the best hyperparameters, namely number of training epochs and $\beta$ in the case based on the NLL loss value. After identifying the best number of epochs, we use all data points to re-train the model under the best hyperparameters. For the proposed cooperative learning strategy, we feed all the data to \cref{alg: CUQalgorithm} and set the iteration $K=2$.

\end{itemize}

\paragraph{Homoscedastic noise}
\label{sec:hyperparams_homo}
The homoscedastic noise case has the same ground truth, but instead of input-dependent noise, we consider constant noise, expressed as:
\begin{equation}
\label{eq:skafte_function_homo}
    y = x \sin{x} + 0.5 \cdot \varepsilon_2
\end{equation}
where $\varepsilon_2 \sim \mathcal{N}(0, 1)$.

\subsection{UCI Regression Datasets}
\label{sec:hypeparams_uci}

We carefully reviewed the experiments and hyperparameters that were found in other studies using UCI regression datasets \cite{Skafte2019, Immer2023, Seitzer2022}. We used similar configurations to these studies. Our dataset splits and randomizations can be found in our code for reproducibility because the UCI dataset results can be sensitive to data splits \cite{Seitzer2022}. We also considered multiple experiments per dataset to minimize the impact of randomization. The dataset size and input/output dimensions of each problem are listed in \cref{tab:rmse_uci} and \cref{tab:tll_uci} under each column within parentheses.  We report the metrics at the original scale and averaged over all outputs. 

\begin{itemize}
    \item \textbf{Architectures:}  We use a one-layer MLP with 50 neurons followed by \textit{ReLU} activation function for all ME and MVE methods. For the proposed cooperative learning strategy, we employ the same architecture for mean net and BNN, and we additionally use an MLP with 5 neurons followed by \textit{ReLU} activation function as the variance network. 

    \item \textbf{Optimizer/Inference:} The methods, including ME, MVE, BBB, as well as the warm-up of the proposed cooperative learning strategy, are trained via \textit{Adam} for $20000$  epochs with a learning rate of $0.001$ (0.01 for training ELBO), in which the MC-Dropout has a Dropout rate of 0.1 for each layer. As for the pSGLD, we set the burn-in epoch to be 5000 and collected 150 posterior samples every 100 epochs (In total 20000 epochs, which is the same as that of \textit{Adam}). For the additional variance network of the proposed cooperative learning strategy, we use \textit{Adam} to train for $10000$ epochs with an early stopping of 100 epochs. The batch size is set to be $256$ for all approaches. 

    \item \textbf{Hyperparameter selection:} We split each dataset into train-test by $80\%\text{-}20\%$ randomly 20 times. In addition, the train set is further divided into $80\%\text{-}20\%$ for training and validation. We search for the best learning rate within $\left\{ 10^{-4}, 3\cdot10^{-4}, 10^{-3}, 3\cdot10^{-3}, 7\cdot10^{-4} \right\}$ and as well as record the best epoch utilizing the validation NLL loss.  After finding the best learning rate and epoch, the final model is trained using all training data points, and metrics are calculated for the test set. It is noted that the presented results of MVE ($\beta\text{-NLL}$) in \cref{tab:rmse_uci} and \cref{tab:tll_uci} are the overall best results among $\left\{ \beta=0.0, \beta=0.25, \beta=0.5, \beta=0.75, \beta=1.0 \right\}$.  For the proposed cooperative learning strategy, we feed all the data to \cref{alg: CUQalgorithm} and set the iteration $K=2$.
 
\end{itemize}

\subsection{Image Regression Datasets}
\label{sec:hyper_params_image_regression}

We take the image regression dataset introduced in \cite{gustafsson2023reliableregressionmodelsuncertainty}, which consists of relatively large-scale problems, with each containing between 6,592 and 20,614 training images depending on the specific task. Each image has a resolution of $64 \times 64$ and the target is a one-dimensional output $y$. Full details can be found in \cite{gustafsson2023reliableregressionmodelsuncertainty}; a summary of the relevant information is provided below:
\begin{itemize}
    \item \textbf{Cells} The dataset contains $10000$, $2000$, and $10000$ images for training, validation, and testing, respectively. The labels have a range of $[0, 200]$, and there is no distribution shift among all the datasets. 
     \item \textbf{Cells-Tail} The training and validation datasets contain images with labels in the range of $[50, 150]$; the test dataset has labels with a range of $[0, 200]$
     \item \textbf{Cells-Gap} The training and validation datasets contain images with labels in the range of $[0, 50] \cup [150, 200]$; the test dataset has labels with a range of $[0, 200]$
     \item \textbf{ChairAngle} The dataset contains $17640$, $4410$, and $11225$ images for training, validation, and testing, respectively. The labels have a range of $[0.1^\circ, 89.9^\circ]$, there is no distribution shift among all the datasets. 
     \item \textbf{ChairAngle-Tail} The training/validation datasets contain images whose labels have a range of $[15^\circ, 75^\circ]$; and the test labels have a range of $[0.1^\circ, 89.9^\circ]$.
     \item \textbf{ChairAngle-Gap} The labels have a range of $[0.1^\circ, 30^\circ] \cup [60^\circ, 89.9^\circ] $, and the test labels are in $[0.1^\circ, 89.9^\circ]$. 
     \item \textbf{Skin} (SkinLesion) The dataset contains $6592$, $1164$, and $2259$ images for training, validation, and testing, respectively. The dataset contains four different sub-datasets, in which the first three are split into train/val with $85\%/15\%$; and the fourth sub-dataset is used as the test dataset. 
     \item \textbf{Aerial} (AreaBuilding) The dataset contains 180 large aerial images with corresponding building segmentation masks. Specifically, the train/val is obtained from two densely populated American cities while the test dataset is from rural European cities. Overall, it
     contains $11184$, $2797$, and $3890$ images for training, validation, and testing, respectively.
\end{itemize}

We leverage the hyperparameters in \cite{gustafsson2023reliableregressionmodelsuncertainty} and define ours as follows: 
\begin{itemize}
    \item \textbf{Architectures:} ResNet34 backbone \cite{He_2016_CVPR} is employed for this problem. We use a two-layer MLP to decode the prediction into a Gaussian distribution for MVE ($\beta$-NLL), MVE (Ensembles), and MVE (MC-Dropout). It is noted that a dropout layer with a dropout rate of 0.1 is followed for each MLP layer. As for the variance net and deep evidential regression, the decoding layer is set to be the corresponding outputs after the ResNet34 backbone.
    \item  \textbf{Optimizer/Inference} We use \textit{Adam} with a learning rate of $0.001$ for MVE, as well as the warm-up step of the proposed cooperative learning strategy, and employ \textit{Adam} to run for $75$ epochs for the above methods with a batch size of 32. As for the pSGLD, we set the burn-in epochs to be $20$, and we sample 10 posterior samples every 2 epochs (in total $40$ epochs). The additional variance network is trained with \textit{Adam} for $20$ epochs.
    
\end{itemize}

\subsection{Plasticity Law Datasets}
\label{sec:hypeparams_plasticity}

\paragraph{Hyperparameter setting for \cref{sec:experiments}}

In the literature of data-driven constitutive laws, several studies address similar problems without considering noise \cite{Dekhovich2023}. We leverage their setups and define the hyperparameter setting as follows: 

\begin{itemize}
    \item \textbf{Architectures:} For the mean network and BNN, we adopt a two-layer GRU architecture with 128 hidden neurons. It is noted that we only apply the Dropout operation to the hidden-to-decoding layer with a Dropout rate of 0.02. The reason is that this inference method does not show compatible performance when making all weights and biases in the Dropout layer. For the proposed cooperative learning strategy, a smaller GRU network with two layers and 8 hidden neurons is employed for the variance network.
    \item  \textbf{Optimizer/Inference} We use \textit{Adam} with a learning rate of $0.001$ for ME, MVE, as well as the warm-up step of the proposed cooperative learning strategy, and employ \textit{Adam} to run for $2000$ epochs for the above methods. As for the pSGLD, we set the burn-in epoch to be $500$, and we sample 150 posterior samples every 10 epochs (in total $2000$ epochs). The additional variance network is trained with \textit{Adam} for $4000$ epochs with an early stopping patience of 50 epochs.
    \item \textbf{Hyperparameter selection:} For every experiment of different training points, we reserve 100 validation data points from the training dataset to determine the best epoch for ME and MVE. Subsequently, we combine the validation points with the training set and retrain the final model using the best epoch configuration. For the same reason, the results depicted for MVE ($\beta\text{-NLL}$) are the overall best among $\left\{ \beta=0.0, \beta=0.25, \beta=0.5, \beta=0.75, \beta=1.0 \right\}$. For the proposed cooperative learning strategy, we set the iteration $K=2$.
    
\end{itemize}

\paragraph{Hyperparameter setting for \cref{sec:discussion_size_var_net}}

To investigate the influence of the variance network architecture, we consider eight configurations where the number of layers varies from $1$ to $2$, and the number of hidden neurons is set to $\left\{8, 16, 64, 128 \right\}$. The largest configuration, $\left\{128, 2 \right\}$, is identical to the mean network. All other hyperparameters are consistent with those used in the previous section.

\subsection{APPA-REAL Dataset}
The APPA-REAL dataset \cite{agustsson2017appareal} contains 4113 training, 1500 validation, and 1978 testing images. In addition, each image has around 38 votes for the apparent age; therefore, we can know the ground truth for this dataset. We adopt a similar hyperparameter set as the image regression dataset since it is also an image regression problem.  

\subsection{Computational Resources}
Each experiment, except the image regression datasets, is conducted on a node of an HPC cluster platform with an Intel Xeon E5-2643v3 CPU with 6 cores of 3.40GHz and 128 GB of RAM. Regarding the image regression datasets, the experiment is conducted based on a platform with an H100 NVL GPU.

\newpage
\section{Description of Plasticity Law Dataset}
\label{sec:des_plasticity_law}

\begin{figure}[h]
\begin{center}
\centerline{\includegraphics[width=0.9\columnwidth]{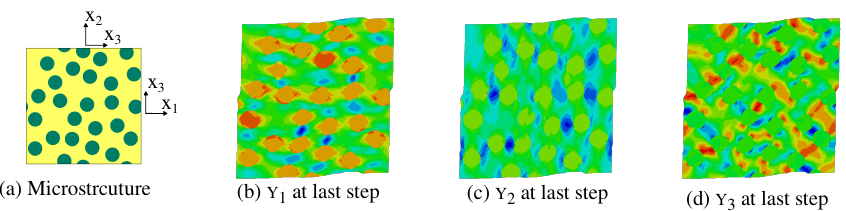}}
\caption{\textbf{Material plasticity law simulation illustration.} Figure. (a) shows an arbitrary realization of material microstructure, and the following figures show the contour plot of this material simulation at the final step.}
\label{fig:sve_contour}
\end{center}
\end{figure}

The fundamental mechanical law of materials is called a constitutive law. It relates average material deformations to average material stresses at any point in a structure. Constitutive laws can model different Physics behaviors, such as elasticity, hyperelasticity, plasticity, and damage. In this paper, we focus on generating datasets for plastically deforming composite materials, coming from prior work \cite{yi2023rvesimulator,Dekhovich2023,Mozaffar2019}. Without loss of generality, the constitutive law of such path-dependent materials can be formulated as follows:
\begin{equation}
\label{eq:plastic_law}
    \mathbf{y}  = f(\mathbf{x}, \Dot{\mathbf{x}}, \tau, \Dot{\tau}, \bm{h})
\end{equation}
\noindent where $\mathbf{y}$, $\mathbf{x}$, $\tau$ are stress, strain, and temperature respectively, $\bm{h}$ is a set of internal variables. The constitutive law can be predicted by micro-scale simulations of material domains that are called stochastic volume elements (SVEs) -- see \cref{fig:sve_contour}. These SVEs are simulated by rigorous Physics simulators based on the Finite Element Method (FEM). Each material SVE is utilized as the basic simulation unit. Many factors bring data uncertainty into the data generation process; we focus on data uncertainties from two aspects: (1) SVE size; and (2) particle distribution. As we randomize particle distribution, the stress obtained for an input deformation exhibits stochasticity (aleatoric uncertainty). Therefore, two datasets are generated from simulations according to \cref{tab:simulation_params_configuration}. 

\begin{table*}[h]
\caption{Parameter configuration material plasticity law simulations (Units: $\text{SI}(mm)$ ).}
\label{tab:simulation_params_configuration}
\centering
\resizebox{\textwidth}{!}{%
\begin{tabular}{lcccclcccc}
\toprule
\multirow{2}{*}{Name} & \multicolumn{3}{c}{Microstructure Parameters} & \multirow{2}{*}{Hardening Law}                              & \multirow{2}{*}{$E_{\mathrm{fiber}}$} & \multirow{2}{*}{Size} & \multirow{2}{*}{$E_{\mathrm{matrix}}$} & \multirow{2}{*}{$\nu_{\mathrm{matrix}}$} & \multirow{2}{*}{$\nu_{\mathrm{fiber}}$} \\
\cmidrule(lr){2-4}
                      & $v_f$               & $r$               & $r_{\mathrm{std}}$ &                                                      &                                       &                      &                                       &                         &                         \\
\midrule
Material 1                   & 0.30               & 0.003             & 0.0                & $\sigma_{y} = 0.5 + 0.5(\Bar{\epsilon})^{0.4}$      & 1                                     & 0.048                & 100                                   & 0.30                    & 0.19                    \\
Material 2                 & 0.30               & 0.003             & 0.0                & $\sigma_{y} = 0.5 + 0.5(\Bar{\epsilon})^{0.4}$      & 1                                     & 0.030                & 100                                   & 0.30                    & 0.19                    \\
\bottomrule
\end{tabular}%
}

\end{table*}

\begin{figure}[hbt!]

\begin{center}
\centerline{\includegraphics[width=\columnwidth]{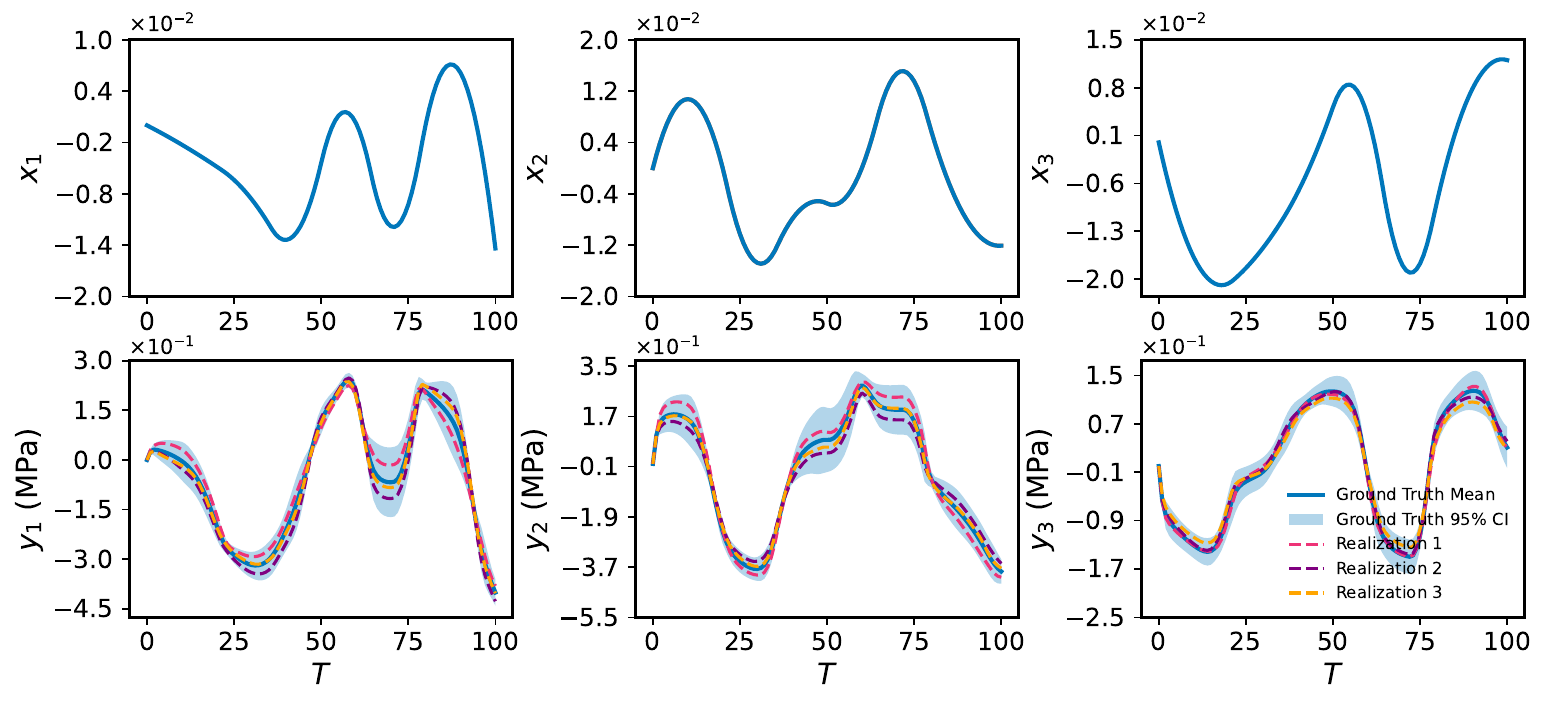}}
\caption{\textbf{Plasticity law data illustration.} The first row is the strain inputs for the material law simulation, and the second row is the stress outputs. Each dashed line represents one particle material microstructure realization in \cref{fig:sve_contour}; the mean and the confidence interval are obtained via multiple realizations.}
\label{fig:plasticity_data_noise}
\end{center}

\end{figure}

According to \cref{tab:simulation_params_configuration}, we have two materials that have different SVE sizes that control the noise sources of the data. Materials with a smaller SVE size have larger data noise. \cref{fig:sve_contour}  and \cref{fig:plasticity_data_noise} illustrate details of the simulations and how the uncertainty in the data originates. Specifically, according to the microstructure configuration in \cref{tab:simulation_params_configuration}, we generate SVEs using the Monte Carlo Sampling strategy \cite{Melro2008} and simulate the stress responses through the commercial software ABAQUS \cite{abaqus2024} with the input of the strain sequence shown in the first row of \cref{fig:plasticity_data_noise}. After simulation, we get a series of contours of stress components in \cref{fig:sve_contour} and average the field (color contours) for each input sequence point, leading to the output sequence. An example of 3 realizations of SVEs and the corresponding 3 output response sequences is shown as dashed lines in the second row of \cref{fig:plasticity_data_noise}. Each realization corresponds to a randomization of the microstructure of the material (particle distribution). By running multiple realizations, we can obtain the statistics of those strain inputs. By definition, the variation in the stress output is the uncertainty of the data. 

Each input deformation sequence and output stress sequence used in training contains 100 points. In total, we use 50 different SVEs to ensure that we have enough realizations to calculate the ground truth aleatoric uncertainty. Overall, we simulate $1000$ sequences for training and $100$ sequences for testing, respectively,  for each problem shown in \cref{tab:simulation_params_configuration}.

The new dataset is made available as open-source in the hope of creating a more interesting problem for assessing future methods, because we had difficulties in finding more challenging heteroscedastic problems with ground truth aleatoric uncertainty to assess our method. This dataset is three-dimensional and history-dependent, i.e., $\mathcal{D} = \{\mathbf{x}_{n,t}, \mathbf{y}_{n,t}\}$ with features $\mathbf{x}_{n,t} \in \mathbb{R}^{3}$ and targets $\mathbf{y}_{n,t} \in \mathbb{R}^3$, where $n=1,..., N$ are the training sequences (deformation paths) and $t=1,...,T$ are the points in each sequence. We highlight two aspects about this dataset. First, the targets $\mathbf{y}$ are history-dependent, so estimating a new state $\mathbf{y}'$ requires to know the sequence of states needed to reach that state, i.e., regression requires recurrent neural network architectures \cite{Dekhovich2023}, specifically adopting a Gated Recurrent Unit (GRU) architecture \cite{cho2014learning, Gan2017} in this work. Furthermore, the dataset was created synthetically by physically-accurate computer simulations of materials under mechanical deformation, so it was possible to generate enough data to determine the ground-truth aleatoric uncertainty (arising from variations within the material). In other words, we have a good estimate of the heteroscedastic noise in the data. 

\end{document}